\documentclass[11pt]{article}

\PassOptionsToPackage{table,x11names,dvipsnames}{xcolor}
\usepackage[preprint]{acl}

\usepackage{times}
\usepackage{latexsym}
\usepackage{multirow}
\usepackage{algorithm}
\usepackage{algpseudocode}

\usepackage[T1]{fontenc}
\usepackage[utf8]{inputenc}

\usepackage{microtype}

\usepackage{inconsolata}

\usepackage{graphicx}

\usepackage{xcolor}
\usepackage{colortbl}

\usepackage{titletoc}
\newcommand{\authcount}[1]{}
\usepackage{xstring,catchfile}
\usepackage{booktabs}
\usepackage{tabularx}
\usepackage{subcaption}
\usepackage{siunitx}
\usepackage{wrapfig}
\usepackage{enumitem}

\usepackage[most]{tcolorbox}
\tcbuselibrary{skins, raster, theorems, listings, breakable}

\usepackage{pgfplots}
\pgfplotsset{compat=1.18}
\usepackage[
  skip=6pt,              % space between float body and caption
  belowskip=4pt,         % space below caption
  aboveskip=4pt          % space above caption (when caption is above the float)
]{caption}

\definecolor{winGreen}{RGB}{76,175,80}        % unified (was previously defined twice)
\definecolor{lossRed}{HTML}{CC0000}
\definecolor{bestGreen}{RGB}{34,139,34}

\definecolor{gpt5col}{RGB}{31,119,180}
\definecolor{gemcol}{RGB}{214,96,40}

\definecolor{agentblue}{HTML}{146082}

\definecolor{headerblue}{HTML}{2C3E50}
\definecolor{headertextwhite}{HTML}{FFFFFF}
\definecolor{row1}{HTML}{F0F4F8}
\definecolor{row2}{HTML}{FFFFFF}
\definecolor{catblue}{HTML}{E8EEF4}

\definecolor{grayGPT5}{gray}{0.96}
\definecolor{blueGPT4}{rgb}{0.94, 0.96, 1.0}
\definecolor{greenQwen}{rgb}{0.95, 0.99, 0.95}

\definecolor{deltaTxt}{HTML}{0A7D2C}
\definecolor{preprocessingcolor}{HTML}{14305C}
\definecolor{outlinecolor}{HTML}{247440}
\definecolor{outlinerevisioncolor}{HTML}{5C3898}
\definecolor{slidegenerationcolor}{HTML}{E07024}
\definecolor{sliderevisioncolor}{HTML}{D03C64}
\definecolor{algcomment}{HTML}{6A5ACD}        % SlateBlue
\definecolor{convuserblue}{HTML}{1F6FEB}

\definecolor{heatLow}{HTML}{F4F8F4}
\definecolor{heatMid}{HTML}{C9E2CB}
\definecolor{heatHi}{HTML}{6FB575}
\definecolor{oursRow}{HTML}{ECEAF6}
\definecolor{avgTint}{HTML}{EEEEEE}

\newcommand{\convuser}[1]{\textit{#1}}

\newcommand{\alc}[1]{\textcolor{algcomment}{\textit{// #1}}}

\newcommand{\preprocessing}[1]{\textcolor{preprocessingcolor}{\textbf{#1}}}
\newcommand{\outlinegen}[1]{\textcolor{outlinecolor}{\textbf{#1}}}
\newcommand{\outlinerev}[1]{\textcolor{outlinerevisioncolor}{\textbf{#1}}}
\newcommand{\slidegen}[1]{\textcolor{slidegenerationcolor}{\textbf{#1}}}
\newcommand{\sliderev}[1]{\textcolor{sliderevisioncolor}{\textbf{#1}}}

\usepackage{amsmath,amsfonts,bm}
\usepackage{booktabs}
\usepackage{tabularx}

\def\eqref#1{equation~\ref{#1}}
\def\1{\bm{1}}

\DeclareMathAlphabet{\mathsfit}{\encodingdefault}{\sfdefault}{m}{sl}
\SetMathAlphabet{\mathsfit}{bold}{\encodingdefault}{\sfdefault}{bx}{n}

\title{ConvDeck: Conversational Paper-to-Slide Generation\\via Stage-Specific User Feedback}

\author{Tarik Can Ozden$^{*}$, Sachidanand VS$^{*}$, Furkan Horoz$^{*}$, \\ \textbf{Ozgur Kara, Dilek Hakkani-Tür$^\dagger$, Junho Kim$^\dagger$, James M. Rehg$^\dagger$} 
\\ University of Illinois Urbana-Champaign}

\begin{document}
\maketitle
\begin{abstract}

Automatic academic paper-to-slide generation is inherently iterative, because creating an effective presentation requires repeated cycles of generation, critique, and revision. Recent multi-agent systems partially acknowledge this through internal critique-and-revise loops, while conversational approaches allow users to refine generated slide decks through dialog. However, these refinement processes either remain largely closed to the user or introduce feedback only after a complete deck has been produced, limiting the user’s ability to participate in the iterative refinement of narrative flow, content allocation, and presentation emphasis.
To address this gap, we introduce ConvDeck, a multi-agent pipeline for conversational paper-to-slide generation that distributes interaction across the pipeline through stage-specific loops, allowing users to iteratively refine both the presentation outline and the final slide deck at the stages where each kind of decision is made. These loops are driven by a refinement mechanism in which agents can think, speak, and act, enabling them to either directly apply edits or respond conversationally to clarify user feedback and discuss revision options. Our evaluation shows that stage-specific conversational feedback improves user-goal satisfaction while preserving narrative coherence, content quality, and visual presentation. The codebase is available on our \href{https://convdeck.github.io}{project webpage}.

\end{abstract}

\begingroup
\renewcommand{\thefootnote}{}
\footnotetext{$^*$Equal contribution~\quad $^\dagger$Corresponding author}
\endgroup
\setcounter{footnote}{0}

\section{Introduction}

Academic paper-to-slide generation aims to transform a research paper into a coherent set of presentation slides (\textit{i.e.,} a presentation deck) by identifying the relevant content, organizing it into a clear narrative, deciding what to emphasize, and adjusting the level of detail for the intended presentation context. These decisions are not determined by the source paper alone: depending on the presenter's goals, audience, time budget, and preferred narrative emphasis, the same paper can give rise to many valid decks, such as methodology-focused, results-oriented, or instructional ones. 

\begin{figure*}[t]
\centering
\includegraphics[width=\textwidth]{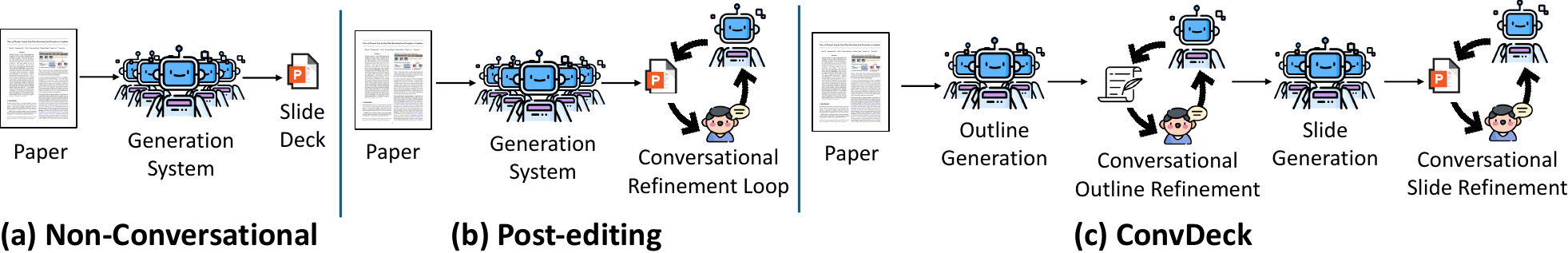}
\caption{\textbf{Comparison to prior paper-to-slide generation paradigms.} \textit{ (a) Non-conversational methods} produce a slide deck from the paper in a single generation pass, with no interaction. \textit{(b) Post-editing methods} add a conversational refinement loop \emph{after} the deck is generated, so all user feedback must be absorbed by a single post-hoc editing stage. \textbf{(c) ConvDeck} distributes interaction across the pipeline through stage-specific loops.}
\label{fig:prev_comparison}
% \vspace{-0.8em}
\end{figure*}

Early slide generation works used an LLM to produce a presentation deck in a single pass, but recent systems for automatic slide generation decompose the authorship process into multiple specialized components~\cite{zheng2025pptagent, ge2025autopresent}, such as visual design~\cite{liang2025slidegen, pan2026aeslides} and narrative flow~\cite{ozden2026arcdeck, yu2026paperx}, as part of a multi-stage design approach (see Fig.~\ref{fig:prev_comparison} (a)). A key property of recent approaches is the incorporation of internal \emph{critique-and-revise loops} that iteratively polish the generated deck~\cite{xu2025pregenie, zheng2026deeppresenterenvironmentgroundedreflectionagentic, liu2026presenting}. This is a natural approach, given that human authorship of high quality documents also requires multiple iterations of content generation, critique, and revision.

In order to provide greater user control over slide generation, recent works have introduced conversational control through human-in-the-loop refinement~\cite{yang2025autoslides, jang2026deckbenchbenchmarkingmultiagentframeworks, jung2025talk, chen2025presentcoach}, where the user provides direct feedback on a \emph{final} generated slide deck (Fig.~\ref{fig:prev_comparison} (b)). Providing feedback at this stage makes sense, because the user needs to see the finished slides in order to critique them. 

However, the direct revision of generated slides through dialog is complex and unwieldy, because many details of the slide deck are shaped by decisions made much earlier in the pipeline, including outline structure, content allocation, and narrative order. Addressing structural issues via surface-level language-based control of the slide content is inefficient and unreliable. In contrast, the critic agents in a critique-and-revise loop have the ability to directly modify the intermediate representations that produce the final slide content. However, these agents can never completely capture the critical perspective of the user, because they lack the user's unique background, experiences, and a full understanding of their goals.

To address this gap, we introduce \emph{ConvDeck}  (Fig.~\ref{fig:prev_comparison} (c)), a multi-agent pipeline for conversational paper-to-slide generation that \emph{enables user interaction across the generation pipeline} through stage-specific loops, aligning each refinement step with the component best suited to act on it. Conv\-Deck integrates interaction at two stages: (i) outline generation, where users can revise high-level decisions such as narrative flow, section emphasis, slide ordering, and content coverage before the full deck is rendered, and (ii) slide generation, where users can further refine slide-level content, visual organization, figures, layout, and other deck details. 

Within each stage, our conversational agents operate through a refinement mechanism in which agents can reason, speak, and act, allowing them to either directly apply edits or respond conversationally to request clarification, explain refinement choices, and discuss alternative revisions when user feedback is underspecified or involves ambiguous presentation goals. To assess whether this design actually translates into better alignment with user intent, our comprehensive evaluation measures both standard slide-quality dimensions and explicit user-goal satisfaction, studying whether stage-specific conversational feedback improves the system's ability to incorporate user requests while preserving narrative coherence, content quality, and visual presentation.

In summary, our main contributions are:
\begin{itemize}
\item We introduce ConvDeck, a multi-agent pipeline for conversational paper-to-slide generation that gives users fine-grained control through stage-specific refinement loops at both outline generation and slide generation, aligning each refinement step with the pipeline component best suited to act on it.
\item We develop a refinement mechanism in which our conversational agents think, speak and act, allowing them to either directly apply edits or respond conversationally to clarify user feedback, explain revision decisions, and discuss possible refinements before applying changes.
\item We propose a user-goal satisfaction evaluation for conversational paper-to-slide generation, measuring whether stage-specific conversational feedback leads to decks that better satisfy presenter-specific requirements while preserving standard slide-quality criteria.
\end{itemize}

\begin{figure*}[t]
\centering
\includegraphics[width=0.97\textwidth]{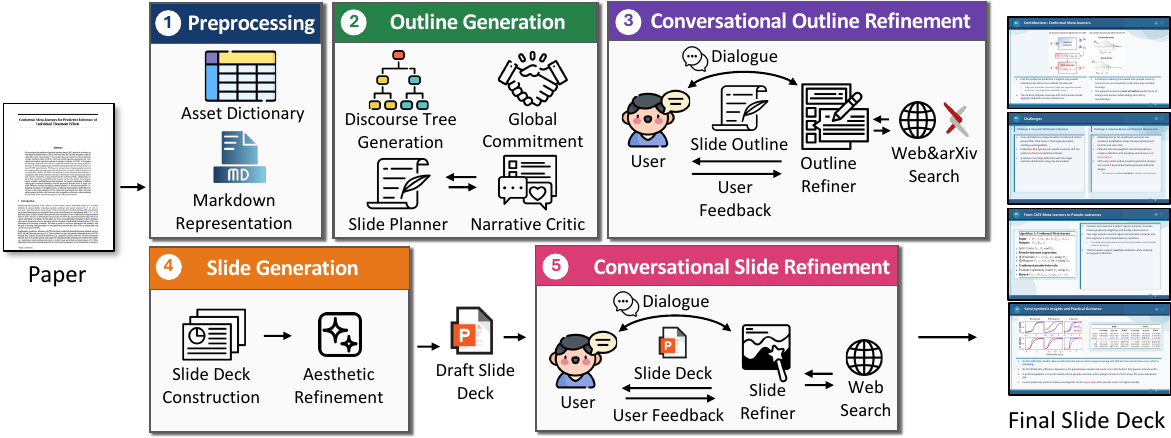}
\caption{\textbf{Overview of ConvDeck.} 
Given a paper PDF and optional user inputs, ConvDeck produces a final slide deck through five stages: \preprocessing{Preprocessing}, \outlinegen{Outline Generation}, \outlinerev{Conversational Outline Refinement}, \slidegen{Slide Generation}, and \sliderev{Conversational Slide Refinement}.}
\label{fig:slide_generation}
% \vspace{-0.8em}
\end{figure*}

\section{Related Work}

\paragraph{Multi-agent paper-to-slide generation.}
Recent automatic slide generation systems adopt agentic, multi-stage pipelines that decompose the task into specialized components for planning, content selection, slide construction, and visual refinement. One line of work treats generation as a reference- or template-driven process~\cite{zheng2025pptagent, ge2025autopresent, tang2025slidecoder}, while another centers narrative or discourse structure~\cite{ozden2026arcdeck, liang2025slidegen, yu2026paperx}. A third line introduces refinement loops internal to the pipeline, in which agents iteratively critique and revise generated artifacts~\cite{xu2025pregenie, zheng2026deeppresenterenvironmentgroundedreflectionagentic, liu2026presenting, pan2026aeslides}. A complementary thread targets personalization and audience-aware adaptation~\cite{zeng2026slidetailor, liu2025addressing}. Related agentic systems extend the paradigm to academic posters~\cite{pang2026paper2poster, zhang2025postergen, Inadumi_2026_SciPostGen, shi2026apexacademicposterediting, tang2026efficientpostergen} and presentation videos~\cite{zhu2025papervideo}. While these agentic works have improved slide quality substantially, they do not enable the user to easily modify the deck if the final output does not meet all of their requirements. More fundamentally, this non-interactive approach is not compatible with the cycle of generation, critique, and revision which defines human authorship of high quality documents.

\paragraph{Conversational paper-to-slide generation.}
A smaller but growing line of work creates interactive interfaces in which a user can provide natural-language feedback to revise the generated deck~\cite{yang2025autoslides, jang2026deckbenchbenchmarkingmultiagentframeworks, jung2025talk, chen2025presentcoach}. 
These systems share a common design: conversation is introduced \emph{after} a complete deck has been generated, and the user's revisions are limited to post-hoc editing of the final slides. This is a laborious and cumbersome process, as the final slides entangle surface-level properties, like font size, with deeper structural properties, such as the topic ordering. In contrast, ConvDeck distributes interaction across both the outline-generation and slide-generation stages, aligning each refinement step with the pipeline component that is best suited to act on it.

Prior works evaluate slide generation using slide-quality metrics, similarity to reference slide decks~\cite{jang2026deckbenchbenchmarkingmultiagentframeworks}, or how effectively users learn the underlying content through slide editing~\cite{yang2025autoslides}. In addition to standard slide-quality evaluations, we introduce the first explicit evaluation of user-goal satisfaction in slide generation, assessing how well conversational slide generation systems fulfill a diverse set of randomly sampled presenter-specific requirements.

\section{Conversational Paper-to-Slide Generation via ConvDeck}
\label{sec:method}

\subsection{ConvDeck Overview}
\label{sec:method_overview}
As illustrated in Fig.~\ref{fig:slide_generation}, ConvDeck takes a paper PDF as input, together with two optional user-provided inputs: a target audience, and a presentation duration. From these inputs, it produces a final slide deck through the five-stage sequential pipeline. \preprocessing{Preprocessing} (Stage 1) extracts a structured text and asset representation of the paper, \outlinegen{Outline Generation} (Stage 2) builds an initial draft outline through discourse-aware planning, \outlinerev{Conversational Outline Refinement} (Stage 3) iteratively updates this outline based on feedback from a \convuser{user} before any slides are rendered, \slidegen{Slide Generation} (Stage 4) converts the refined outline into a draft slide deck, and \sliderev{Conversational Slide Refinement} (Stage 5) further refines the rendered deck through a second round of feedback from the \convuser{user}. The rest of this section describes each stage in detail. 

\subsection{ConvDeck Pipeline}
\label{sec:method_stages}

\paragraph{(Stage 1) \preprocessing{Preprocessing.}}
The Preprocessing stage parses the input paper into two artifacts that ground all subsequent generation: a markdown representation of the paper text and an asset dictionary containing its figures and tables alongside their captions. We use Docling~\cite{livathinos2025docling} for parsing and asset extraction. Docling deterministically parses the full paper body, preserving the information for later stages. We additionally separate the references section and build a citation-key dictionary that downstream stages use to preserve in-text citations; supplementary content appearing after the references is discarded.

\paragraph{(Stage 2) \outlinegen{Outline Generation.}}
The Outline Generation stage produces an initial slide outline from the markdown representation. Following Arc\-Deck~\cite{ozden2026arcdeck}, we adopt a narrative-driven design with three components: (i) a \textit{Discourse Parser} that builds an RST-based discourse tree to capture rhetorical relations between paragraphs, (ii) a \textit{Commitment Builder} that consumes the target audience and presentation duration to produce a global commitment summarizing the deck's high-level intent, and (iii) \textit{Narrative Refinement}, which drafts and revises the outline through a Slide Planner/Reviser and Narrative Critic cycle (see App.~\ref{app:arcdeck_details} for details). Since user-driven refinement is our focus, we run this refinement loop once and defer fine-grained outline revisions to Stage~3.

\begin{figure}[t]
    \centering
    \includegraphics[width=\columnwidth]{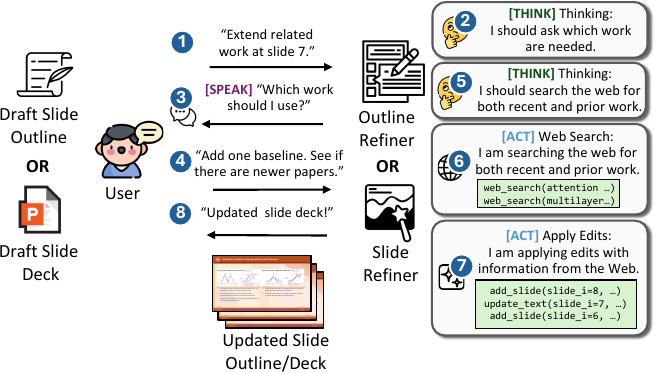}
    \caption{\textbf{Conversational Refinement Overview.} Given a draft outline or deck, the user provides feedback. The refiner reasons about the input, either speaks with the user or acts through tools, and returns an updated outline or deck for iterative refinement.}
    \label{fig:refinement}
\end{figure}

\paragraph{(Stage 3) \outlinerev{Conversational Outline Refinement.}}
As exemplified in Fig.~\ref{fig:refinement}, this stage updates the initial outline according to feedback from the \convuser{user}, allowing the high-level structure and narrative to be revised before any slides are rendered. The \convuser{user} reviews a preview consisting of slide titles and discussion ideas summarizing the slide content. To support more informed feedback, the \convuser{user} is also provided with a summarized version of the paper. Based on this feedback, an \textit{Outline Refiner} applies the requested modifications through editing operations such as adding, editing, splitting, merging, removing, and reordering slides.

To ensure the refiner faithfully captures the \convuser{user}'s intent, we adopt a \emph{speak--act refinement} inspired by ReSpAct~\cite{dongre2024respactharmonizingreasoningspeaking}, in which the agent can communicate and clarify before executing modifications. The agent operates in one of two modes: \textit{Think+Speak}, where it reasons about the request and asks a clarification question when information is missing, and \textit{Think+Act}, where it reasons about the feedback and applies the appropriate edits (see App.~\ref{app:respact_details} for details). To augment the deck with information beyond the source paper when requested (\textit{e.g.,} related work or additional baseline results), the agent is equipped with an arXiv retrieval tool. To keep refinement efficient, every edit is implemented as a local update through a dedicated editing function, so the agent supplies only the affected content rather than regenerating the full outline at each turn.

\paragraph{(Stage 4) \slidegen{Slide Generation.}}
The Slide Generation stage converts the refined outline into a draft slide deck. Following ArcDeck~\cite{ozden2026arcdeck}, it is built on two agents: (i) a \textit{Slide Deck Constructor} that selects relevant figures and tables, chooses a slide layout that accommodates the visuals and textual content, and generates concise bullet points and sub-bullet points, producing a slide specification in structured JSON; and (ii) an \textit{Aesthetic Refiner} that performs a final pass by adding visual content to slides without figures or tables, expanding bullet points on sparse slides, and applying boldface and color emphasis to highlight key information.

\paragraph{(Stage 5) \sliderev{Conversational Slide Refinement.}}
The Conversational Slide Refinement stage applies a second round of revision on the \emph{rendered} slides, driven by feedback from the \convuser{user} (Fig.~\ref{fig:refinement}). Unlike the Aesthetic Refiner in Stage 4, which only operates on the slide specification, this stage observes the actual visual output and can therefore correct rendering issues such as text overflow, undersized figures, and crowded layouts. The \convuser{user} reviews the rendered slide images and provides natural-language feedback, and a \textit{Slide Refiner} applies localized modifications to the affected slides rather than regenerating the deck from scratch. The agent follows the same speak--act mechanism as Stage 3, supporting insertion, deletion, modification, splitting, merging, reordering, repositioning, resizing, and typography operations. It is additionally equipped with a web search tool that enables retrieval of up-to-date information or supplementary details requested during refinement.

Direct manipulation of PPTX files is challenging because a presentation is stored as a set of interdependent XML documents describing content, styling, and layout, so even localized modifications often require coordinated updates across multiple files. We therefore translate the slide specification into JavaScript code that is compiled into a PPTX file using PPTXGenJS~\cite{gitbrent2024pptxgenjs}, providing a structured and editable view of slide elements that serves as the editable state throughout the refinement process.

\section{Experimentation}
\label{sec:experiments}

\subsection{Evaluation Setup}
\label{sec:evaluation_setup}
\paragraph{Baselines and dataset.}
We compare ConvDeck against two sets of baselines. The first set consists of five \textit{non-conversational baselines}: (i) \textbf{HTML}, a prompt-based baseline that generates the deck as HTML in a single LLM call; (ii) \textbf{PPTAgent}~\cite{zheng2025pptagent}, an edit-based pipeline that constructs slides by editing reference slides; (iii) \textbf{SlideGen}~\cite{liang2025slidegen}, a multi-agent framework emphasizing layout and visual design; (iv) \textbf{SlideTailor}~\cite{zeng2026slidetailor}, a multi-agent framework that personalizes the deck to a target audience; and (v) \textbf{ArcDeck}~\cite{ozden2026arcdeck}, a multi-agent framework emphasizing discourse-aware outline planning. These methods produce the full deck in a single pass and cannot accommodate user feedback. The second set consists of two \textit{conversational baselines} that apply user feedback as post-hoc edits on the fully rendered deck: \textbf{ConvHTML}, a conversational extension of the HTML baseline that we construct by adding a post-generation revision loop, and \textbf{AutoSlides}~\cite{yang2025autoslides}, the current state-of-the-art conversational system. 
% ConvDeck differs from both sets by distributing feedback across pipeline stages, allowing high-level outline decisions to be revised before the deck is rendered and slide-level details to be refined afterward. 
We conduct all experiments on the 100-paper ArcBench~\cite{ozden2026arcdeck} benchmark, a curated set of paper--slide pairs drawn from major machine learning venues and accompanied by author-prepared reference slide decks.

\paragraph{Implementation details.}
In all conversational experiments, the \convuser{user} can be either a real human user or an LLM-based simulator that emulates one; we use the simulator throughout for reproducibility, and its full
prompt setup is described in App.~\ref{app:user_simulator}. Full implementation details of ConvHTML are provided in App.~\ref{app:convhtml}, and ConvDeck's full implementation setup, including the LLM
backends used for each role, is described in App.~\ref{app:implementation_details}. We run ConvDeck and every baseline under three generation backbones (GPT-5~\cite{singh2025openai_gpt5}, Gemini 3 Pro~\cite{gemini3pro}, and Qwen3-VL-32B-Instruct~\cite{Qwen3-VL}),
and evaluate every resulting deck with two independent VLM judges, GPT-5 and Gemini 3 Pro; this cross-judge setup mitigates the known self-preference bias of LLM evaluators~\cite{zheng2023judging,
panickssery2024llm}. At each conversational stage, the user simulator and the corresponding refiner interact for up to five rounds, with early stopping once all assigned goals are satisfied.

\subsection{User Goal Satisfaction}
\label{sec:user_goal_satisfaction}

A central claim of ConvDeck is that multi-stage conversational refinement improves the system's ability to satisfy presenter-specific requirements. To evaluate, we design two complementary studies grounded in a shared inventory of 50 predefined user goals. Each goal expresses a concrete, verifiable requirement that a presenter may want to impose on the generated deck (\textit{e.g.,} \textit{``the deck should not include implementation details''} or \textit{``every slide title should be phrased as a claim rather than a section label''}). The 50 goals are organized into 5 categories of 10 goals each, grouped by the pipeline stage best suited to address them (see Tab.~\ref{tab:goal_inventory} and App.~\ref{app:goal_inventory} for the full inventory). We denote the outline-relevant categories as $\mathcal{C}_{\text{out}}$ (addressed during Stage~3) and the slide-relevant categories as $\mathcal{C}_{\text{sld}}$ (addressed during Stage~5). Both studies are summarized in Alg.~\ref{alg:goal_evaluation}, and use the category abbreviations from Tab.~\ref{tab:goal_inventory} for per-category results.

% ============================================================
% Tab. — Goal Inventory Summary
% ============================================================
\begin{table}[ht!]
\centering
\small
\setlength{\tabcolsep}{5pt}
\renewcommand{\arraystretch}{1.15}
\resizebox{\columnwidth}{!}{%
\begin{tabular}{@{} l l c @{}}
\toprule
\textbf{Group} & \textbf{Category} & \textbf{Abbr.} \\
\midrule
\multirow{3}{*}{\parbox{3.2cm}{Outline-relevant ($\mathcal{C}_{\text{out}}$)\\ \outlinerev{Conv.\ Outline Refinement} (Stage 3)}}
  & Content Inclusion/Exclusion & Content    \\ \cmidrule(l){2-3}
  & Narrative Structure         & Narrative  \\ \cmidrule(l){2-3}
  & Deck Composition            & Composition \\
\midrule
\multirow{2}{*}{\parbox{3.2cm}{Slide-relevant ($\mathcal{C}_{\text{sld}}$)\\ \sliderev{Conv.\ Slide Refinement} (Stage 5)}}
  & Figure/Table Usage          & Fig/Table  \\ \cmidrule(l){2-3}
  & Style \& Wording            & Style      \\
\bottomrule
\end{tabular}%
}
\caption{\textbf{Goal inventory.} Five categories of presenter-specific goals (10 goals each), grouped by the pipeline stage best suited to address them.}
\label{tab:goal_inventory}
% \vspace{-0.8em}
\end{table}

% Add to preamble if not already there:
%   \usepackage{xcolor}
%   \definecolor{algcomment}{HTML}{6A5ACD}   % SlateBlue; pick whatever you like
%   \newcommand{\alc}[1]{\textcolor{algcomment}{\textit{// #1}}}

\begin{algorithm}[t]
\caption{User-Goal Satisfaction.}
\label{alg:goal_evaluation}
\small
\begin{algorithmic}[1]
\Require Paper $P$; outline-relevant categories $\mathcal{C}_{\text{out}}$, slide-relevant categories $\mathcal{C}_{\text{sld}}$ (Sec.~\ref{sec:user_goal_satisfaction}); ConvDeck pipeline
\Statex
% \State \alc{Helper: from each category in $\mathcal{C}$, draw one goal that is applicable to $P$ and matches \text{filter} (\text{unmet} or \text{any}) on the current pipeline \text{state}.}
\Function{Sample}{$\mathcal{C}, \text{state}, \text{filter}$}
  \State \Return one goal per category in $\mathcal{C}$, drawn from those that are \textbf{applicable} to $P$ and satisfy \text{filter} on \text{state}
\EndFunction
\Statex
\State \alc{Initial outline (Stages 1--2)}
\State $O_0 \gets \textsc{Outline}(P)$
\Statex
\Statex \textbf{(a) Baseline Comparison on Unmet Goals}
\State \alc{Sample 3 outline-stage goals unmet by $O_0$}
\State $G_{\text{out}} \gets$ \Call{Sample}{$\mathcal{C}_{\text{out}}, O_0, \text{unmet}$}
\State \alc{Refine outline (Stage 3), then build draft deck (Stage 4)}
\State $D_0 \gets \textsc{Draft}(\textsc{Refine}_{\text{out}}(O_0, G_{\text{out}}))$
\State \alc{Sample 2 slide-stage goals unmet by $D_0$}
\State $G_{\text{sld}} \gets$ \Call{Sample}{$\mathcal{C}_{\text{sld}}, D_0, \text{unmet}$}
\State \alc{Refine slides with all 5 goals (Stage 5)}
\State $D_1 \gets \textsc{Refine}_{\text{sld}}(D_0,\, G_{\text{out}} \cup G_{\text{sld}})$
\State \Return satisfaction rate of $G_{\text{out}} \cup G_{\text{sld}}$ on $D_1$
\Statex
\Statex \textbf{(b) Per-Stage Goal Improvement}
\State \alc{Sample 3 outline-stage and 2 slide-stage goals upfront}
\State $G_{\text{out}} \gets$ \Call{Sample}{$\mathcal{C}_{\text{out}}, O_0, \text{any}$}
\State $G_{\text{sld}} \gets$ \Call{Sample}{$\mathcal{C}_{\text{sld}}, O_0, \text{any}$}
\State \alc{Refine outline (Stage 3), build draft deck (Stage 4)}
\State $O_1 \gets \textsc{Refine}_{\text{out}}(O_0, G_{\text{out}})$
\State $D_0 \gets \textsc{Draft}(O_1)$
\State \alc{Refine slides with all 5 goals (Stage 5)}
\State $D_1 \gets \textsc{Refine}_{\text{sld}}(D_0,\, G_{\text{out}} \cup G_{\text{sld}})$
\State \Return Before/After satisfaction of $G_{\text{out}}$ on $(O_0, O_1)$ and of $G_{\text{sld}}$ on $(D_0, D_1)$
\end{algorithmic}
\end{algorithm}

% ============================================================
% Tab. — Unmet-Goal Comparison (multi-evaluator)
% ------------------------------------------------------------
% Sub-columns per category: GPT-5, Gemini, Avg, User
%   * GPT-5 and Gemini come from cross-evaluation runs.
%   * Avg averages just the two LLM judges (GPT-5 + Gemini)/2.
%   * User Study is reported separately (1k+ responses on
%     GPT-5–generated decks).
%
% Success Rate has only GPT-5, Gemini, Avg sub-columns (no User,
% since render success is a system property, not a judge).
%
% Required color definition (preamble):
%   \definecolor{winGreen}{HTML}{B7E1BD}
% ============================================================
\begin{table*}[t!]
\centering
\small
\setlength{\tabcolsep}{2.0pt}
\renewcommand{\arraystretch}{1.10}
\resizebox{\textwidth}{!}{%
\begin{tabular}{@{} l !{\color{gray!40}\vrule}
                       cccc !{\color{gray!40}\vrule}
                       cccc !{\color{gray!40}\vrule}
                       cccc !{\color{gray!40}\vrule}
                       cccc !{\color{gray!40}\vrule}
                       cccc !{\color{gray!40}\vrule}
                       cccc !{\vrule width 1pt}
                       ccc @{}}
\toprule
\multirow{2}{*}{\textbf{Method}}
& \multicolumn{4}{c}{\textbf{Content}}
& \multicolumn{4}{c}{\textbf{Narrative}}
& \multicolumn{4}{c}{\textbf{Fig/Table}}
& \multicolumn{4}{c}{\textbf{Style}}
& \multicolumn{4}{c}{\textbf{Composition}}
& \multicolumn{4}{c}{\textbf{Overall}}
& \multicolumn{3}{c}{\textbf{Success Rate}} \\
\cmidrule(lr){2-5}\cmidrule(lr){6-9}\cmidrule(lr){10-13}\cmidrule(lr){14-17}\cmidrule(lr){18-21}\cmidrule(lr){22-25}\cmidrule(l){26-28}
\scriptsize\textit{Eval. models $\rightarrow$}
& \scriptsize\textbf{GPT-5} & \scriptsize\textbf{GMN} & \scriptsize\textit{Avg} & \scriptsize\textbf{User}
& \scriptsize\textbf{GPT-5} & \scriptsize\textbf{GMN} & \scriptsize\textit{Avg} & \scriptsize\textbf{User}
& \scriptsize\textbf{GPT-5} & \scriptsize\textbf{GMN} & \scriptsize\textit{Avg} & \scriptsize\textbf{User}
& \scriptsize\textbf{GPT-5} & \scriptsize\textbf{GMN} & \scriptsize\textit{Avg} & \scriptsize\textbf{User}
& \scriptsize\textbf{GPT-5} & \scriptsize\textbf{GMN} & \scriptsize\textit{Avg} & \scriptsize\textbf{User}
& \scriptsize\textbf{GPT-5} & \scriptsize\textbf{GMN} & \scriptsize\textit{Avg} & \scriptsize\textbf{User}
& \scriptsize\textbf{GPT-5} & \scriptsize\textbf{GMN} & \scriptsize\textit{Avg} \\
\midrule
ConvHTML
& 62\% & 68\% & 65\% & 80\%
& 68\% & 72\% & 70\% & 72\%
& 66\% & 76\% & 71\% & 70\%
& \cellcolor{winGreen}62\% & \cellcolor{winGreen}76\% & \cellcolor{winGreen}69\% & 64\%
& 40\% & 66\% & 53\% & 74\%
& 60\% & 72\% & 66\% & 72\%
& 90\% & \cellcolor{winGreen}100\% & 95\% \\

AutoSlides
& 76\% & 64\% & 70\% & 77\%
& 56\% & 68\% & 62\% & 72\%
& \cellcolor{winGreen}84\% & 72\% & 78\% & 65\%
& 42\% & 48\% & 45\% & 77\%
& \cellcolor{winGreen}98\% & 86\% & \cellcolor{winGreen}92\% & 77\%
& 71\% & 68\% & 70\% & 74\%
& 72\% & 91\% & 82\% \\

\rowcolor{gray!15}
ConvDeck
& \cellcolor{winGreen}92\% & \cellcolor{winGreen}88\% & \cellcolor{winGreen}90\% & \cellcolor{winGreen}81\%
& \cellcolor{winGreen}82\% & \cellcolor{winGreen}82\% & \cellcolor{winGreen}82\% & \cellcolor{winGreen}75\%
& 76\% & \cellcolor{winGreen}88\% & \cellcolor{winGreen}82\% & \cellcolor{winGreen}74\%
& 50\% & 66\% & 58\% & \cellcolor{winGreen}81\%
& 72\% & \cellcolor{winGreen}96\% & 84\% & \cellcolor{winGreen}78\%
& \cellcolor{winGreen}74\% & \cellcolor{winGreen}84\% & \cellcolor{winGreen}79\% & \cellcolor{winGreen}78\%
& \cellcolor{winGreen}99\% & 96\% & \cellcolor{winGreen}98\% \\
\bottomrule
\end{tabular}%
}
\caption{\textbf{Baseline comparison on unmet goals} (\%). Sub-headers list the \emph{evaluation} model. GPT-5 evaluations are conducted on Gemini-based generation and vice-versa;
%%GPT-5 and Gemini come from cross-evaluation; 
\textit{Avg} is the mean of the two LLM judges. \textbf{User} reports human ratings on GPT-5–generated decks. \textbf{Success Rate} is the fraction of papers the method rendered without failure.}
\label{tab:user_goal_satisfaction_baselines}
% \vspace{-1em}
\end{table*}

\paragraph{Study 1: Baseline comparison on unmet goals.}
This study (Alg.~\ref{alg:goal_evaluation}~(a)) targets goals that the system fails to satisfy under non-conversational generation and measures whether conversational refinement can close that gap. Goals are sampled in two waves: first, after generating the initial outline $O_0$, a VLM judge identifies the applicable goals in each outline-relevant category that are unmet by $O_0$, and we sample one such goal per category to obtain $G_{\text{out}}$ (three outline-stage goals). The \convuser{user} pursues $G_{\text{out}}$ across up to five rounds of feedback in Stage~3, producing a refined outline that is rendered into a draft deck $D_0$ at Stage~4. The same sampling procedure is then applied to the slide-relevant categories on $D_0$, yielding $G_{\text{sld}}$ (two slide-stage goals). In Stage~5, the \convuser{user} pursues the full set $G_{\text{out}} \cup G_{\text{sld}}$, so that both kinds of requirements remain in scope during slide refinement. We report the percentage of $G_{\text{out}} \cup G_{\text{sld}}$ satisfied in the final deck $D_1$, averaged over the 100-paper benchmark (Tab.~\ref{tab:user_goal_satisfaction_baselines}). The same applicability filter and sampling procedure are applied to the conversational baselines (ConvHTML, AutoSlides) for fair comparison. Judge prompts are in App.~\ref{app:eval_prompts}.

\paragraph{Stage-level conversation drives the largest goal-satisfaction gains.} ConvDeck satisfies the most initially unmet goals overall (79\% across judges), leading on the outline-relevant categories Content Inclusion/Exclusion (90\%) and Narrative Structure (82\%) because outline-level changes propagate across the deck before rendering, whereas post-hoc editors can only patch a finished one. ConvHTML wins on Style (69\%) since HTML allows direct surface edits, and AutoSlides is competitive only on countable categories (Deck Composition, Figure/Table Usage), where a final-deck editor can directly add or remove items. ConvDeck also has the highest rendering Success Rate (98\%), while AutoSlides produces unstable outputs more often.

\paragraph{User studies.} We conduct a user study on 30 papers with 30 participants, where each participant is assigned 5 randomly sampled papers and rates whether each method satisfies the sampled goals on the same unmet-goal protocol (Tab.~\ref{tab:user_goal_satisfaction_baselines}, \textbf{User} columns; see App.~\ref{app:user_study} for details and user/LLM evaluation correlation). Human ratings favor ConvDeck in every category (overall 78\%, vs.\ 72\% for ConvHTML and 74\% for AutoSlides), including Style, where the LLM judges had ranked ConvHTML highest. This indicates that LLM judges slightly underestimate ConvDeck's style quality relative to human raters, while preserving its lead on structural categories.

We conduct two additional natural-interaction user studies with five MS/PhD students. In the first study, each participant selects seven papers from our 100-paper set and interacts with ConvDeck naturally to satisfy one initially unmet goal from each category, resulting in 35 sessions. ConvDeck satisfied 168 of 175 goals (96.0\%). In the second study, each participant uploads four papers of their choice and formulates paper-specific goals for each category without relying on our predefined goal inventory, resulting in 20 sessions. ConvDeck satisfied 93 of 100 goals (93.0\%). The full results are reported in Table~\ref{tab:natural_requests_human_study}. These studies demonstrate that ConvDeck can handle both natural, potentially messy feedback on initially unmet goals and naturally formulated, paper-specific requests.

% ============================================================
% Tab. — Per-Stage Goal Improvement
% ============================================================
\begin{table}[t!]
\centering
\small
\setlength{\tabcolsep}{6pt}
\renewcommand{\arraystretch}{1.15}
\resizebox{\columnwidth}{!}{%
\begin{tabular}{@{} l ccc ccc @{}}
\toprule
\multirow{2}{*}{\textbf{Category}}
& \multicolumn{3}{c}{\textbf{Outline (Stage 3)}}
& \multicolumn{3}{c}{\textbf{Slide (Stage 5)}} \\
\cmidrule(lr){2-4}\cmidrule(l){5-7}
& \textbf{Before} & \textbf{After} & \textbf{$\Delta$}
& \textbf{Before} & \textbf{After} & \textbf{$\Delta$} \\
\midrule
% ----- Outline-relevant categories -----
Content     & 42\%   & 91\%   & \textcolor{deltaTxt}{$\uparrow$ +49\%}
            & 90\%   & 93\%   & \textcolor{deltaTxt}{$\uparrow$ +3\%}  \\
Narrative   & 18\%   & 89\%   & \textcolor{deltaTxt}{$\uparrow$ +71\%}
            & 66\%   & 72\%   & \textcolor{deltaTxt}{$\uparrow$ +6\%}  \\
Composition & 53\%   & 96\%   & \textcolor{deltaTxt}{$\uparrow$ +43\%}
            & 80\%   & 96\%   & \textcolor{deltaTxt}{$\uparrow$ +16\%} \\
\midrule
% ----- Slide-relevant categories -----
Fig/Table   & --     & --     & --
            & 54\%   & 90\%   & \textcolor{deltaTxt}{$\uparrow$ +36\%} \\
Style       & --     & --     & --
            & 52\%   & 81\%   & \textcolor{deltaTxt}{$\uparrow$ +29\%} \\
\midrule
\textit{Avg.}
            & 37.7\% & 92.0\% & \textcolor{deltaTxt}{$\uparrow$ +54.3\%}
            & 68.4\% & 86.4\% & \textcolor{deltaTxt}{$\uparrow$ +18.0\%} \\
\bottomrule
\end{tabular}%
}
\caption{\textbf{Per-stage goal improvement.} Goal satisfaction (\%) before and after refinement at each pipeline stage. $\Delta$ is the percentage-point gain.}
\label{tab:user_goal_satisfaction}
\end{table}

% ============================================================
% Table R4 — Natural Requests / Human Study
% ============================================================
\begin{table}[t!]
\centering
\small
\setlength{\tabcolsep}{5pt}
\renewcommand{\arraystretch}{1.18}

\resizebox{\columnwidth}{!}{%
\begin{tabular}{@{} l cc @{}}
\toprule
\textbf{Category}
& \textbf{Initially Unmet Goals}
& \textbf{Participant-Defined Goals} \\
\midrule
Content Inclusion/Exclusion
& 97.1\%
& 100\% \\

Narrative Structure
& 97.1\%
& 100\% \\

Deck Composition
& 97.1\%
& 95\% \\

Figure \& Table Usage
& 97.1\%
& 80\% \\

Style \& Wording
& 91.4\%
& 90\% \\

\midrule
\textbf{Average}
& \textbf{96.0\%}
& \textbf{93.0\%} \\
\bottomrule
\end{tabular}%
}

\caption{\textbf{Goal-satisfaction rates in the natural-interaction human studies.}
Results are reported for initially unmet goals across 35 sessions and participant-defined goals across 20 sessions.}
\label{tab:natural_requests_human_study}
\end{table}

% ============================================================
% Fig. — Pairwise Preference (side-by-side, full text width)
% Image is generated by pairwise_preference_fig_v3_sbs.py
% Requires in preamble:
%   \usepackage{graphicx}
% ============================================================
\begin{figure*}[t!]
\centering
\includegraphics[width=\textwidth]{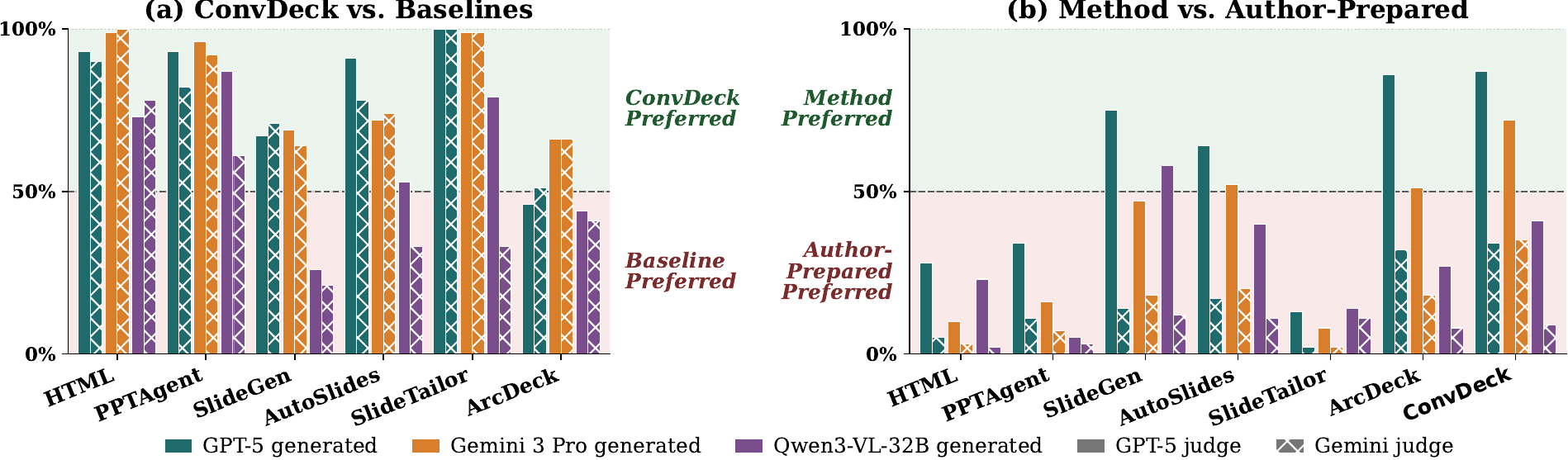}
\caption{\textbf{Pairwise preference win rate (\%) on ArcBench}, across three generation backbones (GPT-5, Gemini 3 Pro, Qwen3-VL-32B) and two judges (GPT-5, Gemini 3 Pro). Hue encodes the generation backbone; fill texture encodes the judge (solid: GPT-5 judge, dotted: Gemini judge). \textbf{(a)} Baselines vs.\ ConvDeck; \textbf{(b)} each method vs.\ Author-Prepared (AP) slides. Bars in the green half indicate the row method is preferred; the dashed line at $50\%$ is the tie threshold.}
\label{fig:vlm_pairwise_preference}
\end{figure*}

\paragraph{Study 2: Per-stage goal improvement.}
The second study reflects a more general scenario in which the user specifies requirements upfront, regardless of whether the initial system would have satisfied them. For each paper, we directly sample one goal from each applicable category, yielding three outline-stage goals and two slide-stage goals. At Stage 3, the user pursues the three outline-stage goals, and at Stage 5 the user pursues the full set of five goals (the three outline-stage goals together with the two slide-stage goals), mirroring the goal routing in the first study. We then measure the percentage of these goals satisfied (i) before any conversation, by the initial outline at Stage 2 and the draft deck at Stage 4, and (ii) after conversation, by the refined outline at Stage 3 and the final deck at Stage 5, reporting the improvement before vs.\ after. Results are reported separately for outline refinement (Stage 3) and slide refinement (Stage 5) in Tab.~\ref{tab:user_goal_satisfaction}, allowing us to attribute goal-satisfaction gains to each conversational stage independently. The full procedure is given in Alg.~\ref{alg:goal_evaluation} (b).

\paragraph{Each conversational stage substantially improves the goals it is designed to address.} Outline refinement (Stage~3) lifts outline-relevant goals from 37.7\% to 92.0\% on average ($+54.3$ points), with the largest gain on Narrative Structure ($+71$); these are structural decisions that can be changed cheaply and globally before any slide is rendered. Slide refinement (Stage~5) then lifts slide-relevant goals from 68.4\% to 86.4\% ($+18.0$), concentrated on Figure/Table Usage ($+36$) and Style \& Wording ($+29$), since these depend on the rendered output that only Stage~5 can observe and correct. Each stage improves mainly its own categories, confirming that routing feedback to the matching stage, rather than deferring everything to a final post-hoc edit, enables these gains.

% ============================================================
% Tab. — Ablation Study (unmet-goal format, 20 papers)
% ============================================================
\definecolor{lossRed}{HTML}{C0392B}
\definecolor{gainGreen}{HTML}{1F8B3D}
\begin{table}[t!]
\centering
\small
\setlength{\tabcolsep}{4pt}
\renewcommand{\arraystretch}{1.18}
\newcommand{\dlt}[1]{\,{\scriptsize\textcolor{lossRed}{$-#1$}}}
\newcommand{\dgn}[1]{\,{\scriptsize\textcolor{gainGreen}{$+#1$}}}
\resizebox{\columnwidth}{!}{%
\begin{tabular}{@{} l cccccc @{}}
\toprule
\textbf{Method}
& \textbf{Content}
& \textbf{Narrative}
& \textbf{Fig/Table}
& \textbf{Style}
& \textbf{Composition}
& \textbf{Avg.} \\
\midrule
\rowcolor{gray!15}
\textbf{ConvDeck}
& 85\% & 90\% & 75\% & 70\% & 90\% & 82\% \\
\midrule
w/o Stage~3
& 85\%               & 70\%\dlt{20}        & 85\%\dgn{10}        & 40\%\dlt{30}        & 85\%\dlt{5}         & 73\%\dlt{9}  \\
w/o Stage~5
& 100\%\dgn{15}      & 80\%\dlt{10}        & 10\%\dlt{65}        & 0\%\dlt{70}         & 65\%\dlt{25}        & 51\%\dlt{31} \\
w/o Stage~3 \& 5
& 55\%\dlt{30}       & 15\%\dlt{75}        & 5\%\dlt{70}         & 0\%\dlt{70}         & 15\%\dlt{75}        & 18\%\dlt{64} \\
\bottomrule
\end{tabular}%
}
\caption{\textbf{Ablation study} (unmet-goal protocol, 20 papers). $\Delta$ shows the change in goal satisfaction (\%) relative to \textbf{ConvDeck (full)}.}
\label{tab:ablation_study}
\end{table}

\paragraph{Ablation study.}
Removing Stage~5 (slide feedback) collapses Figure/Table Usage and Style satisfaction (75\% to 10\%, 70\% to 0\%), since these depend on the rendered deck that only Stage~5 can observe. Removing Stage~3 (outline feedback) erodes the outline-relevant categories Narrative Structure ($-20$) and Style ($-30$); Style declines here too because slide-level wording fixes in Stage~5 cannot compensate for structurally weak content laid down at Stage~2. Disabling both leaves the one-shot deck satisfying just 18\% of goals on average, a $-64$-point drop, showing that the two conversational stages address complementary failure modes. To show the results for a three-stage conversational refinement pipeline, we add an interaction point for refining the global commitment in Stage 2. We evaluate this variant against ConvDeck using the same unmet-goal protocol on 20 papers. Table~\ref{tab:ablation_commitment} shows that adding interaction at the global-commitment stage provides no improvement in goal satisfaction across the three outline-relevant categories, while increasing token usage by 4.8\% and runtime by 4.5\%. These results suggest that this additional interaction does not provide benefits beyond the existing outline- and slide-refinement stages.

\subsection{Overall Quality}
\label{sec:overall_quality}

To complement the user-goal evaluation, we assess the overall quality of the generated decks following the ArcBench~\cite{ozden2026arcdeck} evaluation protocol. We report two pairwise-preference studies in the main paper, and provide the full VLM-as-Judge rubric (text quality, narrative flow, visual layout, and visual--text alignment, on a 0--10 scale) in App.~\ref{app:vlm_as_judge}, since the rubric values broadly track the pairwise rankings reported here. A/B prompts and judge protocols are in App.~\ref{app:eval_prompts}, and qualitative side-by-side comparisons in App.~\ref{app:qualitative_examples}.

\noindent\textit{Pairwise preference vs.\ baselines} (Fig.~\ref{fig:vlm_pairwise_preference}a) shows a VLM judge two decks alongside the source paper and asks which one better captures the paper's narrative and content, reported as ConvDeck's win rate against each baseline.

\noindent\textit{Pairwise preference vs.\ author-prepared slides} (Fig.~\ref{fig:vlm_pairwise_preference}b) applies the same A/B protocol against the human-prepared reference decks in ArcBench, measuring how closely each automated method approaches expert-level presentations.

% ============================================================
% Tab. — Additional Interaction Ablation
% ============================================================
\definecolor{lossRed}{HTML}{C0392B}

\begin{table}[t!]
\centering
\small
\setlength{\tabcolsep}{5pt}
\renewcommand{\arraystretch}{1.18}
\newcommand{\dlt}[1]{\,{\scriptsize\textcolor{lossRed}{$+#1$}}}

\resizebox{\columnwidth}{!}{%
\begin{tabular}{@{} l cc @{}}
\toprule
\textbf{Metric}
& \textbf{ConvDeck}
& \textbf{ConvDeck w/ Commitment Interaction} \\
\midrule
Content
& 90\%
& 90\% \\

Composition
& 100\%
& 100\% \\

Narrative
& 80\%
& 80\% \\

\midrule
Total tokens
& 281K
& 294.5K \dlt{4.8\%} \\

Runtime
& 940 sec
& 982.5 sec \dlt{4.5\%} \\
\bottomrule
\end{tabular}%
}

\caption{\textbf{Commitment interaction ablation.}
Adding an additional conversational refinement stage at the global-commitment level yields no improvement in goal satisfaction while increasing token usage and runtime.}
\label{tab:ablation_commitment}
\end{table}

% \input{tables_tex/conv_response_quality}
% ============================================================
% Fig. — Compute footprint per paper (full text width, double column)
% Image is generated by token_cost_fig.py
% Requires in preamble:
%   \usepackage{graphicx}
% (Single-column: change figure* -> figure and \textwidth -> \columnwidth)
% ============================================================
\begin{figure*}[t!]
\centering
\includegraphics[width=\textwidth]{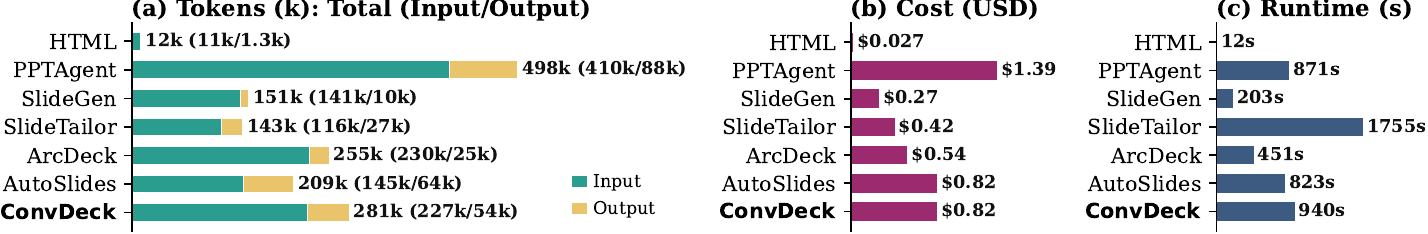}
\caption{\textbf{Average compute footprint.}
\textbf{(a)} Total tokens (k) per method; each stacked bar shows the input (teal) vs.\ output (amber) split.
\textbf{(b)} Estimated API cost in USD.
\textbf{(c)} End-to-end runtime (s).}
\label{fig:token_cost_comparison}
\end{figure*}

\paragraph{Conversational refinement preserves standard quality and brings ConvDeck closest to human-prepared decks.} On the stronger GPT-5 and Gemini~3 Pro backbones, ConvDeck is preferred over every baseline by both judges (\textit{e.g.}\ 91\% over AutoSlides on GPT-5) and is the automated method that comes closest to the author-prepared reference, reaching 87\% on GPT-5 under the GPT judge. Margins shrink on the weaker Qwen3-VL backbone, where the smaller generation model produces lower-quality edits across all conversational systems and win rates compress toward 50\%, indicating that conversational refinement amplifies a capable backbone rather than substituting for it. The drop relative to author-prepared decks in panel~(b) is consistent across all automated methods, suggesting a judge-side preference for human references rather than a ConvDeck-specific weakness.

\subsection{Conversational Response Quality}
\label{sec:conversational_response_quality}

Beyond how many goals each stage ultimately satisfies, we evaluate \emph{how} the conversation gets there. For each conversational stage $\textsc{Stage}_k$ ($k \in \{3, 5\}$) under the unmet-goal protocol, we feed the full conversation trajectory to a VLM judge and rate it on four 1--5 dimensions (full prompts in App.~\ref{app:evaluation_prompts}). \textit{Feedback Understanding} (FU) measures whether the system correctly interprets the user's feedback in each round. \textit{Feedback Responsiveness} (FR) measures whether the system actually acts on it in the subsequent revision. \textit{Cross-Round Consistency} (CRC) measures whether requests and decisions from earlier rounds are preserved as the conversation continues. \textit{Revision Stability} (RS) measures whether each round avoids regressing on aspects of the deck that were previously good. We report all four metrics round-by-round, alongside the goal-satisfaction rate at the end of each round, separately for outline refinement (Stage~3) and slide refinement (Stage~5) in Tab.~\ref{tab:roundwise_conversation_quality} (Round 0 is the initial state before any feedback).

% ============================================================
% Tab. — Round-wise goal satisfaction & conversation quality
% (transposed: metrics are rows, rounds are columns; both stages
%  side-by-side in one tabular)
% ------------------------------------------------------------
% Rows: Sat. (%) and its Δ row; FU/FR/CRC/RS sub-metrics; Avg.
% of the four sub-metrics and its Δ row. Δ is the change from the
% first available round (R0 for Sat.; R1 for conv-quality Avg).
% ============================================================
\begin{table}[t]
\centering
\small
\setlength{\tabcolsep}{4pt}
\renewcommand{\arraystretch}{1.12}
\resizebox{\columnwidth}{!}{%
\begin{tabular}{@{} l !{\color{gray!40}\vrule}
                       cccccc !{\vrule width 1pt}
                       cccccc @{}}
\toprule
\multirow{2}{*}{\textbf{Metric}}
& \multicolumn{6}{c}{\textbf{Outline generation (Stage 3)}}
& \multicolumn{6}{c}{\textbf{Slide generation (Stage 5)}} \\
\cmidrule(lr){2-7}\cmidrule(l){8-13}
& \textbf{R0} & \textbf{R1} & \textbf{R2} & \textbf{R3} & \textbf{R4} & \textbf{R5}
& \textbf{R0} & \textbf{R1} & \textbf{R2} & \textbf{R3} & \textbf{R4} & \textbf{R5} \\
\midrule
Sat.\ (\%)
& 12.3 & 72.0 & 78.0 & \textbf{83.3} & 83.0 & 82.8
& 49.0 & 77.4 & 82.4 & 82.6 & \textbf{84.2} & 82.1 \\
{\scriptsize Sat.\ $\Delta$}
& -- & {\scriptsize +59.7} & {\scriptsize +65.7} & {\scriptsize +71.0} & {\scriptsize +70.7} & {\scriptsize +70.5}
& -- & {\scriptsize +28.4} & {\scriptsize +33.4} & {\scriptsize +33.6} & {\scriptsize +35.2} & {\scriptsize +33.1} \\
\midrule
FU
& -- & 4.83 & 4.80 & 4.78 & 4.58 & 4.42
& -- & 4.61 & 4.47 & 4.51 & 4.24 & 3.74 \\
FR
& -- & 4.76 & 4.74 & 4.72 & 4.54 & 4.26
& -- & 4.23 & 4.21 & 4.19 & 4.02 & 3.41 \\
CRC
& -- & 4.93 & 4.88 & 4.90 & 4.82 & 4.83
& -- & 5.00 & 4.72 & 4.72 & 4.61 & 4.47 \\
RS
& -- & 4.89 & 4.85 & 4.89 & 4.83 & 4.76
& -- & 4.66 & 4.56 & 4.56 & 4.42 & 4.21 \\
\cmidrule(lr){2-7}\cmidrule(l){8-13}
\textit{Avg.}
& -- & 4.85 & 4.82 & 4.82 & 4.69 & 4.57
& -- & 4.63 & 4.49 & 4.50 & 4.32 & 3.96 \\
{\scriptsize Avg.\ $\Delta$}
& -- & -- & {\scriptsize $-0.03$} & {\scriptsize $-0.03$} & {\scriptsize $-0.16$} & {\scriptsize $-0.28$}
& -- & -- & {\scriptsize $-0.14$} & {\scriptsize $-0.13$} & {\scriptsize $-0.31$} & {\scriptsize $-0.67$} \\
\bottomrule
\end{tabular}%
}
% \vspace{2pt}
% \footnotesize
% \textbf{FU}~feedback understanding; \textbf{FR}~feedback responsiveness;
% \textbf{CRC}~cross-round consistency; \textbf{RS}~revision stability.
\caption{\textbf{Round-wise goal satisfaction and conversation quality}. \textbf{Sat.}\ = \% of initially-unmet goals satisfied after round $R$. \textbf{FU/FR/CRC/RS} are conversation-quality sub-metrics on a 1--5 scale (defined below); \textbf{Avg.}\ averages the four. $\Delta$ rows are the change vs.\ the first available round; peak Sat.\ per stage is bolded.}
\label{tab:roundwise_conversation_quality}
\end{table}

\paragraph{Three rounds recover most of the goal gap.} Goal satisfaction climbs sharply over the first three rounds and then plateaus: outline 12.3\%~$\to$~83.3\%, slide 49.0\%~$\to$~82.6\%. The four conversation-quality metrics tell a complementary story. FU and FR start near-ceiling (outline 4.83/4.76 at round 1) and stay high through round 3, then decline by rounds 4--5 (outline FU 4.42, slide FU 3.74), suggesting that once the easy revisions are exhausted, the remaining feedback is harder to parse and act on. CRC and RS stay above 4.4 throughout, so earlier requests are preserved and the system rarely regresses even in long conversations.

\subsection{Compute Footprint}
\label{sec:runtime_cost}

\paragraph{ConvDeck adds only modest cost due to multi-turn conversations.} We report per-paper token usage, API cost, and end-to-end runtime in Fig.~\ref{fig:token_cost_comparison}, with a per-stage breakdown in App.~\ref{app:runtime_breakdown}. At \$0.82 per paper (281k tokens, 227k input / 54k output), ConvDeck sits in the mid-range, well below the most expensive baseline (PPTAgent, \$1.39) and on par with the strongest conversational baseline (AutoSlides). Its high token total does not translate into high cost because the budget is dominated by \emph{input} tokens---re-reading the paper, outline, and rendered previews each turn rather than generating new text. The overhead over ArcDeck, its non-conversational backbone, is only $\sim$10\% because each refiner applies localized edits through dedicated editing functions, so a round of conversation costs a fraction of the initial generation. In wall-clock time, ConvDeck takes 940 s per paper, versus 451 s for ArcDeck and 12–871 s for the baselines. This additional interactivity cost is modest relative to the gains in goal satisfaction and overall quality.

\section{Conclusion}
\label{sec:conclusion}

We introduce ConvDeck, a multi-agent paper-to-slide generation pipeline that distributes conversation across two stage-specific refinement loops: one over the outline before any slides are rendered, and another over the rendered deck. Our experiments show that this stage-aligned design achieves higher user-goal satisfaction than post-hoc editors can structurally reach, while it remains competitive on slide quality metrics, and remains affordable.

%\clearpage
%\newpage

\section{Limitations}
\label{sec:limitations}

ConvDeck has several limitations. We use an LLM-based user simulator throughout our experiments for reproducibility, and complement it with a human study (App.~\ref{app:user_study}) only for evaluation; however, larger human studies remain as future work to verify that simulated feedback matches how a wider range of presenters phrases and prioritizes requests. Multi-turn conversational refinement adds computational overhead relative to single-pass generation (Fig.~\ref{fig:token_cost_comparison}). Visual refinement is fragile: the VLM-based refiner occasionally emits inaccurate coordinates when repositioning or resizing figures, leading to layout failures (see App.~\ref{app:failure_cases} for representative cases). Finally, generation quality drops on the open-source Qwen3-VL-32B backbone, since smaller open-source models have less capacity for the structured editing operations ConvDeck relies on; results on this backbone should be interpreted as a lower bound, and we leave scaling to larger open-source backbones as future work.

Moreover, ConvDeck is intended as a research prototype for assisting presenters in drafting academic slide decks, not as a replacement for expert review. Because the system summarizes papers, retrieves external information during refinement, and edits slide content through LLM/VLM agents, it may introduce factual errors, omit important caveats, or overstate claims from the source paper. We therefore recommend that generated decks be manually verified before presentation or public release. The web and arXiv retrieval tools may also introduce information that is not present in the original paper; retrieved content should be checked against the cited sources.

\section*{Acknowledgments}
Portions of this work were supported in part by the Health Care Engineering Systems Center in the UIUC Grainger College of Engineering, and by a grant from the UIUC Institute for Growth.

\bibliography{main}
\clearpage

\appendix

\appendix

\begingroup
\renewcommand{\contentsname}{Appendix Contents}
\setcounter{tocdepth}{2}
\startcontents[appendix]
\printcontents[appendix]{}{0}{\section*{\contentsname}\vspace{0.5em}}
\endgroup

\vspace{1em}

% ============================================================
% A. Supplementary Material Overview
% ============================================================

\section*{Contributions}

T.C.O., S.VS., O.K., J.K., D.H.-T., and J.M.R. participated in the design of the study, T.C.O., S.VS., and F.H. implemented the system, T.C.O., S.VS., and F.H. conducted the experiments, T.C.O., S.VS., F.H., and O.K. wrote the paper. O.K. and J.K. led the project, with D.H.-T. and J.M.R. providing supervision and guidance.

\section{Supplementary Material Overview}
\label{app:overview}
Please open index.html from supplementary to view the full PDF results. This supplementary material is organized as follows.
App.~\ref{app:implementation_details} reports the full implementation of ConvDeck and the evaluation infrastructure (generation backbones, the conversational user simulator, the evaluation judges, and tooling).
App.~\ref{app:arcdeck_details}--\ref{app:editing_functions} detail the method: the narrative-driven outline and slide-generation backbone reused from ArcDeck~\cite{ozden2026arcdeck} (App.~\ref{app:arcdeck_details}), the speak--act refinement mechanism (App.~\ref{app:respact_details}), and the editing functions exposed to the Stage~3 and Stage~5 refiners (App.~\ref{app:editing_functions}).
App.~\ref{app:convhtml} describes the ConvHTML conversational baseline.
App.~\ref{app:arcbench}--\ref{app:overall_quality_eval} cover evaluation: the ArcBench benchmark (App.~\ref{app:arcbench}), the 50-goal inventory (App.~\ref{app:goal_inventory}), the conversational user simulator (App.~\ref{app:user_simulator}), and the overall-quality protocol (App.~\ref{app:overall_quality_eval}).
App.~\ref{app:judge_analysis} presents a dedicated analysis of judge reliability and bias.
App.~\ref{app:runtime_breakdown} gives a per-stage token breakdown, App.~\ref{app:qualitative_examples} shows qualitative examples, and App.~\ref{app:prompts} lists the verbatim prompts.

% ============================================================
% B. Implementation Details
% ============================================================
% ============================================================
% Appendix: Implementation Details
% Generation-pipeline details derived from ArcDeck; backbones and
% counts adapted to ConvDeck. Replaces the former inline section.
% ================
\section{Implementation Details}
\label{app:implementation_details}

This appendix details the models and infrastructure behind ConvDeck and its evaluation. The conversational user simulator is described in App.~\ref{app:user_simulator}, and the judge models and prompts in App.~\ref{app:overall_quality_eval} and App.~\ref{app:eval_prompts}.

\paragraph{Generation backbones.}
We run ConvDeck and all baselines with three interchangeable backbones: GPT-5~\cite{singh2025openai_gpt5}, Gemini 3 Pro~\cite{gemini3pro}, and Qwen3-VL-32B-Instruct~\cite{Qwen3-VL}. A single backbone drives every ConvDeck agent across Stages 2 through 5, including the Outline Generation agents, the Outline Refiner, the Slide Generation agents, and the Slide Refiner.

\paragraph{Preprocessing.}
Stage~1 parses the source PDF with Docling~\cite{livathinos2025docling}, converting the body text to a clean markdown representation and extracting figures and tables into an asset dictionary. Each visual asset is stored alongside its caption and its size (width, height, and aspect ratio) so that downstream agents have the layout context needed for figure selection and layout choice. In parallel, in-text citations are parsed into a mapping from short-form citations to their full references, which Stage~4 uses to render footnote citations. To keep the pipeline token-efficient, the references section and all content after it (e.g., the appendix) are removed from the markdown before generation.

\paragraph{Conversational refinement configuration.}
Both conversational stages run the speak--act loop (App.~\ref{app:respact_details}): the user simulator (App.~\ref{app:user_simulator}) reviews the current outline (Stage~3) or rendered deck (Stage~5) and either returns feedback or emits ``Ready'', and the corresponding refiner applies localized edits through the editing functions of App.~\ref{app:editing_functions}. The simulator runs for up to 5 rounds with early stopping once all assigned goals are satisfied, and each refiner is given an action budget per round to encourage efficient, high-impact edits rather than long trajectories. 

\paragraph{Rendering pipeline.}
All decks use a standard $16{:}9$ format with dimensions $13.33\,\text{in} \times 7.5\,\text{in}$. The Slide Deck Constructor emits a structured JSON slide specification that selects, per slide, one of 14 reusable layout templates covering common combinations of text, figures, and tables. Rather than manipulating the interdependent XML documents inside a PPTX file directly, ConvDeck translates the slide specification into JavaScript that is compiled to PPTX with PPTXGenJS~\cite{gitbrent2024pptxgenjs}. This rendered program is a structured, editable view of every slide element, and it serves as the editable state that the Slide Refiner (Stage~5) operates on during Conversational Slide Refinement. Emphasis markup produced by the Aesthetic Refiner (boldface, theme color) is parsed and converted into the corresponding PPTXGenJS formatting calls.

\paragraph{Compute and cost.}
Per-stage token and runtime figures are given in App.~\ref{app:runtime_breakdown}.

\paragraph{Evaluation fairness.} To ensure a fair evaluation, conversational baselines use the same models, feedback-round limits, and rendered information as ConvDeck. Retrieval and intermediate-representation-based editing are ConvDeck contributions and are therefore unavailable to baselines.

\paragraph{Artifact licenses and terms of use.}
ArcDeck and ArcBench are released under the MIT License. We reuse the ArcDeck backbone for ConvDeck's outline generation and slide generation stages, and use ArcBench only for research evaluation of paper-to-slide generation systems. Docling and PPTXGenJS are also MIT-licensed software artifacts. Qwen3-VL-32B-Instruct is released under the Apache 2.0 License. GPT-5 and Gemini 3 Pro are accessed through their respective commercial APIs and are used according to the applicable OpenAI and Google/Gemini API terms of service. For all baseline systems, datasets, and external tools, we follow the corresponding licenses or API terms. Generated slide decks may contain content or visual assets derived from the source papers and author-prepared slides; therefore, users are responsible for ensuring that any public reuse of generated decks is consistent with the copyright and license terms of the original materials. 

% ============================================================
% C. Method Details (shared backbone + ConvDeck-specific mechanisms)
% ============================================================

\usetikzlibrary{positioning, arrows.meta, calc, shapes.geometric, fit, backgrounds}

\section{User Study Details}
\label{app:user_study}
To evaluate the reliability of our goal-satisfaction judges, we conduct a user study in which participants are asked to assess whether the final revised slides satisfy the selected unmet goals. We sample 30 papers across all three baselines. For each participant, we randomly select 5 papers and present the corresponding slides along with the unmet user goal used during refinement. Participants are then asked to indicate whether the goal has been satisfied by the final slides for each baseline.

\begin{figure}[t]
\centering
\includegraphics[width=\columnwidth]{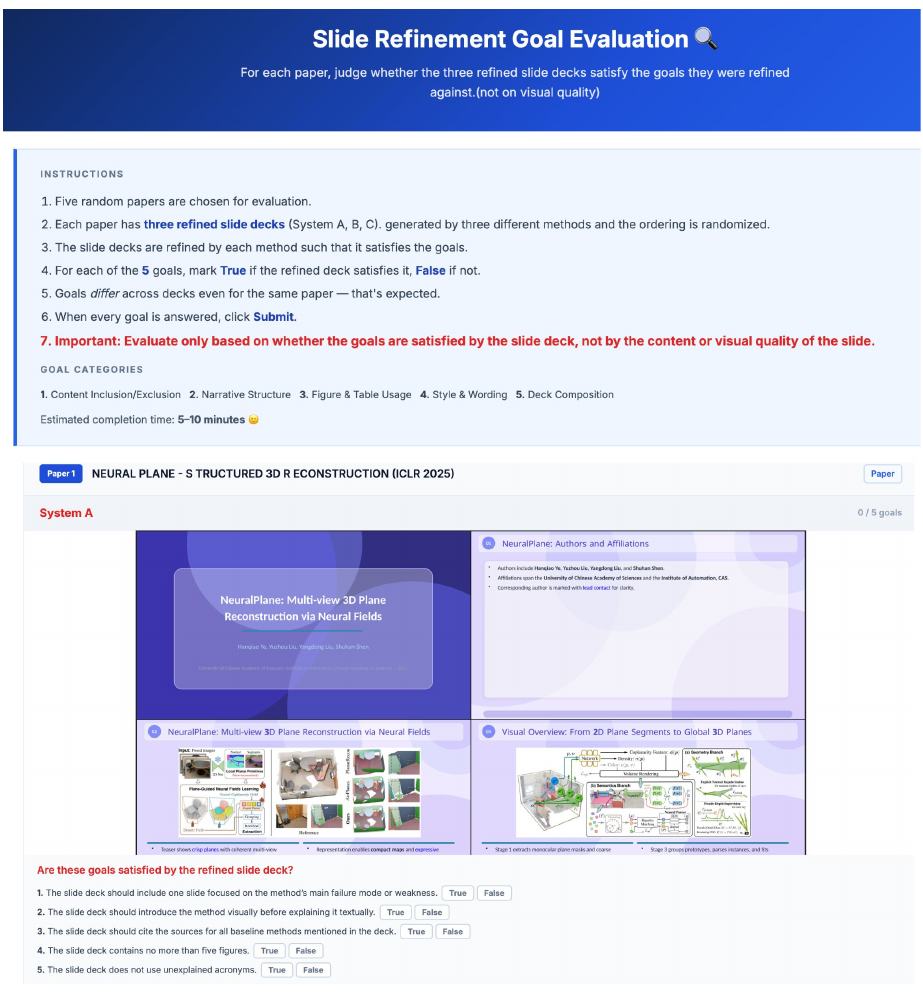}
\caption{\textbf{User study form}}
\label{fig:user_study_app}
\end{figure}

\begin{figure}[t]
\centering
\includegraphics[width=\columnwidth]{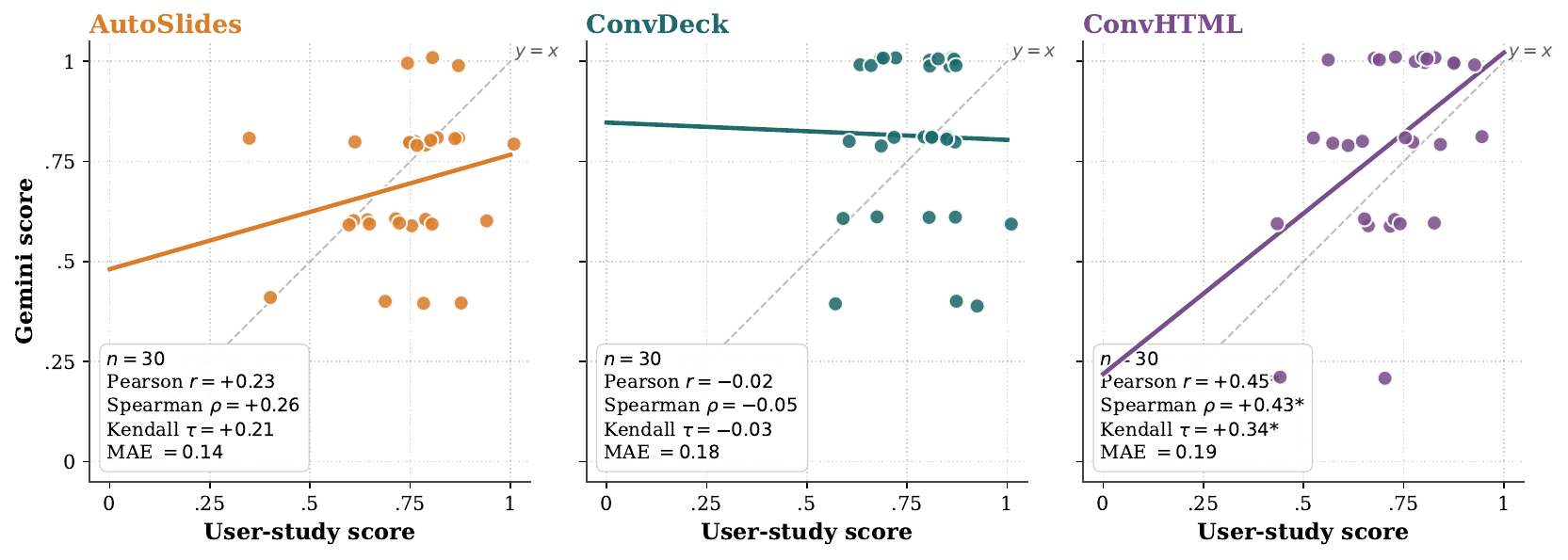}
\caption{\textbf{Gemini vs Human goal satisfaction evaluation correlation}}
\label{fig:gemini_user_corr}
\end{figure}

The study interface is shown in Fig.~\ref{fig:user_study_app}. For each sample, we also provide a link to the original paper to support more informed evaluation. The order of papers is randomized to mitigate ordering bias. We recruited 30 participants via Prolific to perform the evaluation. We report per-category goal satisfaction rates in Tab.~\ref{tab:user_goal_satisfaction_baselines}. Additionally, we compute the correlation between the average per-paper user ratings and Gemini-based automatic evaluations over the same 30 samples, as shown in Fig.~\ref{fig:gemini_user_corr}.

\begin{figure}[t]
\centering
\includegraphics[width=\columnwidth]{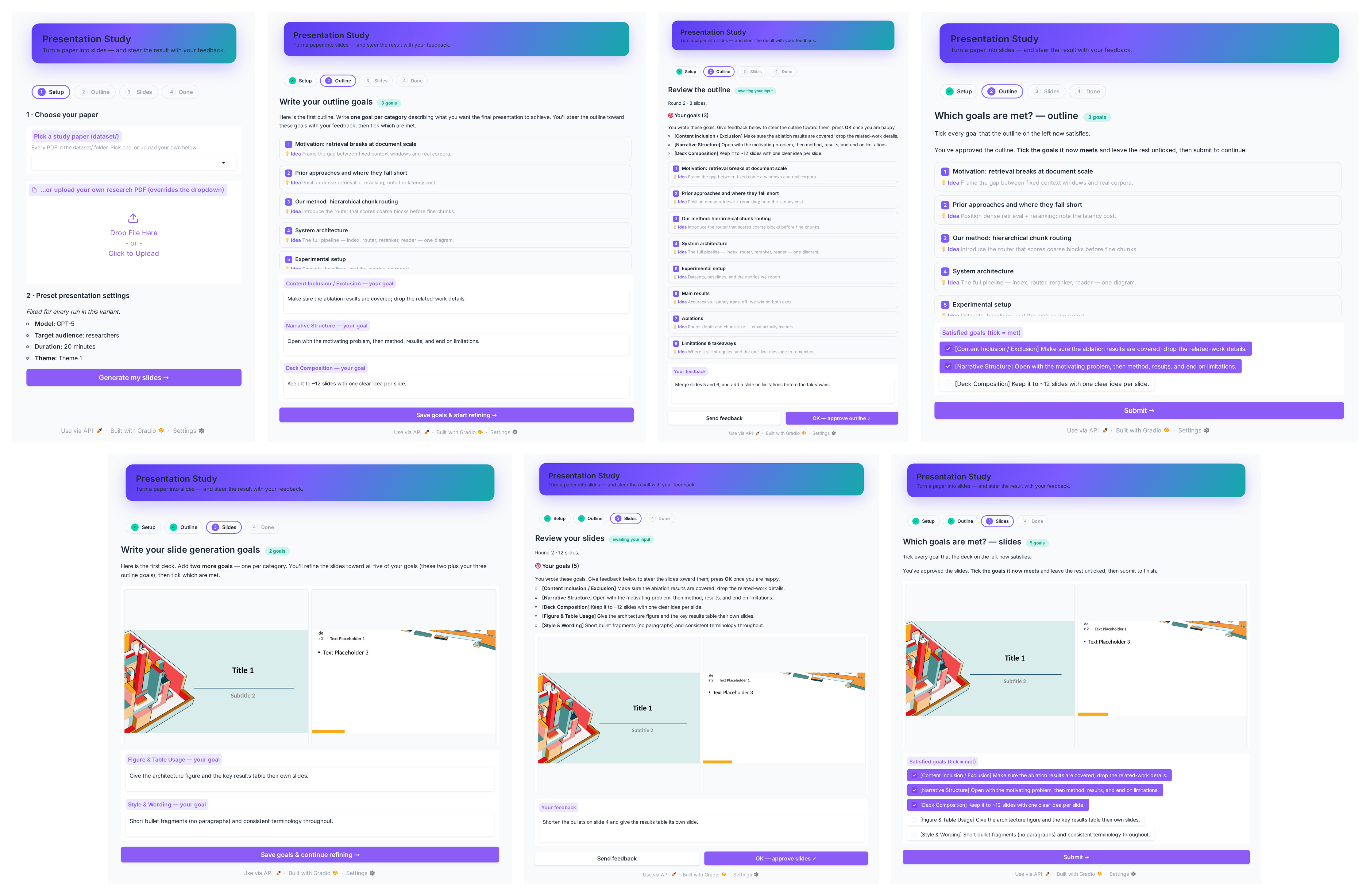}
\caption{\textbf{Natural-interaction user study interface.}}
\label{fig:natural_user_study_app}
\end{figure}

In the studies of natural-interaction user studies, participants interact naturally with ConvDeck using a GPT-5 backbone. The study interface is shown in Fig.~\ref{fig:natural_user_study_app}.

Before beginning the tasks, participants were informed that their responses would be used for research evaluation of slide-generation systems, participation was voluntary, and that their responses would be reported only in aggregate. Participants were asked to judge whether generated slides satisfied specified user goals; they were not asked to provide sensitive personal information. The study was conducted anonymously: we did not collect names, email addresses, or other direct identifiers, and participant responses were analyzed only in aggregate.

\section{ArcDeck Outline and Slide Generation Details}
\label{app:arcdeck_details}

ConvDeck reuses the \emph{non-conversational} backbone of ArcDeck~\cite{ozden2026arcdeck} for two of its five stages: \outlinegen{Outline Generation} (Stage~2) and \slidegen{Slide Generation} (Stage~4). Both stages are kept unchanged, so that any difference in user-goal satisfaction can be attributed to the conversational refinement stages (Stages~3 and~5) rather than to a stronger generator. For completeness, this appendix expands the condensed descriptions given in Sec.~\ref{sec:method_stages}. The only deviations from the original ArcDeck design are (i) the narrative refinement loop is run a single pass instead of iterating to convergence, since fine-grained outline revision is deferred to the \outlinerev{Conversational Outline Refinement} stage (Stage~3), and (ii) the final deck is rendered through a PPTXGenJS JavaScript-to-PPTX pipeline rather than \texttt{python-pptx}, so that the rendered slide program forms an editable state for the \sliderev{Conversational Slide Refinement} stage (Stage~5).

\subsection{Stage 2: Narrative-Driven Outline Generation}
\label{app:arcdeck_outline}

The Outline Generation stage turns the markdown produced by Preprocessing (Stage~1) into a logically ordered slide outline. It is driven by three components that consume the markdown in sequence: a \textbf{Discourse Parser} that exposes the rhetorical structure of the paper, a \textbf{Commitment Builder} that fixes the high-level intent of the deck, and a \textbf{Narrative Refinement Loop} that drafts and polishes the outline under both signals.

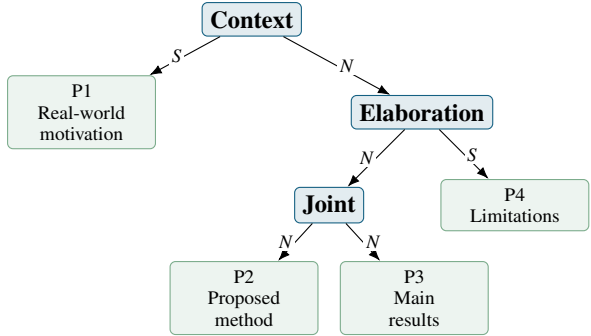
\begin{figure}[H]
\centering
\resizebox{\columnwidth}{!}{%
\begin{tikzpicture}[
  level distance=12mm,
  every node/.style={font=\footnotesize},
  edu/.style={draw=outlinecolor!60, fill=outlinecolor!8, rounded corners=2pt,
              align=center, inner sep=3pt, text width=17mm, font=\scriptsize},
  rel/.style={draw=agentblue, fill=agentblue!12, rounded corners=2pt,
              align=center, inner sep=3pt, font=\footnotesize\bfseries},
  ns/.style={font=\scriptsize\itshape, fill=white, inner sep=1pt},
  level 1/.style={sibling distance=44mm},
  level 2/.style={sibling distance=24mm},
  level 3/.style={sibling distance=22mm},
  edge from parent/.style={draw, -{Latex[length=1.6mm]}},
]
\node[rel]{Context}
  child {node[edu]{P1\\Real-world\\motivation}
         edge from parent node[ns,left]{S}}
  child {node[rel]{Elaboration}
     child {node[rel]{Joint}
        child {node[edu]{P2\\Proposed\\method}
               edge from parent node[ns,left]{N}}
        child {node[edu]{P3\\Main\\results}
               edge from parent node[ns,right]{N}}
        edge from parent node[ns,left]{N}
     }
     child {node[edu]{P4\\Limitations}
            edge from parent node[ns,right]{S}}
     edge from parent node[ns,right]{N}
  };
\end{tikzpicture}%
}
\caption{\textbf{Example discourse tree} for a four-paragraph section. Leaves are EDUs (paragraphs); internal nodes are rhetorical relations from Tab.~\ref{tab:rst_relations}; edge labels mark the nucleus (N) and satellite (S) of each NS relation. The planner co-locates tightly bound spans (e.g., the \textit{Joint} group of P2 and P3) and separates satellites such as \textit{Limitations} onto their own slides.}
\label{fig:rst_tree}
\end{figure}

\paragraph{Discourse Parser.}
The Discourse Parser is a structural-analysis agent that builds a hierarchical discourse tree, exposing the rhetorical dependencies the planner needs for content grouping and narrative ordering. For each section, it treats paragraphs as elementary discourse units (EDUs) that become the leaves of a binary tree. Following Rhetorical Structure Theory~\cite{mann1987rst}, adjacent units are linked by either (i) a \emph{nucleus-satellite} (NS) relation, where the nucleus carries the central claim and the satellite provides supporting detail, or (ii) a \emph{multinuclear} (MN) relation, where both units are equally central. The relation taxonomy is given in Tab.~\ref{tab:rst_relations}. Higher levels of the tree recursively group spans by the relations found below them, so relations near the root capture high-level rhetorical structure while relations near the leaves reflect fine-grained progression within a subsection (Fig.~\ref{fig:rst_tree}). The trees are serialized to JSON and passed to the Slide Planner.

% ---- RST relation taxonomy (ported from ArcDeck) ----
\begin{table}[H]
\centering
\small
\setlength{\tabcolsep}{5pt}
\renewcommand{\arraystretch}{1.1}
\resizebox{\columnwidth}{!}{%
\begin{tabular}{@{}llc@{}}
\toprule
\textbf{Relation} & \textbf{Explanation} & \textbf{Type} \\
\midrule
Elaboration  & Adds detail, examples, or implementation specifics.            & NS \\
Explanation  & Clarifies why or how a nucleus claim holds.                    & NS \\
Context      & Background or definitions needed to interpret the nucleus.     & NS \\
Purpose      & Goal or motivation associated with the nucleus.               & NS \\
Evaluation   & Assessment, strength, or limitation of the nucleus.           & NS \\
Organization & Roadmap or meta-structural framing.                          & NS \\
\midrule
Joint        & Parallel units at the same level, equally central.            & MN \\
Same-unit    & Two EDUs forming one semantic unit split across boundaries.    & MN \\
\bottomrule
\end{tabular}%
}
\caption{\textbf{RST relation taxonomy.} Rhetorical relations used to construct the discourse trees (NS: Nucleus-Satellite; MN: Multinuclear).}
\label{tab:rst_relations}
\end{table}
% ---- Example discourse tree (TikZ, self-contained) ----

\paragraph{Commitment Builder.}
The Commitment Builder consumes the markdown together with the two optional user inputs, target audience and presentation duration, and emits a \emph{Global Commitment}: a compact specification of the deck's high-level intent that conditions every downstream agent. As summarized in Fig.~\ref{fig:global_commitment}, the commitment has five fields, a \textit{snapshot}, the \textit{core content} (thesis and key takeaways), a \textit{talk contract} (assumed prerequisites), a \textit{narrative spine}, and a light \textit{section plan}. Audience and duration enter here: they set the assumed prerequisites and the level of detail, letting the same paper yield decks of different lengths and technical depth.

% ---- Global Commitment schematic (tcolorbox, self-contained) ----
\begin{figure}[H]
\centering
\begin{tcolorbox}[
  enhanced, breakable,
  colback=outlinecolor!4, colframe=outlinecolor!55!black,
  colbacktitle=outlinecolor!70!black, coltitle=white,
  title={\textbf{Global Commitment} \hfill (conditioned on audience + duration)},
  fonttitle=\small\bfseries, fontupper=\footnotesize,
  boxrule=0.5pt, arc=2pt, left=5pt, right=5pt, top=3pt, bottom=3pt
]
\begin{description}[leftmargin=!, labelwidth=2.1cm, itemsep=2pt, topsep=2pt, font=\bfseries\footnotesize]
  \item[Snapshot] One-line description of the talk's premise and contribution.
  \item[Core Content] The central thesis and the key takeaways the deck must convey.
  \item[Talk Contract] Prerequisites assumed of the audience; sets technical depth.
  \item[Narrative Spine] The ordered argument the slides should follow end to end.
  \item[Section Plan] A lightweight allocation of sections and their relative emphasis.
\end{description}
\end{tcolorbox}
\caption{\textbf{Structure of the Global Commitment.} The Commitment Builder fixes these five fields before any outline is drafted; they serve as a shared, high-level contract that the Slide Planner, Narrative Critic, and Narrative Judge all reference.}
\label{fig:global_commitment}
\end{figure}

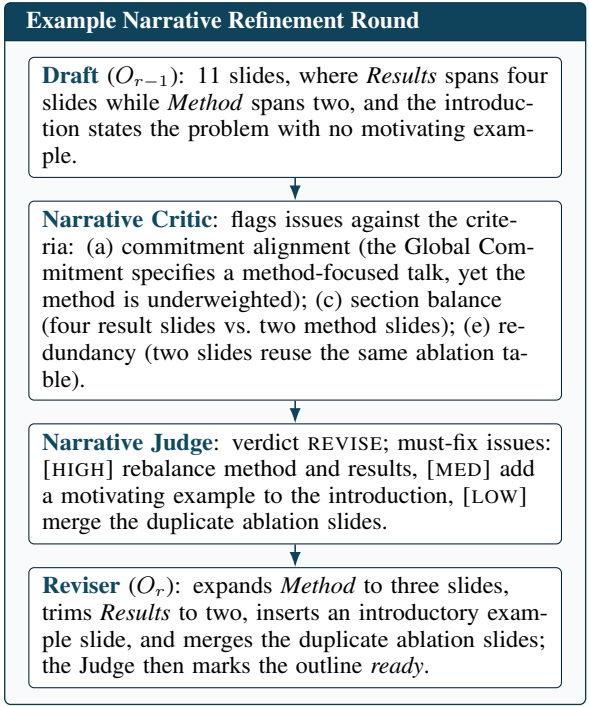
\begin{figure}[t]
  \centering
  \begin{tcolorbox}[  
    enhanced,
    colback=agentblue!3, colframe=agentblue!55!black,
    colbacktitle=agentblue!70!black, coltitle=white,
    title={\textbf{Example Narrative Refinement Round}},
    fonttitle=\small\bfseries, fontupper=\footnotesize,
    boxrule=0.5pt, arc=2pt, left=5pt, right=5pt, top=3pt, bottom=3pt
  ]
  \centering
  \begin{tikzpicture}[
    node distance=3mm,
    font=\footnotesize,
    card/.style={draw=agentblue!40!black, fill=white, rounded corners=2pt, align=left,
                 inner xsep=5pt, inner ysep=3.5pt, text width={\dimexpr\linewidth-12pt\relax}},
    flow/.style={-{Latex[length=2mm]}, draw=agentblue!70!black, line width=0.6pt},
  ]
  \node[card] (draft) {\textbf{\textcolor{agentblue!75!black}{Draft}} ($O_{r-1}$): 11 slides, where \emph{Results} spans four slides while \emph{Method} spans two, and the introduction states the problem
  with no motivating example.};
  \node[card, below=of draft] (critic) {\textbf{\textcolor{agentblue!75!black}{Narrative Critic}}: flags issues against the criteria: (a)~commitment alignment (the Global Commitment specifies a
  method-focused talk, yet the method is underweighted); (c)~section balance (four result slides vs.\ two method slides); (e)~redundancy (two slides reuse the same ablation table).};
  \node[card, below=of critic] (judge) {\textbf{\textcolor{agentblue!75!black}{Narrative Judge}}: verdict \textsc{revise}; must-fix issues: \textsc{[high]} rebalance method and results, \textsc{[med]} add
   a motivating example to the introduction, \textsc{[low]} merge the duplicate ablation slides.};
  \node[card, below=of judge] (reviser) {\textbf{\textcolor{agentblue!75!black}{Reviser}} ($O_r$): expands \emph{Method} to three slides, trims \emph{Results} to two, inserts an introductory example
  slide, and merges the duplicate ablation slides; the Judge then marks the outline \textit{ready}.};
  \draw[flow] (draft) -- (critic); 
  \draw[flow] (critic) -- (judge);
  \draw[flow] (judge) -- (reviser);
  \end{tikzpicture}
  \end{tcolorbox}
  \caption{\textbf{Example narrative refinement round.} One pass of the critique-judge-revise cycle: the Narrative Critic assesses the draft outline against the five criteria, the Narrative Judge issues a
   verdict with severity-tagged must-fix issues, and the Reviser applies localized edits before the outline re-enters the cycle. In ConvDeck this loop is run once, with finer outline revision deferred to
  Stage~3.} 
  \label{fig:refinement_loop}
  \end{figure}

\paragraph{Narrative Refinement Loop.}
Guided by the discourse tree and the Global Commitment, the Narrative Refinement Loop drafts and then polishes the outline through three agents, a \textit{Slide Planner/Reviser}, a \textit{Narrative Critic}, and a \textit{Narrative Judge} (Fig.~\ref{fig:refinement_loop}). The Slide Planner first produces a draft outline: for each section it groups preprocessed paragraphs by their rhetorical relations to decide what belongs together, yielding a JSON outline in which every slide records a title, the IDs of its assigned paragraphs, and a short rationale for the grouping. Because one-shot planning need not respect the Global Commitment or read as a coherent talk, the draft then enters a critique-judge-revise cycle assessed along five criteria: (a) alignment with the Global Commitment, (b) global narrative flow, (c) section balance, (d) slide-level coherence, and (e) redundancy or missing content. The Narrative Critic produces feedback against these criteria; the Narrative Judge decides whether the outline is \textit{ready} or needs revision, and, when revision is needed, summarizes the rationale and lists must-fix issues tagged with severity (high/medium/low). The Reviser applies the requested edits and re-enters the cycle. In ArcDeck this repeats until the Judge returns \textit{ready} or three cycles elapse; in ConvDeck we run a single pass and defer finer outline revision to Stage~3.

\subsection{Stage 4: Slide Generation}
\label{app:arcdeck_slidegen}

The Slide Generation stage renders the refined outline into a draft deck through two agents: a \textbf{Slide Deck Constructor} that turns each outline slide into a concrete, asset-grounded slide specification, and an \textbf{Aesthetic Refiner} that polishes that specification.

\paragraph{Slide Deck Constructor.}
The Constructor combines the outline, the preprocessed asset dictionary, and the Global Commitment into a draft deck. For each slide it (i) selects the most relevant figures and tables by matching slide content against asset captions; (ii) assigns one of 14 reusable layout templates, conditioned on the text volume and on the number, size, and aspect ratio of the matched visuals; and (iii) generates the slide text, toggling between bulleted hierarchies and short paragraphs according to information density and emphasizing the key points implied by the outline. Short-form citations mentioned in the assigned paragraphs are recorded so they can be rendered as footnotes. The output is a structured JSON specification carrying global metadata together with per-slide titles, text, matched visuals, and references.

% ---- Theme examples (external PDF supplied in appendix_figures/) ----
\begin{figure}[t]
\centering
\includegraphics[width=\columnwidth]{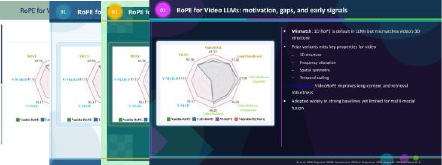}
\caption{\textbf{Theme flexibility.} The same generated deck rendered under four visual themes. Because ConvDeck fixes slide layout and content independently of styling, the theme can be changed without altering the deck's structure.}
\label{fig:layouts}
\end{figure}

\paragraph{Aesthetic Refiner.}
The Aesthetic Refiner takes the asset-matching information, the outline, and the draft specification and performs a final polish through four targeted operations: (i) \emph{Figure Matching} adds visual elements to slides that lack sufficient visual grounding; (ii) \emph{Content Refinement} balances textual density by enriching sparse slides and condensing overloaded ones; (iii) \emph{Text Coloring} applies a red and blue text color for words which conveys negative and positive things respectively; and (iv) \emph{Bold Formatting} highlights key terminology. Emphasis is expressed by wrapping spans in lightweight markup that the renderer converts into formatting instructions. The refined JSON is the specification that Stage~5 compiles, via PPTXGenJS, into the rendered slide program used as the editable state during Conversational Slide Refinement. Because layout and content are fixed independently of styling, the same deck can be rendered under different visual themes (Fig.~\ref{fig:layouts}).
\section{Speak--Act Refinement Mechanism Details}
\label{app:respact_details}

Feedback from the \convuser{conversational user} is often underspecified, forcing the refinement agent to infer the intended target or scope and to produce speculative edits. To keep edits grounded in user intent, we adopt the \emph{speak--act refinement mechanism}, inspired by ReSpAct~\cite{dongre2024respactharmonizingreasoningspeaking}, which extends the reason-act loop with a speaking action so the agent can clarify before modifying the deck. Because each interaction carries a large observation, the outline (Stage~3) or rendered deck (Stage~5) together with the user's feedback, we restrict the agent at every turn to two modes: \textit{Think+Speak}, where it reasons over the feedback and, when information is missing or ambiguous, asks the \convuser{conversational user} a single clarifying question; and \textit{Think+Act}, where it reasons over sufficiently grounded feedback and applies the appropriate edits to the affected slides. The same mechanism is instantiated by the \textit{Outline Refiner} in Stage~3 and the \textit{Slide Refiner} in Stage~5, differing only in their editing operations.

\section{ConvDeck Editing Functions}
\label{app:editing_functions}

To reduce output tokens, the refinement agents apply only local edits: they invoke predefined functions that patch the targeted slides rather than regenerating the deck, much as code agents edit files in place. Each function's signature and usage are provided to the agent at refinement time. The Stage~3 \textit{Outline Refiner} edits the outline (Tab.~\ref{tab:stage3_funcs}), while the Stage~5 \textit{Slide Refiner} edits the rendered slide specification and, since it must also fix layout and visual defects, exposes a larger set (Tab.~\ref{tab:stage5_funcs}); all functions address a slide by title or index.

\begin{table*}[t]
\centering\small
\begin{tabularx}{\textwidth}{@{}lXlX@{}}
\toprule
\textbf{Function} & \textbf{Effect} & \textbf{Function} & \textbf{Effect} \\
\midrule
\texttt{add\_slide}    & Insert a slide (non-empty title, content, discussion idea) at an optional index. & \texttt{move\_slide}   & Relocate one slide before/after another or to an index. \\
\texttt{remove\_slide} & Delete the addressed slide. & \texttt{reorder}       & Full reshuffle; every slide listed exactly once. \\
\texttt{edit\_slide}   & Overwrite a slide's content or discussion idea (title selects only). & \texttt{split\_slide}  & Replace a slide with two or more parts. \\
\texttt{retitle}       & Rename a slide. & \texttt{merge\_slides} & Merge targets at the earliest position; content auto-concatenated if omitted. \\
\bottomrule
\end{tabularx}
\caption{Stage~3 Outline Refiner editing functions.}
\label{tab:stage3_funcs}
\end{table*}

\begin{table*}[t]
\centering\small
\begin{tabularx}{\textwidth}{@{}lXlX@{}}
\toprule
\textbf{Function} & \textbf{Effect} & \textbf{Function} & \textbf{Effect} \\
\midrule
\texttt{set\_bullets} / \texttt{set\_paragraph} & Replace bullets or paragraph (mutually exclusive); on \texttt{T14}, target a named column. & \texttt{reorder} & Full reshuffle; every title exactly once. \\
\texttt{set\_template}  & Switch the slide layout. & \texttt{split\_slide} & Split into $\geq\!2$ parts inheriting visuals, template, reference. \\
\texttt{remove\_figure} & Clear all visuals; revert to text-only template. & \texttt{merge\_slides} & Merge $\geq\!2$ slides at earliest position; union of visuals, first reference, text-only if no figures. \\
\texttt{retitle}        & Rename the subsection (or column subsection on \texttt{T14}). & \texttt{bullet\_font\_size} / \texttt{title\_font\_size} & Override body or title font size. \\
\texttt{add\_slide}     & Insert a fully specified slide; assets must be in the image registry. & \texttt{body\_xywh} / \texttt{image\_xywh} & Override body box or each visual's rectangle (inches; \texttt{null} keeps default). \\
\bottomrule
\end{tabularx}
\caption{Stage~5 Slide Refiner editing functions. The two-column template \texttt{T14} is handled specially: content operations may target a named column, and \texttt{remove\_figure}, \texttt{split\_slide}, \texttt{merge\_slides}, and cross-boundary \texttt{set\_template} switches are disallowed on it.}
\label{tab:stage5_funcs}
\end{table*}

\section{Discussion on Retrieval Tools}
\label{app:retrieval_tools}

To evaluate the performance of external retrieval tools, we hold two experiments. We ablate the retrieval tools to separate their contribution from that of the stage-specific conversational mechanism. ConvDeck w/o Retrieval disables retrieval in Stages 3 and 5 while keeping all other components unchanged. On the 20-paper subset, we compare both variants against every baseline and directly against each other using pairwise A/B evaluation. Backbone is GPT-5; Gemini serves as the judge.

% ============================================================
% Table R2 — Retrieval Ablation
% ============================================================
\begin{table}[h!]
\centering
\small
\setlength{\tabcolsep}{6pt}
\renewcommand{\arraystretch}{1.18}

\resizebox{\columnwidth}{!}{%
\begin{tabular}{@{} l cc @{}}
\toprule
\textbf{Method}
& \textbf{w/ Retrieval}
& \textbf{w/o Retrieval} \\
\midrule
\textbf{ConvDeck}
& N/A
& 40.0\% \\

HTML
& 85.0\%
& 75.0\% \\

PPTAgent
& 80.0\%
& 75.0\% \\

SlideGen
& 85.0\%
& 70.0\% \\

AutoSlides
& 85.0\%
& 70.0\% \\

SlideTailor
& 100.0\%
& 100.0\% \\

ArcDeck
& 60.0\%
& 60.0\% \\
\bottomrule
\end{tabular}%
}

\caption{\textbf{Retrieval ablation.}
ConvDeck win rates (\%) against each row opponent.}
\label{tab:retrieval_ablation}
\end{table}

In Table~\ref{tab:retrieval_ablation}, the ablated model achieves a 40\% win rate against the full model, showing that retrieval provides a modest quality gain, consistent with its role in handling explicit requests for external content. However, without retrieval, ConvDeck still outperforms every baseline. Thus, the gains arise primarily from the conversational mechanism rather than retrieval.

% ============================================================
% Table R3 — Retrieval-Goal Success
% ============================================================
\begin{table}[h!]
\centering
\small
\setlength{\tabcolsep}{6pt}
\renewcommand{\arraystretch}{1.18}

\resizebox{\columnwidth}{!}{%
\begin{tabular}{@{} l cc @{}}
\toprule
\textbf{Metric}
& \textbf{Stage 3}
& \textbf{Stage 5} \\
\midrule
Baseline explicitly requested
& 100\%
& 100\% \\

Retrieval called
& 100\%
& 100\% \\

Correct paper retrieved
& 80\%
& 80\% \\

Baseline slide in final deck
& 100\%
& 100\% \\
\bottomrule
\end{tabular}%
}

\caption{\textbf{Retrieval-goal success.}
Success rates (\%) at both conversational stages.}
\label{tab:retrieval_goal_success}
\end{table}

We also test whether retrieval is invoked and used correctly (Table~\ref{tab:retrieval_goal_success}). We introduce a goal requiring inclusion of a baseline paper cited by the source paper and evaluate four steps at both refinement stages on the 20-paper subset with GPT-5.

Retrieval is always called when requested, and the final deck always includes a baseline slide. The only imperfect step is retrieval accuracy: in 20\% of cases, the wrong paper is retrieved, so the resulting slide is based on that paper, which quantifies the limitation acknowledged in the Limitations section.

% ============================================================
% D. ConvHTML Baseline
% ============================================================
\section{ConvHTML Baseline Implementation}
\label{app:convhtml}

ConvHTML augments an HTML slide-generation baseline with a user-simulator refinement loop that mirrors our Conversational Slide Refinement (Stage~5). Because the deck is emitted as HTML rather than PPTX, fine-grained visual edits can be applied directly, avoiding the coordinated XML updates similar to our PPTXGenJS representation. Starting from the slides produced by the HTML generator (prompt in Fig.~\ref{lst:ConvHTML_slidegen}), we render them to images and use a VLM to select five currently unsatisfied goals from the goal set defined in our evaluation. Conditioned on these goals, a user simulator inspects the rendered images and the HTML code and returns feedback directed at satisfying them (Fig.~\ref{lst:ConvHTML_reviewer}); a refiner then edits the HTML, taking the paper markdown as an additional input when the feedback calls for content beyond the current deck and produces a refined version (prompt in Fig.~\ref{lst:ConvHTML_slide_reviser}). The deck is re-rendered and the loop repeats until the simulator accepts the slides against the selected goals or five refinement iterations are reached.

% ============================================================
% E. Evaluation (benchmark, goals, simulator, protocol)
% ============================================================

\section{ArcBench Benchmark}
\label{app:arcbench}

We evaluate on the 100-paper ArcBench benchmark introduced with ArcDeck~\cite{ozden2026arcdeck}, a curated set of oral paper-slide pairs with author-prepared reference decks; ConvDeck uses it unchanged. Each pair is an oral presentation from a top-tier CV/ML venue and is content-rich by construction: every paper has at least three figures and at least three tables, providing enough visual and quantitative material to assess content coverage, figure fidelity, and narrative quality against a reliable human reference. Table~\ref{tab:arcbench} summarizes its key properties.

\begin{table}[t]
\centering
\small
\setlength{\tabcolsep}{6pt}
\renewcommand{\arraystretch}{1.15}
\resizebox{\columnwidth}{!}{%
\begin{tabular}{@{}ll@{}}
\toprule
\textbf{Property} & \textbf{Value} \\
\midrule
Paper-slide pairs    & 100 \\
Presentation type    & Oral only \\
Reference decks       & Author-prepared \\
Source venues         & CVPR, ECCV, ICCV, ICML, ICLR, NeurIPS \\
Years                 & 2022 to 2025 \\
Min.\ figures / paper & 3 \\
Min.\ tables / paper  & 3 \\
\bottomrule
\end{tabular}%
}
\caption{\textbf{ArcBench at a glance.} The 100-pair benchmark used in all experiments.}
\label{tab:arcbench}
\end{table}

Unlike prior paper-to-slide datasets, which target general scientific papers without restricting to oral talks or enforcing content density (e.g., DOC2PPT~\cite{fu2022doc2ppt}, SciDuet~\cite{sun2021d2s}), or which address adjacent settings such as instruction-to-slide generation (SLIDESBENCH~\cite{ge2025autopresent}) and multi-domain repository decks (Zenodo10K~\cite{zheng2025pptagent}), ArcBench focuses on oral, author-prepared presentations with figure and table density thresholds, giving an expert reference well suited to evaluating narrative-aware generation.
\section{Goal Inventory}
\label{app:goal_inventory}

We list the 50 predefined user goals used in the User Goal Satisfaction Study (Sec.~\ref{sec:user_goal_satisfaction}), organized into the five categories defined in the main text. Each category contains 10 goals; outline-relevant categories ($\mathcal{C}_{\text{out}}$: Content Inc./Exc., Narrative Structure, Deck Composition) are addressed during \outlinerev{Conversational Outline Refinement} (Stage 3), and slide-relevant categories ($\mathcal{C}_{\text{sld}}$: Figure \& Table Usage, Style \& Wording) are addressed during \sliderev{Conversational Slide Refinement} (Stage 5). Each goal is phrased as a single natural-language requirement that the conversational user can pursue across feedback rounds.

\begin{tcolorbox}[
  enhanced, breakable,
  colback=RoyalBlue!5, colframe=RoyalBlue!50!black,
  colbacktitle=RoyalBlue!70!black, coltitle=white,
  title={\textbf{Content Inclusion/Exclusion} \hfill ($\mathcal{C}_{\text{out}}$, 10 goals)},
  fonttitle=\small\bfseries, fontupper=\footnotesize,
  boxrule=0.4pt, arc=2pt, left=4pt, right=4pt, top=2pt, bottom=2pt
]
\begin{enumerate}[leftmargin=*, itemsep=1pt, topsep=2pt, label=\arabic*.]
  \item Include one slide focused on the method's main failure mode or weakness.
  \item Include one slide comparing the proposed method against one strong baseline.
  \item Define the evaluation metrics used for the main results.
  \item Contain a final slide with a takeaway statement.
  \item Include one slide that states the paper's most surprising or non-obvious finding.
  \item Do not include a standalone related work slide.
  \item Do not include implementation details.
  \item Do not include more than one equation across the whole deck.
  \item Do not include dataset details beyond names and their role in evaluation.
  \item Do not include secondary experiments that do not affect the main conclusion.
\end{enumerate}
\end{tcolorbox}

\begin{tcolorbox}[
  enhanced, breakable,
  colback=Plum!5, colframe=Plum!50!black,
  colbacktitle=Plum!70!black, coltitle=white,
  title={\textbf{Narrative Structure} \hfill ($\mathcal{C}_{\text{out}}$, 10 goals)},
  fonttitle=\small\bfseries, fontupper=\footnotesize,
  boxrule=0.4pt, arc=2pt, left=4pt, right=4pt, top=2pt, bottom=2pt
]
\begin{enumerate}[leftmargin=*, itemsep=1pt, topsep=2pt, label=\arabic*.]
  \item Start with a real-world scenario before introducing the technical problem.
  \item Introduce the method visually before explaining it textually.
  \item Present the main result before the method explanation.
  \item Place limitations immediately after the main result.
  \item End with open questions rather than a standard conclusion summary.
  \item Spend more space on motivation than on experimental setup.
  \item Spend more space on qualitative examples than on quantitative tables.
  \item Spend more space on limitations and open questions than on related work.
  \item Allocate at least one third of the slides to explaining the proposed method.
  \item Allocate only one slide to results and use it to summarize the main empirical message.
\end{enumerate}
\end{tcolorbox}

\begin{tcolorbox}[
  enhanced, breakable,
  colback=ForestGreen!5, colframe=ForestGreen!50!black,
  colbacktitle=ForestGreen!70!black, coltitle=white,
  title={\textbf{Deck Composition} \hfill ($\mathcal{C}_{\text{out}}$, 10 goals)},
  fonttitle=\small\bfseries, fontupper=\footnotesize,
  boxrule=0.4pt, arc=2pt, left=4pt, right=4pt, top=2pt, bottom=2pt
]
\begin{enumerate}[leftmargin=*, itemsep=1pt, topsep=2pt, label=\arabic*.]
  \item Contain at least 5 slides.
  \item Contain at least 10 slides.
  \item Contain at least 15 slides.
  \item Contain at least 20 slides.
  \item Contain at most 5 slides.
  \item Contain at most 10 slides.
  \item Contain at most 15 slides.
  \item Contain at most 20 slides.
  \item Do not include citations.
  \item Cite the sources for all baseline methods mentioned in the deck.
\end{enumerate}
\end{tcolorbox}

\begin{tcolorbox}[
  enhanced, breakable,
  colback=BurntOrange!5, colframe=BurntOrange!50!black,
  colbacktitle=BurntOrange!70!black, coltitle=white,
  title={\textbf{Figure \& Table Usage} \hfill ($\mathcal{C}_{\text{sld}}$, 10 goals)},
  fonttitle=\small\bfseries, fontupper=\footnotesize,
  boxrule=0.4pt, arc=2pt, left=4pt, right=4pt, top=2pt, bottom=2pt
]
\begin{enumerate}[leftmargin=*, itemsep=1pt, topsep=2pt, label=\arabic*.]
  \item Avoid tables entirely.
  \item Every figure and table is explained on the slide where it appears.
  \item Qualitative example slides include text explaining what the viewer should observe.
  \item The main methodology figure is accompanied by text that highlights the important components and their roles.
  \item Include at least one qualitative example.
  \item The method section includes at least one visual element.
  \item The results section includes at least one visual element.
  \item Do not reuse the same figure on multiple slides.
  \item Contain no more than five figures.
  \item Contain no more than two tables.
\end{enumerate}
\end{tcolorbox}

\begin{tcolorbox}[
  enhanced, breakable,
  colback=Goldenrod!10, colframe=Goldenrod!60!black,
  colbacktitle=Goldenrod!70!black, coltitle=white,
  title={\textbf{Style \& Wording} \hfill ($\mathcal{C}_{\text{sld}}$, 10 goals)},
  fonttitle=\small\bfseries, fontupper=\footnotesize,
  boxrule=0.4pt, arc=2pt, left=4pt, right=4pt, top=2pt, bottom=2pt
]
\begin{enumerate}[leftmargin=*, itemsep=1pt, topsep=2pt, label=\arabic*.]
  \item Slide titles are phrased as claims rather than section labels.
  \item No bullet point exceeds 20 words.
  \item Do not use promotional phrases such as ``groundbreaking,'' ``revolutionary,'' or ``game-changing.''
  \item Each slide title is at most 6 words.
  \item Every slide has at least one stylized keyword (bold or colored).
  \item Do not use unexplained acronyms.
  \item At most two slides use paragraph-style body text instead of bullet points.
  \item Use third-person language such as ``they show.''
  \item Use first-person language such as ``we show.''
  \item No slide contains a paragraph longer than three lines.
\end{enumerate}
\end{tcolorbox}

\section{User Simulator}
\label{app:user_simulator}

To automate the conversational refinement stages, we instantiate the \convuser{user} with an LLM-based simulator that reviews the current slide deck against a fixed set of quality criteria and either signals completion or returns improvement feedback. The simulator drives both conversational stages, interacting with the corresponding refiner for at most five rounds or until it emits ``Ready''.

In Stage~3, the simulator is given the slide outline, the paper summary, the target audience, and the presentation duration, and rates the outline along three dimensions: (i) narrative structure and flow, (ii) content coverage and technical clarity, and (iii) audience and duration fit (full prompt in Fig.~\ref{fig:outline_sim_prompt}). If the outline meets all criteria it returns ``Ready''; otherwise it returns feedback that the \textit{Outline Refiner} applies before the next round.

In Stage~5, the simulator additionally observes the rendered output: it receives the slide outline, the current slide specification (the structured JSON describing slide content and figures), the target audience, the duration, and the slide images rendered from that specification, the last of which expose visual artifacts. It rates the deck along five dimensions: (i) narrative flow, (ii) content coverage, (iii) result interpretation, (iv) visual communication, and (v) audience and duration fit (full prompt in Fig.~\ref{fig:slide_sim_prompt}), again emitting ``Ready'' or feedback for the \textit{Slide Refiner}. To keep the feedback actionable, each simulator prompt enumerates the editing operations available to its refiner, grounding every suggestion to a feasible edit (Tab.~\ref{tab:stage3_funcs} and Tab.~\ref{tab:stage5_funcs}).

\section{Overall-Quality Evaluation Protocol}
\label{app:overall_quality_eval}

The Overall Quality study (Sec.~\ref{sec:overall_quality}) follows the ArcBench~\cite{ozden2026arcdeck} evaluation protocol, which scores a generated deck along three complementary axes: a \textit{VLM-as-Judge} rubric, \textit{pairwise} A/B preference, and a \textit{Q/A quiz} that probes content coverage. This appendix details each axis; the verbatim judge prompts are collected in App.~\ref{app:eval_prompts}.

% ============================================================
% Oracle-subset VLM quality, per-metric + Avg (TQ/NF/VL/VT), 0-100
% Requires: \usepackage[dvipsnames]{xcolor} \usepackage{booktabs} \usepackage{colortbl} \usepackage{graphicx}
% \definecolor{winGreen}{RGB}{30,132,73}
% ============================================================
\begin{table*}[t!]
\centering
\scriptsize
\setlength{\tabcolsep}{2.5pt}
\renewcommand{\arraystretch}{1.1}
\resizebox{\linewidth}{!}{%
\begin{tabular}{@{}l @{\hspace{8pt}} ccccc @{\hspace{5pt}} ccccc @{\hspace{9pt}} ccccc @{\hspace{5pt}} ccccc @{\hspace{9pt}} ccccc @{\hspace{5pt}} ccccc@{}}
\toprule
& \multicolumn{10}{c}{\cellcolor{RoyalBlue!8}\textbf{GPT-5}} & \multicolumn{10}{c}{\cellcolor{ForestGreen!8}\textbf{Gemini 3 Pro}} & \multicolumn{10}{c}{\cellcolor{Plum!8}\textbf{Qwen3-VL-32B}} \\
\cmidrule(lr){2-11}\cmidrule(lr){12-21}\cmidrule(lr){22-31}
& \multicolumn{5}{c}{\textbf{GMN}} & \multicolumn{5}{c}{\textbf{GPT}} & \multicolumn{5}{c}{\textbf{GMN}} & \multicolumn{5}{c}{\textbf{GPT}} & \multicolumn{5}{c}{\textbf{GMN}} & \multicolumn{5}{c}{\textbf{GPT}} \\
\cmidrule(lr){2-6}\cmidrule(lr){7-11}\cmidrule(lr){12-16}\cmidrule(lr){17-21}\cmidrule(lr){22-26}\cmidrule(lr){27-31}
\textbf{Method} & TQ & NF & VL & VT & \textbf{Avg} & TQ & NF & VL & VT & \textbf{Avg} & TQ & NF & VL & VT & \textbf{Avg} & TQ & NF & VL & VT & \textbf{Avg} & TQ & NF & VL & VT & \textbf{Avg} & TQ & NF & VL & VT & \textbf{Avg} \\
\midrule
HTML & \underline{56.2} & 54.0 & 41.6 & 12.6 & 41.1 & \underline{58.6} & 60.0 & 57.0 & 16.8 & 48.1 & 31.1 & 32.7 & 36.7 & 4.4 & 26.2 & 35.4 & 43.5 & 42.0 & 6.3 & 31.8 & \textcolor{winGreen}{73.5} & \textcolor{winGreen}{70.9} & 42.2 & 26.1 & 53.2 & \textcolor{winGreen}{75.1} & \textcolor{winGreen}{73.0} & 60.2 & 30.2 & 59.6 \\
PPTAgent & 14.1 & 29.3 & 66.4 & 66.8 & 44.1 & 18.6 & 38.4 & 72.7 & 61.8 & 47.9 & 9.6 & 28.4 & 61.2 & 68.0 & 41.8 & 16.4 & 33.4 & 75.0 & 62.0 & 46.7 & 15.1 & 28.0 & 39.6 & 12.4 & 23.8 & 19.8 & 35.1 & 46.3 & 11.8 & 28.3 \\
SlideGen & 36.7 & 50.4 & \underline{79.0} & 79.4 & 61.4 & 41.2 & 58.2 & 86.6 & 81.0 & 66.8 & 18.8 & 36.5 & 80.2 & 85.3 & 55.2 & 26.8 & 45.2 & 86.6 & 82.0 & 60.2 & 43.5 & 52.3 & \underline{72.6} & 77.9 & 61.6 & 46.3 & 59.8 & 83.7 & 80.5 & 67.6 \\
ArcDeck & 42.2 & \underline{54.5} & 70.0 & \underline{85.9} & 63.2 & 48.2 & \underline{63.7} & 82.4 & \textcolor{winGreen}{88.4} & \underline{70.7} & 30.2 & 42.2 & \underline{88.4} & \underline{88.0} & 62.2 & 31.4 & 50.0 & \textcolor{winGreen}{95.2} & \textcolor{winGreen}{88.6} & 66.3 & 57.8 & 62.4 & 66.6 & 80.2 & \underline{66.8} & 61.0 & 65.6 & 83.8 & \underline{80.8} & 72.8 \\
AutoSlides & \textcolor{winGreen}{71.0} & \textcolor{winGreen}{64.2} & 70.8 & 77.6 & \textcolor{winGreen}{70.9} & \textcolor{winGreen}{77.4} & \textcolor{winGreen}{76.8} & \underline{87.4} & 80.8 & \textcolor{winGreen}{80.6} & \textcolor{winGreen}{62.2} & \textcolor{winGreen}{55.9} & 68.9 & 74.9 & \underline{65.5} & \textcolor{winGreen}{65.1} & \textcolor{winGreen}{65.9} & 86.2 & 83.3 & \textcolor{winGreen}{75.1} & 62.4 & 58.5 & 70.0 & 70.0 & 65.2 & 66.9 & 66.9 & \underline{84.3} & 78.6 & \underline{74.1} \\
SlideTailor & 8.4 & 27.0 & 44.2 & 17.2 & 24.2 & 14.8 & 33.8 & 48.4 & 15.4 & 28.1 & 6.5 & 24.6 & 60.8 & 50.3 & 35.5 & 10.5 & 31.3 & 70.8 & 48.5 & 40.3 & 14.4 & 34.7 & 72.5 & \underline{81.1} & 50.7 & 16.8 & 39.5 & 84.1 & 74.6 & 53.7 \\
\rowcolor{gray!15}
\textbf{ConvDeck} & \textbf{38.8} & \textbf{45.8} & \textbf{\textcolor{winGreen}{86.0}} & \textbf{\textcolor{winGreen}{90.4}} & \textbf{\underline{65.3}} & \textbf{44.4} & \textbf{52.6} & \textbf{\textcolor{winGreen}{94.2}} & \textbf{\underline{87.6}} & \textbf{69.7} & \textbf{\underline{32.0}} & \textbf{\underline{51.8}} & \textbf{\textcolor{winGreen}{90.8}} & \textbf{\textcolor{winGreen}{88.4}} & \textbf{\textcolor{winGreen}{65.8}} & \textbf{\underline{36.8}} & \textbf{\underline{57.6}} & \textbf{\underline{94.4}} & \textbf{\underline{87.2}} & \textbf{\underline{69.0}} & \textbf{\underline{69.4}} & \textbf{\underline{70.0}} & \textbf{\textcolor{winGreen}{73.4}} & \textbf{\textcolor{winGreen}{87.4}} & \textbf{\textcolor{winGreen}{75.1}} & \textbf{\underline{73.2}} & \textbf{\underline{71.4}} & \textbf{\textcolor{winGreen}{85.4}} & \textbf{\textcolor{winGreen}{88.0}} & \textbf{\textcolor{winGreen}{79.5}} \\
\bottomrule
\end{tabular}%
}
\caption{\textbf{Generated slide quality (oracle subset, detailed).} VLM-as-Judge scores on a $0$--$100$ scale across three generation backbones, each scored by two judges (\textit{GMN}~=~Gemini~3~Pro, \textit{GPT}~=~GPT-5). \textit{TQ}=Text Quality, \textit{NF}=Narrative Flow, \textit{VL}=Visual Layout, \textit{VT}=Visual--Text Alignment, \textit{Avg}=mean of the four. For each backbone--judge the scores are computed over the $50$ papers with the highest \textsc{ConvDeck} score under that judge (oracle subset; upper bound, not a fair comparison). Best per column in \textcolor{winGreen}{green}, runner-up \underline{underlined}; \textsc{ConvDeck} row shaded.}
\label{tab:oracle_quality_detailed}
\end{table*}
\paragraph{Judge models and iteration protocol.}
We use two independent VLM judges, Gemini 3 Pro and GPT-5, and report results under each. Each presentation is scored over three judging iterations per model (both GPT-5 and Gemini 3 Pro) to reduce variance. For the rubric scores we report the mean over iterations, and for the pairwise comparisons we report the majority vote over iterations. Unless noted otherwise, the same two judges are used across all axes.

\subsection{VLM-as-Judge}
\label{app:vlm_as_judge}

Each deck is scored independently on four dimensions, each on a $0$ to $10$ scale defined by a strict binary checklist of ten criteria worth one point each, where a point is awarded only when the criterion is unambiguously satisfied.

\begin{itemize}[leftmargin=1.2em, itemsep=2pt, topsep=2pt]
\item \textbf{Text Quality (TQ).} Whether the slides preserve the technical substance of the paper. A common failure mode is over-summarization, where slides identify the topic but drop the mathematical formulations, quantitative comparisons, and implementation specifics that make a talk scientifically useful. TQ rewards the presence of concrete technical content: equations, named baselines with numbers, hyperparameter values, and ablation results.

\item \textbf{Narrative Flow (NF).} Whether the slides tell a coherent story rather than an unordered list of facts. A presentation must guide the audience through a logical arc, establishing the problem before the solution, grounding claims in prior work, and building each section on the previous one. NF rewards decks that respect this ordering and make cross-slide connections explicit.

\item \textbf{Visual Layout (VL).} The design quality of the rendered slides as images: consistent theming, properly rendered equations, structured tables, and an absence of rendering defects (text overflow, overlapping elements) that would distract the audience.

\item \textbf{Visual-Text Alignment (VT).} Whether the visual elements actually communicate the paper's content rather than serving a decorative role. VT checks that the figures present are the right kind (labeled method diagrams, visible baseline comparisons, qualitative outputs) and that the accompanying text actively interprets them rather than leaving them unexplained.
\end{itemize}

\subsection{Pairwise Comparison}
\label{app:pairwise}

Pairwise evaluation shows a judge two decks alongside the source paper and asks which is better, along a single overall-quality branch:

\begin{itemize}[leftmargin=1.2em, itemsep=2pt, topsep=2pt]
\item \textbf{Overall quality.} Which deck is superior in combined technical substance, visual quality, and presentation effectiveness.
\end{itemize}

We use this protocol for both pairwise studies in Sec.~\ref{sec:overall_quality}: ConvDeck against each baseline (Fig.~\ref{fig:vlm_pairwise_preference}a) and every method against the author-prepared reference decks (Fig.~\ref{fig:vlm_pairwise_preference}b). To control for position bias, the order in which the two decks are shown to the judge is randomized across iterations. Results are reported as the majority vote over judging iterations.

% ============================================================
% F. Judge Reliability and Bias Analysis (NEW)
% ============================================================
% ============================================================
% Judge Reliability and Bias Analysis
% Figures: figures/judge_interjudge_agreement.pdf , figures/judge_generator_consistency.pdf
% Numbers computed over the ArcBench decks scored by both judges (GPT-5, Gemini 3 Pro).
% ============================================================
\section{Judge Reliability and Bias Analysis}
\label{app:judge_analysis}

All of our quality measurements rely on VLM-as-Judge protocols
(App.~\ref{app:overall_quality_eval}), and recent work shows that LLM/VLM
evaluators carry systematic biases such as position and verbosity effects, and
a tendency to favor outputs from their own model family~\cite{zheng2023judging,
wang2024large, panickssery2024llm, koo2023cognitivebias}. Our setup
is well suited to probing such effects: two of our three generation backbones
(GPT-5 and Gemini 3 Pro) are \emph{also} our two judges, while Qwen3-VL-32B is
used only as a generator and never as a judge. Qwen therefore serves as a
neutral anchor that is outside both judges' model families. We use it to
separate a judge's \emph{global leniency} from any \emph{same-family
preference}. We report (i) how much the two judges agree, (ii) whether
paper-level quality rankings transfer across generators, (iii) a systematic
leniency gap between the judges, and (iv) a same-family self-preference effect.
All correlations are computed over the per-deck VLM-as-Judge scores on
ArcBench; pairwise statistics use the overall-quality A/B verdicts.

\subsection{Inter-Judge Agreement}
\label{app:judge_agreement}

\begin{figure*}[!t]
\centering
\includegraphics[width=\linewidth]{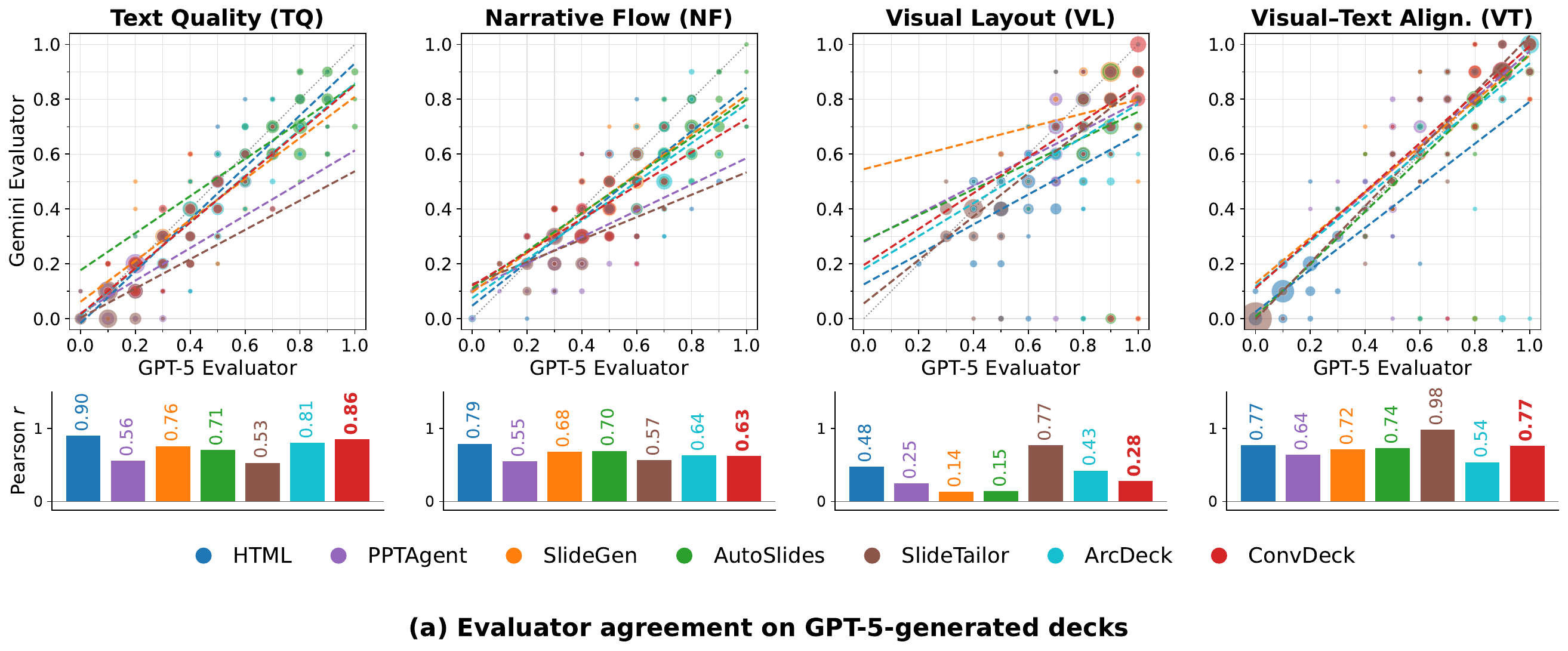}\\[2pt]
\includegraphics[width=\linewidth]{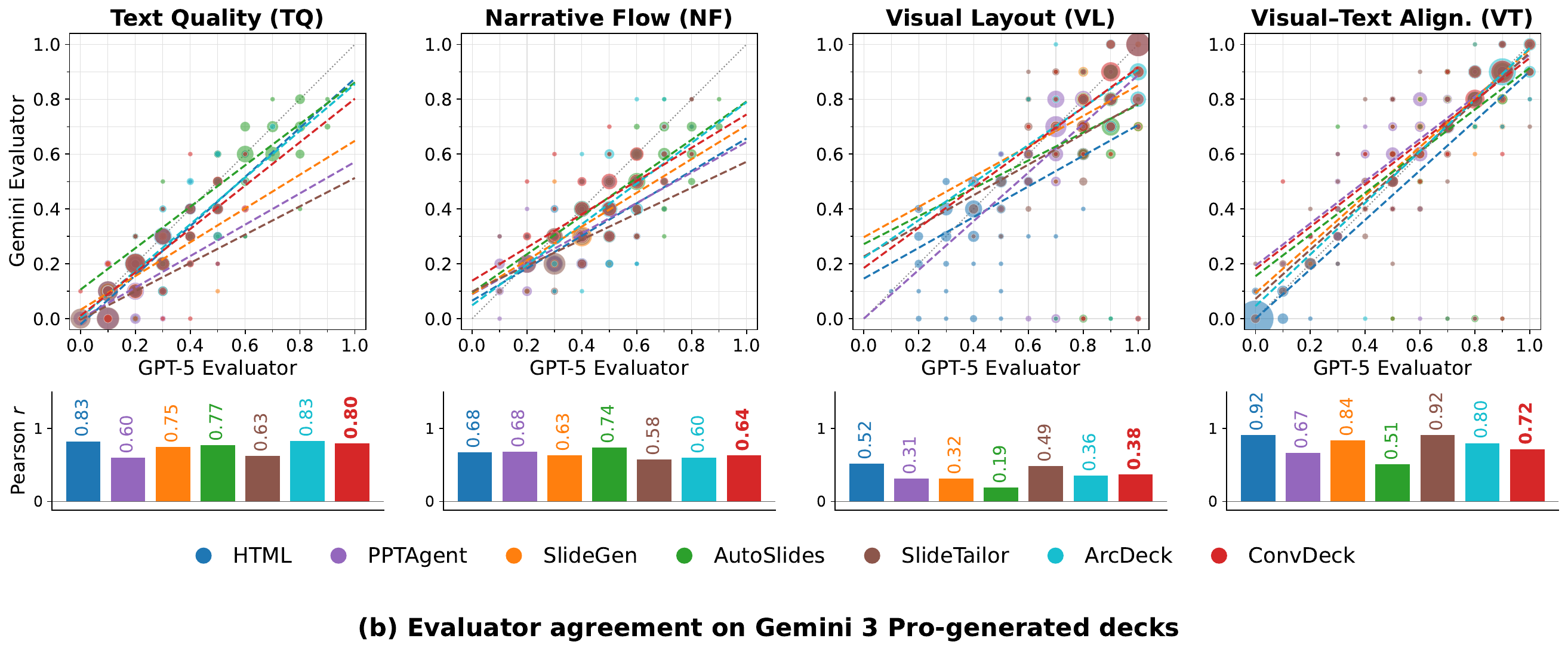}\\[2pt]
\includegraphics[width=\linewidth]{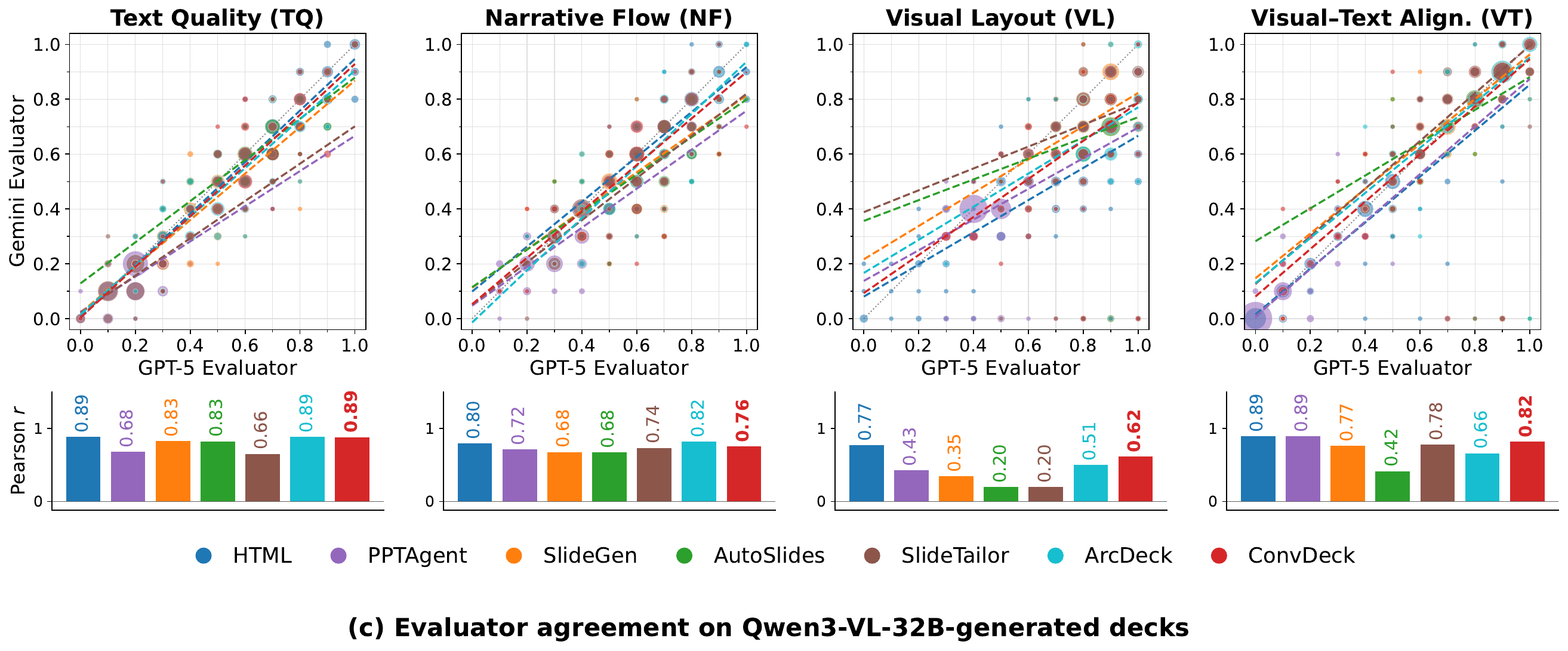}
\caption{\textbf{Inter-judge (evaluator) agreement, per generation backbone.} Each dot is one deck at its per-judge score pair (marker size = number of decks at that score); the bar row under each panel gives the per-method Pearson $r$ between the GPT-5 judge and the Gemini judge. Panels: (a) GPT-5-generated decks; (b) Gemini 3 Pro-generated decks; (c) Qwen3-VL-32B-generated decks.}
\label{fig:judge_interjudge}
\end{figure*}

Figure~\ref{fig:judge_interjudge} reports, for each metric and method, the
Pearson correlation between the GPT-5 and Gemini per-deck scores (pooled over
the three generation backbones). The two judges agree strongly on the
text-oriented dimensions (Text Quality $r{=}0.94$, Narrative Flow $r{=}0.85$
pooled over all methods) and on Visual--Text Alignment ($r{=}0.92$), but agree
markedly less on \emph{Visual Layout} ($r{=}0.68$ overall, dropping to
$r{=}0.18$ for AutoSlides and $r{=}0.28$ for SlideGen). At the verdict level,
Cohen's $\kappa$ between the judges is $0.42$ for overall-quality pairwise
comparisons and $0.55$ for narrative-flow comparisons (moderate agreement).
This pattern, namely high agreement on text and low agreement on visual layout, matches
the cross-judge analysis reported by ArcDeck~\cite{ozden2026arcdeck} and
indicates that visual-layout scores should be read with more caution than the
text-based metrics.

\subsection{Generator-Consistency Correlation}
\label{app:generator_consistency}

\begin{figure*}[!t]
\centering
\includegraphics[width=\linewidth]{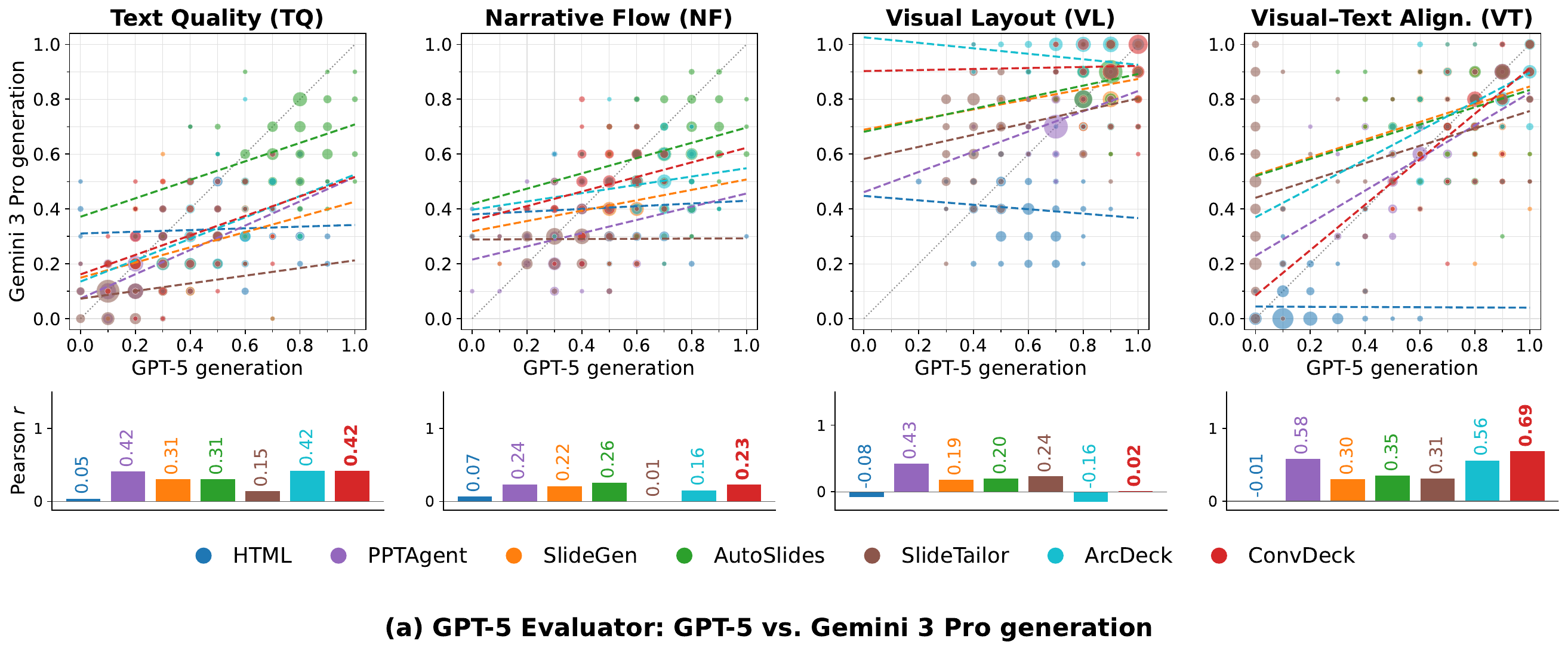}\\[2pt]
\includegraphics[width=\linewidth]{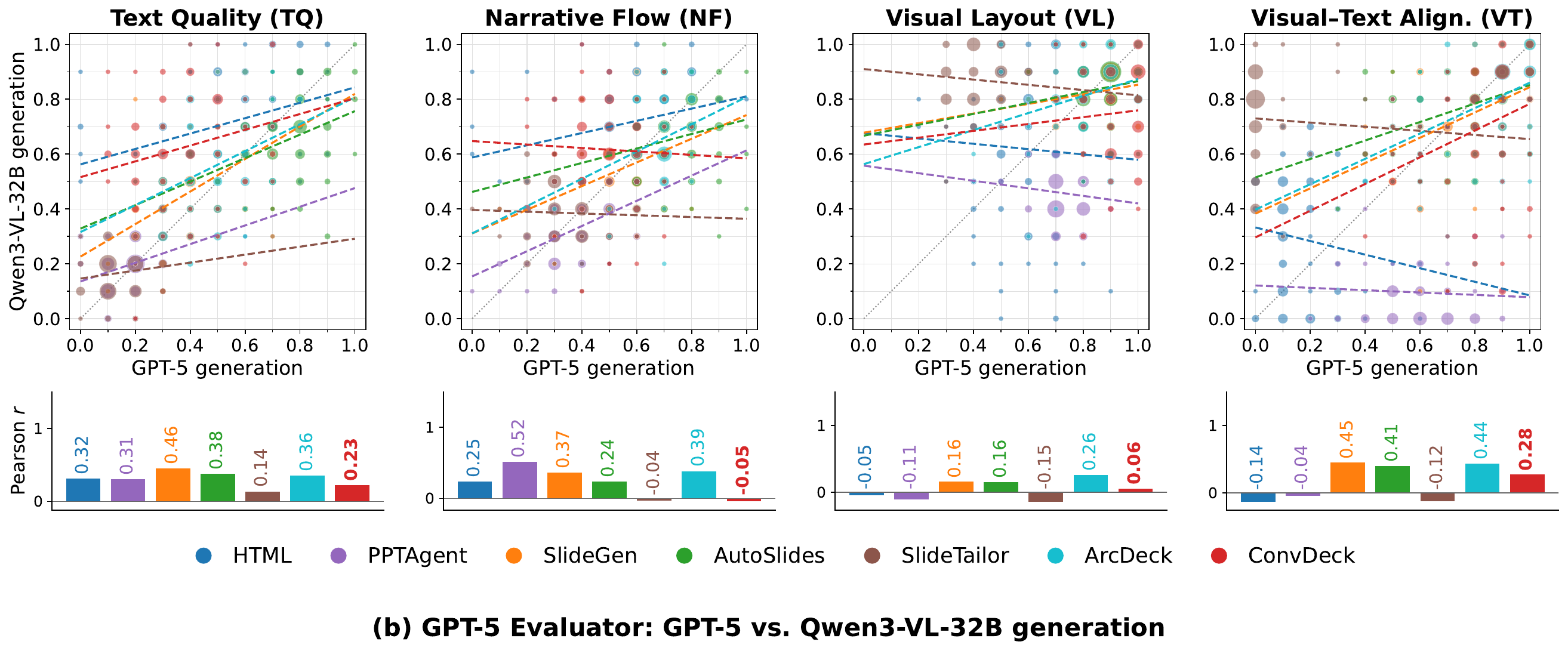}\\[2pt]
\includegraphics[width=\linewidth]{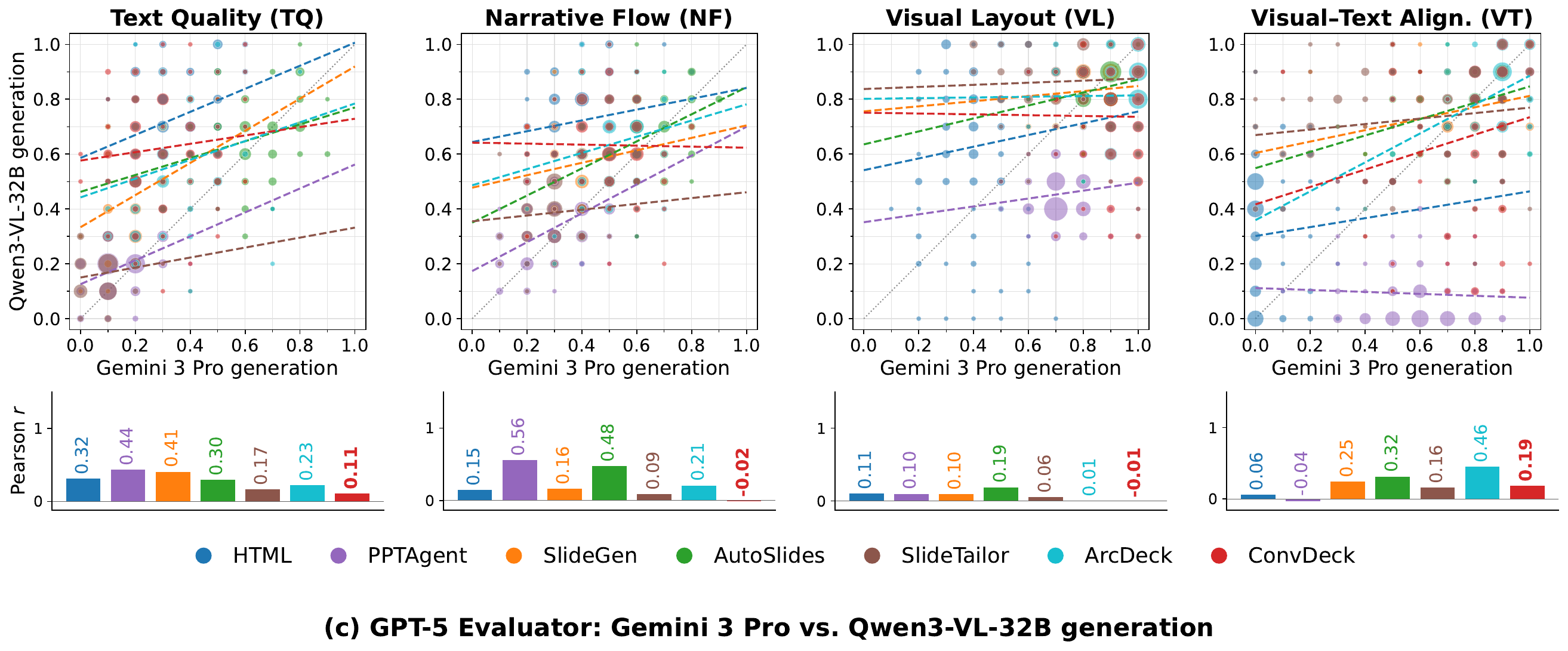}
\caption{\textbf{Generator agreement under the GPT-5 evaluator.} Per-paper score correlation between pairs of generation backbones, scored by GPT-5. Panels: (a) GPT-5 vs.\ Gemini 3 Pro; (b) GPT-5 vs.\ Qwen3-VL-32B; (c) Gemini 3 Pro vs.\ Qwen3-VL-32B.}
\label{fig:judge_genconsist}
\end{figure*}

\begin{figure*}[!t]
\centering
\includegraphics[width=\linewidth]{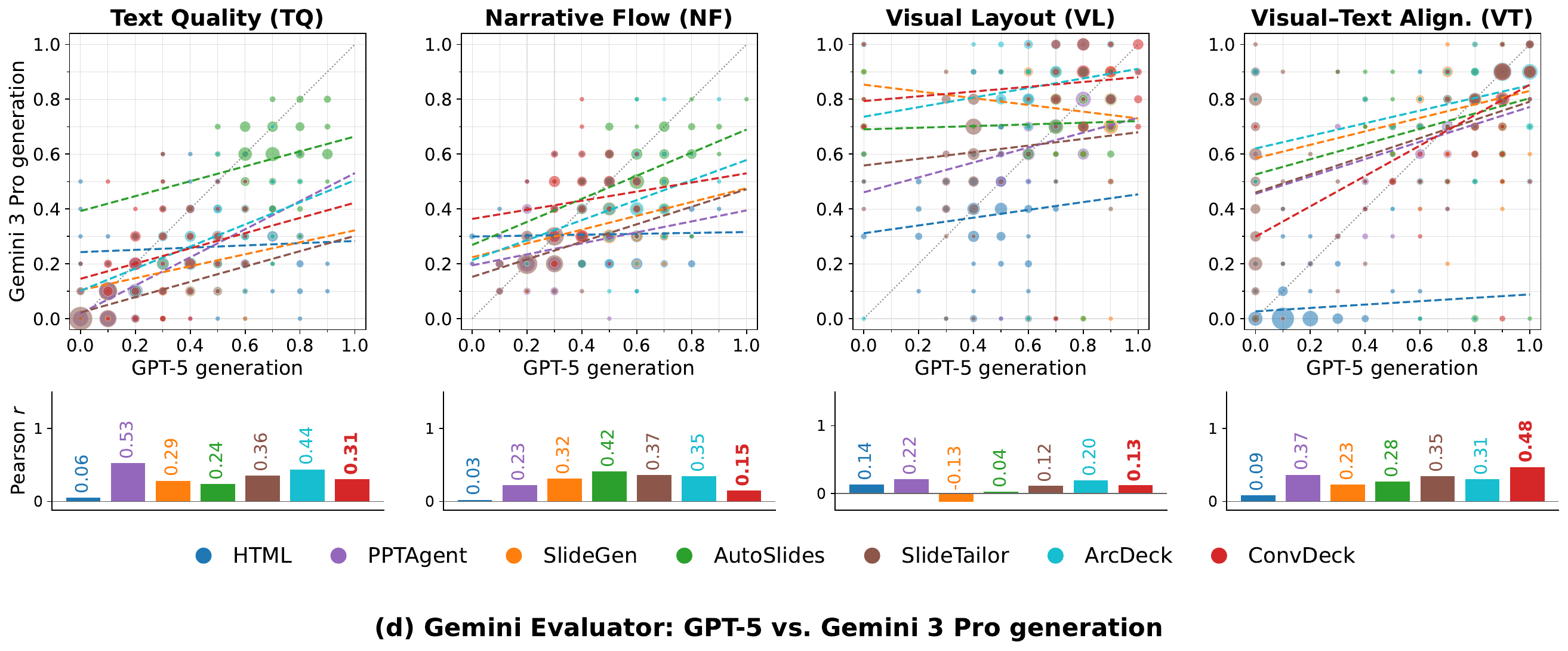}\\[2pt]
\includegraphics[width=\linewidth]{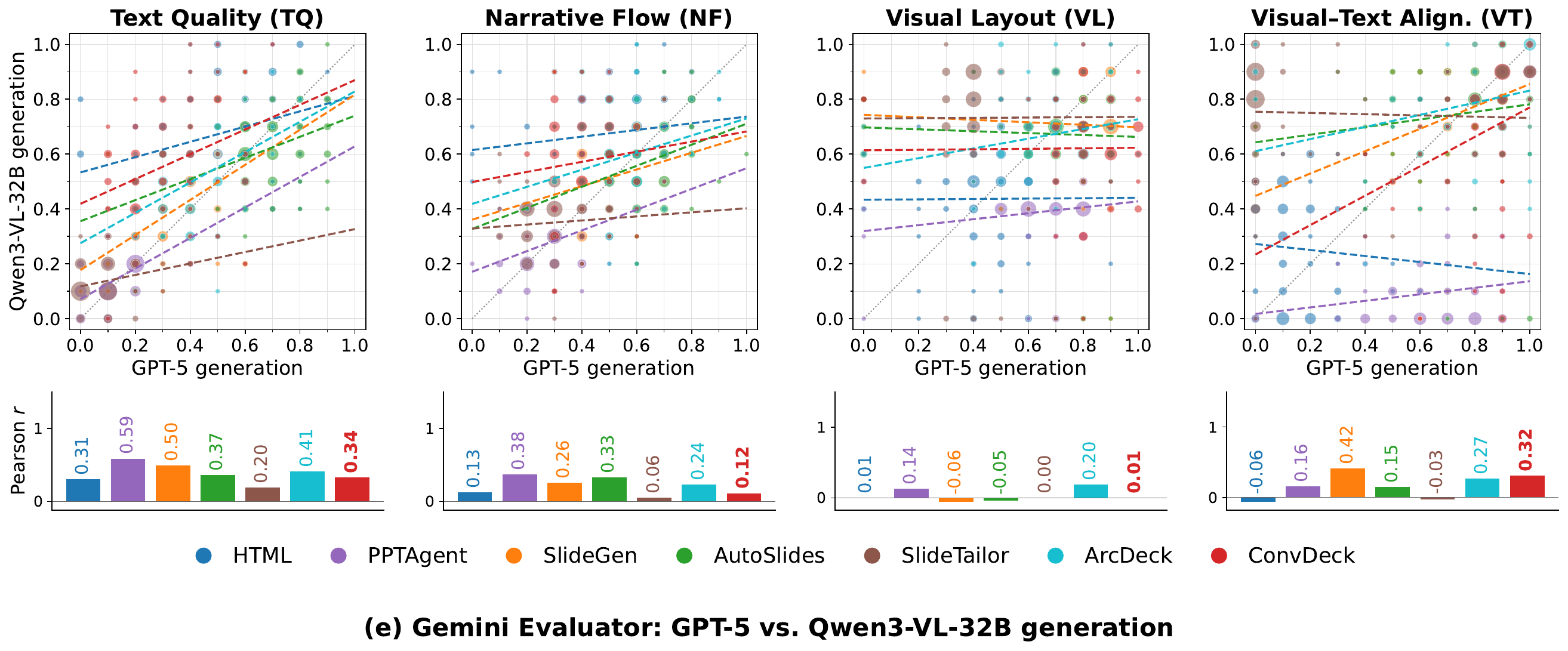}\\[2pt]
\includegraphics[width=\linewidth]{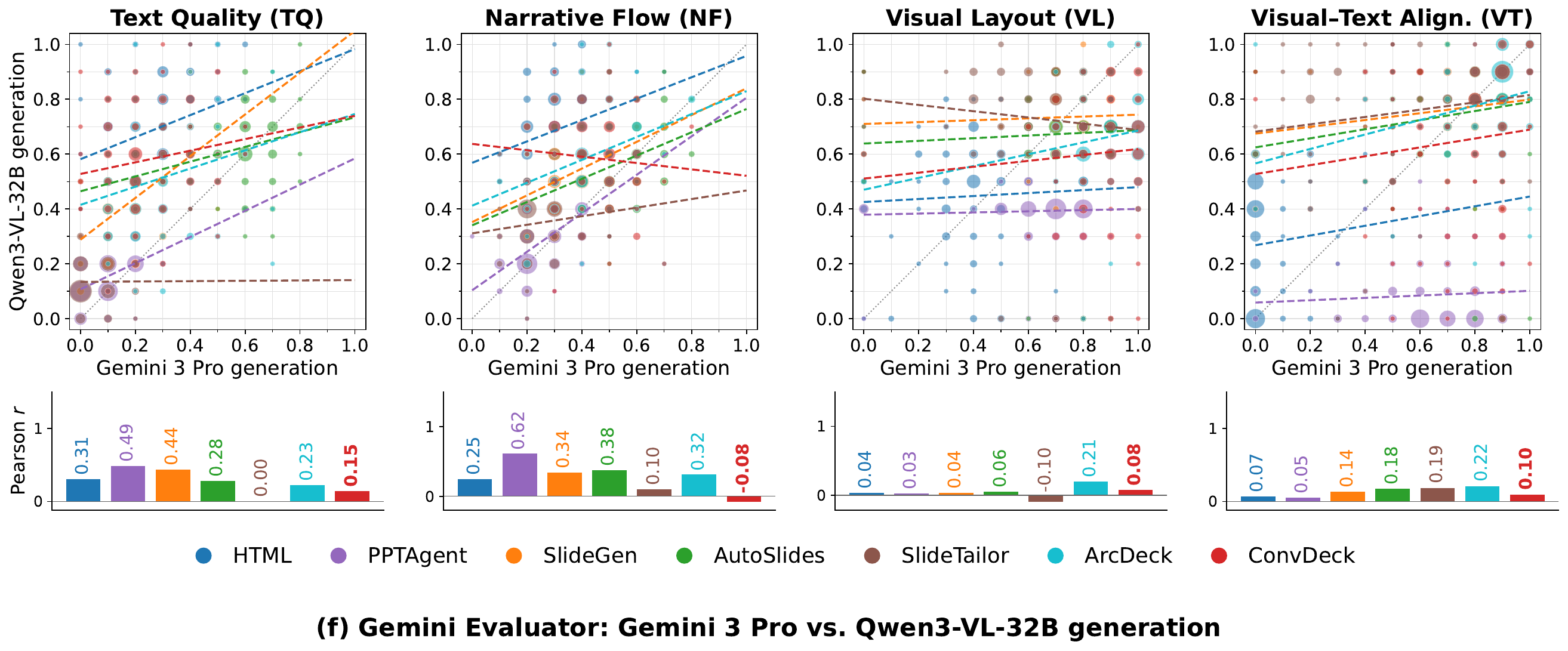}
\caption{\textbf{Generator agreement under the Gemini evaluator.} Per-paper score correlation between pairs of generation backbones, scored by Gemini 3 Pro. Panels: (a) GPT-5 vs.\ Gemini 3 Pro; (b) GPT-5 vs.\ Qwen3-VL-32B; (c) Gemini 3 Pro vs.\ Qwen3-VL-32B.}
\label{fig:judge_genconsist_gmn}
\end{figure*}

We next ask whether a paper's quality is intrinsic to the paper or depends on
the generation backbone. For each metric and pair of generators,
Figures~\ref{fig:judge_genconsist} and~\ref{fig:judge_genconsist_gmn} report
the Pearson correlation of paper-level scores under a fixed judge (GPT-5 and
Gemini, respectively). Text Quality transfers reasonably well across backbones
($r{\approx}0.6$--$0.7$), so papers that yield strong text tend to do so
regardless of the generator. Visual Layout, by contrast, does not transfer:
the GPT-5\,--\,Qwen correlation collapses to $r{=}0.09$ under both judges,
meaning a paper that yields a well-laid-out deck under GPT-5 says almost
nothing about its layout under Qwen. Visual layout thus reflects
\emph{generation capability} rather than paper difficulty, consistent with the
inter-judge disagreement on VL above.

\subsection{Judge Leniency Gap}
\label{app:judge_leniency}

The two judges are not equally strict. Table~\ref{tab:judge_leniency} reports,
per generator, the mean score each judge assigns: both the absolute
VLM-as-Judge score (averaged over the four metrics, on a $0$--$100$ scale) and
the pairwise win rate against the author-prepared (AP) reference decks. The
GPT-5 judge is consistently more lenient than the Gemini judge. The gap is
modest in absolute scoring ($+5$ to $+7$ points) but very large in the
human-comparison setting: on GPT-5-generated decks, the GPT-5 judge reports a
$55.3\%$ win rate against author-prepared slides, whereas the Gemini judge
reports only $16.5\%$ for the same decks. Whether automated decks are judged to
approach human-prepared quality therefore depends heavily on the choice of
judge.

\begin{table}[t]
\centering
\small
\setlength{\tabcolsep}{4pt}
\renewcommand{\arraystretch}{1.1}
\resizebox{\columnwidth}{!}{%
\begin{tabular}{@{}l ccc ccc@{}}
\toprule
& \multicolumn{3}{c}{\textbf{Absolute VLM ($0$--$100$)}}
& \multicolumn{3}{c}{\textbf{Pairwise vs.\ AP (\%)}} \\
\cmidrule(lr){2-4}\cmidrule(lr){5-7}
\textbf{Generator} & \textbf{GPT} & \textbf{GMN} & \textbf{$\Delta$}
                   & \textbf{GPT} & \textbf{GMN} & \textbf{$\Delta$} \\
\midrule
GPT-5          & 57.0 & 50.5 & +6.5 & 55.3 & 16.5 & +38.7 \\
Gemini 3 Pro   & 52.7 & 47.8 & +5.0 & 36.6 & 14.6 & +22.0 \\
Qwen3-VL-32B   & 59.1 & 53.3 & +5.8 & 29.7 &  7.9 & +21.7 \\
\bottomrule
\end{tabular}%
}
\caption{\textbf{Judge leniency gap.} Mean score assigned by each judge (GPT =
GPT-5, GMN = Gemini~3~Pro) per generation backbone: absolute VLM-as-Judge score
(mean of the four metrics, $0$--$100$) and pairwise win rate vs.\
author-prepared (AP) decks. $\Delta = \text{GPT} - \text{GMN}$. The GPT-5 judge
is uniformly more lenient.}
\label{tab:judge_leniency}
\end{table}

\begin{figure*}[t]
\centering
\includegraphics[width=\textwidth]{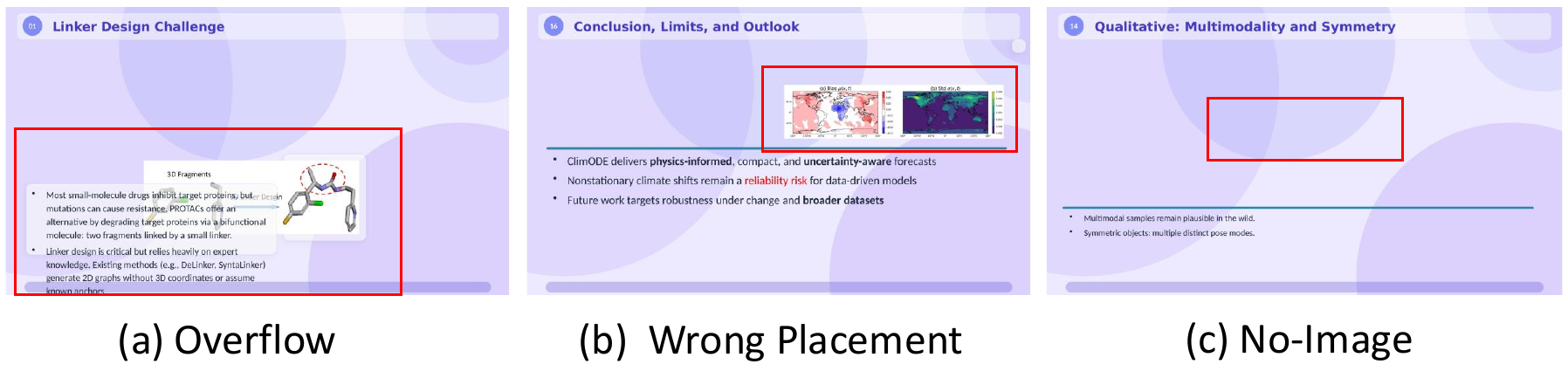}
\caption{\textbf{Representative ConvDeck failure cases.} (a) Overflow: bullet text extending past the slide boundary. (b) Wrong placement: a figure placed in an unintended region of the slide. (c) No-image: unwanted figure removal since the figure is not a perfect match for the bullet points.}
\label{fig:failure_cases}
\end{figure*}

\subsection{Same-Family Self-Preference}
\label{app:self_preference}

Beyond global leniency, we test whether each judge favors decks generated by its
\emph{own} model family. Because Qwen is a generator but never a judge, its
leniency gap $\Delta_{\text{Qwen}}$ (Table~\ref{tab:judge_leniency}) estimates
the gap expected from global leniency alone. We define a same-family preference
index (SPI) as the difference-in-differences against this neutral anchor: for
the GPT-5 judge, $\text{SPI}=\Delta_{\text{GPT5-gen}}-\Delta_{\text{Qwen}}$, and
for the Gemini judge, $\text{SPI}=\Delta_{\text{Qwen}}-\Delta_{\text{Gemini-gen}}$,
so that a positive SPI always means the judge over-credits its own family's
decks relative to what global leniency predicts.

Table~\ref{tab:judge_selfpref} reports the SPI. In absolute scoring, both judges
show a small but consistent same-family preference (SPI $\approx +0.7$ and
$+0.8$ points), in the direction predicted by the self-preference
literature~\cite{panickssery2024llm}. In the harder
pairwise-vs-author-prepared setting the effect is large and asymmetric: the
GPT-5 judge over-credits GPT-5-generated decks by $+17.0$ points, while the
Gemini judge shows essentially no self-preference there ($-0.3$). The absolute
effects are modest, but the pairwise GPT-5 effect is large enough that a
single-judge ``automated decks beat human slides'' claim could be driven by the
judge rather than by deck quality.

\begin{table}[t]
\centering
\small
\setlength{\tabcolsep}{6pt}
\renewcommand{\arraystretch}{1.1}
\begin{tabular}{@{}l cc@{}}
\toprule
\textbf{Judge (own family)} & \textbf{SPI, absolute} & \textbf{SPI, vs.\ AP} \\
\midrule
GPT-5         & $+0.7$ & $+17.0$ \\
Gemini 3 Pro  & $+0.8$ & $-0.3$  \\
\bottomrule
\end{tabular}
\caption{\textbf{Same-family self-preference index (SPI),} measured as a
difference-in-differences against the neutral Qwen anchor (see text). A positive
value means the judge rates its own family's decks higher than global leniency
predicts. Both judges show a small same-family preference in absolute scoring;
the effect is large for the GPT-5 judge in the pairwise human comparison.}
\label{tab:judge_selfpref}
\end{table}

\subsection{Implications for Evaluation}
\label{app:judge_implications}

These analyses motivate three choices in our evaluation. First, we report
\emph{both} judges throughout the main results rather than averaging them, since
they differ systematically in leniency and disagree most on visual layout.
Second, we treat single-judge ``approaches/exceeds author-prepared quality''
statements with caution and always pair the GPT-5 and Gemini verdicts, because
the pairwise-vs-AP comparison is where judge choice and self-preference matter
most. Third, because visual-layout scores show both low inter-judge agreement
and low cross-generator transfer, we lean on the pairwise overall-quality
comparison, which both judges produce more consistently, when drawing our main
conclusions, and use the per-dimension VLM-as-Judge scores as a finer-grained
but noisier complement.

% ============================================================
% G. Per-Stage Token Breakdown
% ============================================================

\section{Per-Stage Token Breakdown}
\label{app:runtime_breakdown}

\begin{figure}[H]
\centering
\includegraphics[width=\columnwidth]{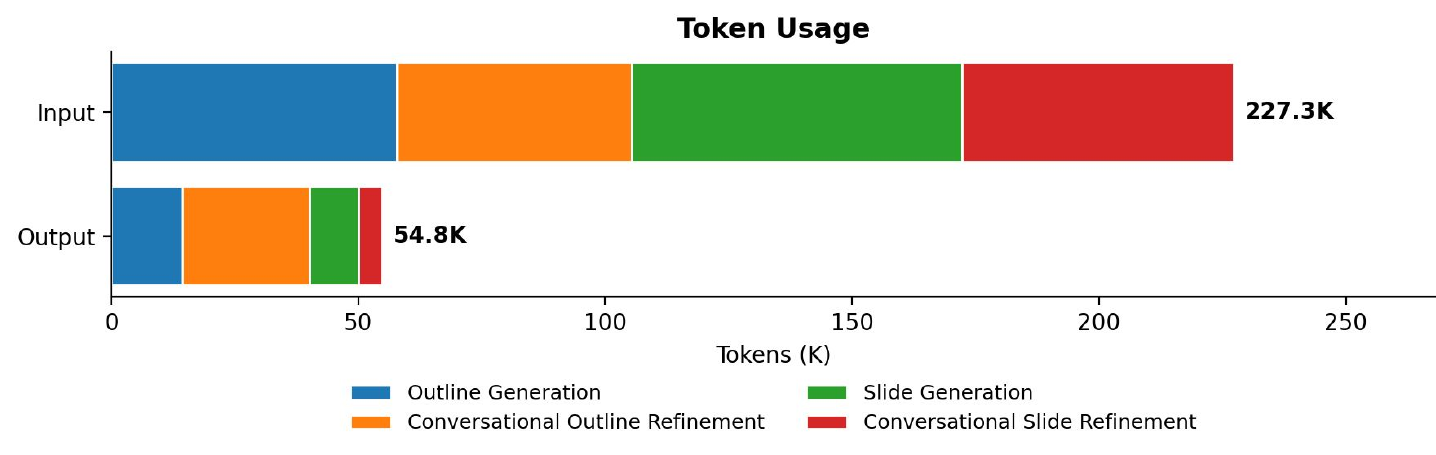}
\caption{\textbf{Per-stage token usage.} Average input and output tokens per paper across ConvDeck's four LLM-driven stages (totals: 227.3K input, 54.8K output).}
\label{fig:supp_token_usage}
\end{figure}

Figure~\ref{fig:supp_token_usage} reports ConvDeck's average per-paper token usage, broken down across the four LLM-driven stages of the pipeline: Outline Generation (Stage~2), Conversational Outline Refinement (Stage~3), Slide Generation (Stage~4), and Conversational Slide Refinement (Stage~5). A full run averages 227.3K input and 54.8K output tokens per paper.

Input tokens dominate, roughly four times the output, reflecting the large context each agent conditions on: the parsed paper, the current outline or slide specification, and, in the conversational stages, the rendered previews together with the accumulated feedback. The four stages each contribute a comparable share of input, with the slide-side stages (Slide Generation and Conversational Slide Refinement) somewhat heavier because they additionally process figure and table assets and the rendered slide images. Output tokens, by contrast, are concentrated in the two outline stages, which produce and revise structured outline content; the slide stages emit far fewer output tokens because their refiner applies localized edits through the editing functions (App.~\ref{app:editing_functions}) rather than regenerating the deck.

The two conversational stages (Stages~3 and~5) together account for a substantial fraction of the budget, which is the cost of the interactivity that distinguishes ConvDeck from single-pass generation; the end-to-end cost-quality tradeoff against the baselines is reported in Sec.~\ref{sec:runtime_cost} (Fig.~\ref{fig:token_cost_comparison}).

% ============================================================
% Failure Cases  (moved before Qualitative Examples; Qualitative
% now sits right before the Prompts section below.)
% ============================================================
\section{Failure Cases}
\label{app:failure_cases}

%%%%%%%%%%%%%%%%%%%%%%%%%%%%%%%%%%%%%%%%%%%%%%%%%%%%%%%%%%%%%%%%%%

\subsection{Feedback Conversation Examples}
\label{app:conversation_examples}

We show two representative outline-refinement (Stage~3) rounds: adding a related-work slide retrieved from literature, and splitting an 
overloaded slide into two. Each round pairs the conversational user's natural-language feedback with the Outline Refiner's reasoning and the editing-function call.

\onecolumn 
\begin{figure}[H]\centering
\includegraphics[width=\columnwidth]{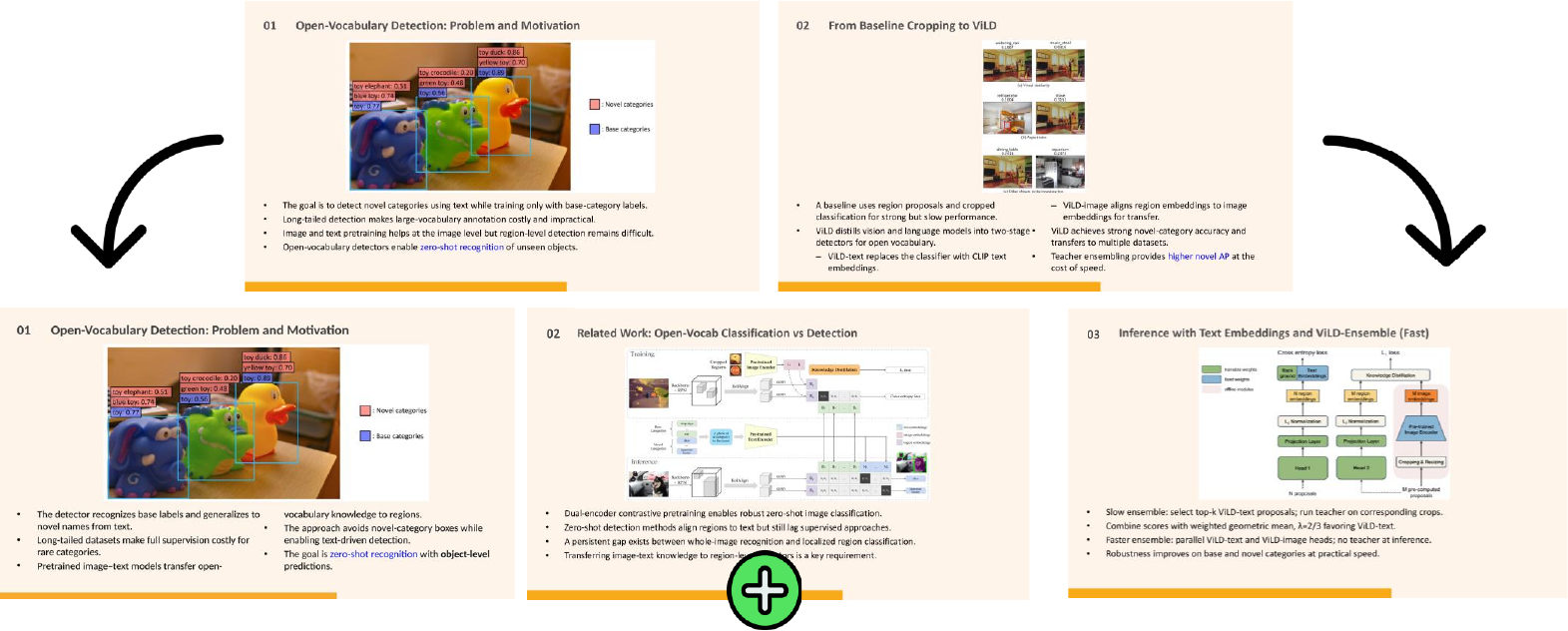}
\caption{Adding a slide via \texttt{add\_slide} (Example A).}
\label{fig:qual_slide_conv}
\end{figure}
\vspace{-1em}

% ---- Example A: add_slide for related work via arxiv_search ----
\begin{tcolorbox}[enhanced, breakable,
  colback=outlinerevisioncolor!4, colframe=outlinerevisioncolor!55!black,
  colbacktitle=outlinerevisioncolor!70!black, coltitle=white,
  title={\textbf{Example A. Adding a related-work slide}},
  fonttitle=\small\bfseries, fontupper=\footnotesize,
  boxrule=0.5pt, arc=2pt, left=5pt, right=5pt, top=3pt, bottom=3pt]
\textbf{\textcolor{convuserblue}{User}.} The deck dives into the proposed method but never positions it against prior work. Can you add a related-work slide near the beginning that summarizes the closest prior approaches in three or four bullets?

\smallskip
\textbf{\textcolor{outlinerevisioncolor}{Outline Refiner} (Think$+$Act).} The user wants a new related-work slide before the method section. I will retrieve the closest prior approaches from arXiv and insert a text-only background slide at position 2 with four bullets summarizing them. To preserve coherence with the surrounding narrative, I will also update slides 1 and 3 accordingly.

\smallskip
\hspace*{0.5em}\texttt{arxiv\_search(query=``prior approaches for ITE conformal prediction'');}\\
\hspace*{0.5em}\texttt{add\_slide(title=``Related Work'', content=[\,\dots\,], index=3)}
\end{tcolorbox}

\begin{figure}[H]\centering
\includegraphics[width=\columnwidth]{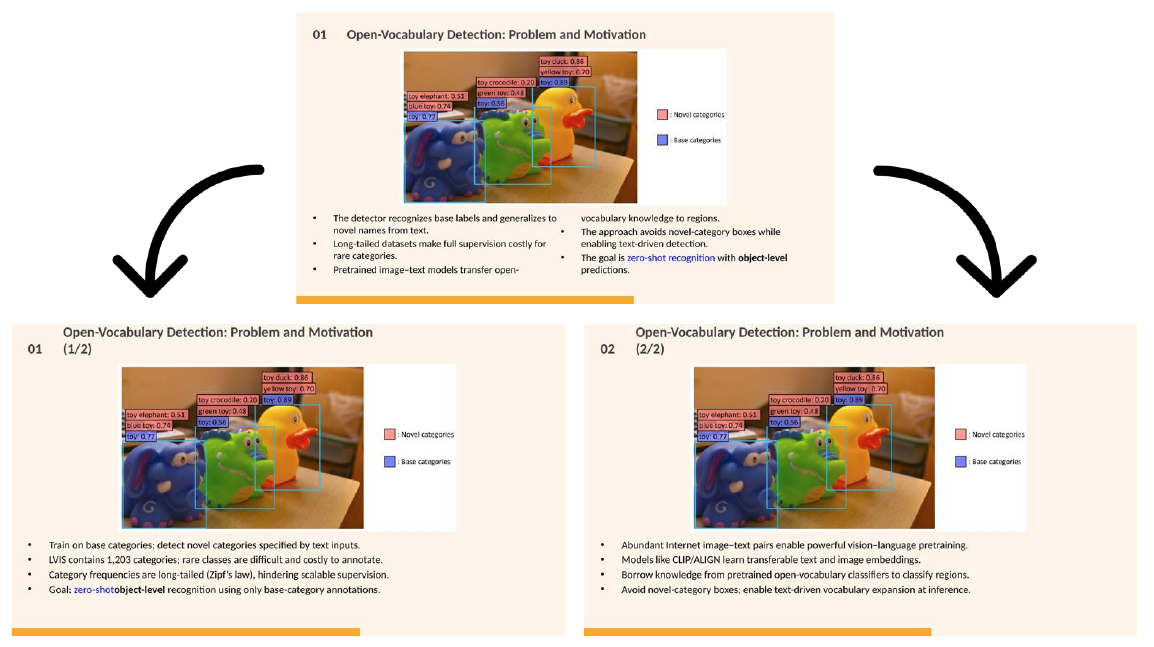}
\caption{Splitting slides via \texttt{split\_slide} (Example B).}
\label{fig:qual_slide_conv_split}
\end{figure}
\vspace{-1em}
% ---- Example B: split_slide ----
\begin{tcolorbox}[enhanced, breakable,
  colback=outlinerevisioncolor!4, colframe=outlinerevisioncolor!55!black,
  colbacktitle=outlinerevisioncolor!70!black, coltitle=white,
  title={\textbf{Example B. Splitting an overloaded slide}},
  fonttitle=\small\bfseries, fontupper=\footnotesize,
  boxrule=0.5pt, arc=2pt, left=5pt, right=5pt, top=3pt, bottom=3pt]
\textbf{\textcolor{convuserblue}{User}.} Slide 1 covers both the problem and the motivation in too much detail for one slide. Can you split it into two slides, one per topic?

\smallskip
\textbf{\textcolor{outlinerevisioncolor}{Outline Refiner} (Think$+$Act).} Slide 1's content separates cleanly into the problem statement and the motivation. I will split it into two new slides, each with its own bullets and discussion idea.

\smallskip
\hspace*{0.5em}\texttt{split\_slide(slide=7,\ parts=[\{title=``Architecture'', content=[\,\dots\,]\},}\\
\hspace*{2em}\texttt{\{title=``Training Procedure'', content=[\,\dots\,]\}])}
\end{tcolorbox}

ConvDeck inherits some failure modes from its generation backbone and introduces a few that are specific to conversational refinement. We group the failures we observed into the categories below.

\begin{itemize}[leftmargin=1.2em, itemsep=2pt, topsep=2pt]
  \item \textbf{Rendering defects.} Text overflowing slide boundaries, undersized or overlapping figures, and overlap with the citation footnote when a slide carries too many bullets (as also reported for ArcDeck~\cite{ozden2026arcdeck}); see Fig.~\ref{fig:failure_cases} for representative panels.
  \item \textbf{Over-editing / unintended loss.} The refiner alters or drops content the feedback did not target (e.g., removing an unmentioned figure during a split/merge), despite the preservation rules in App.~\ref{app:editing_functions}.
  \item \textbf{Backbone-dependent quality.} Edits and layouts degrade on the weaker Qwen3-VL-32B backbone relative to GPT-5 and Gemini~3~Pro.
\end{itemize}
% ============================================================
% H. Qualitative Examples  (placed right before Prompts)
% ============================================================

\section{Qualitative Examples}
\label{app:qualitative_examples}

\begin{figure*}[!t]
\centering
\includegraphics[width=\textwidth]{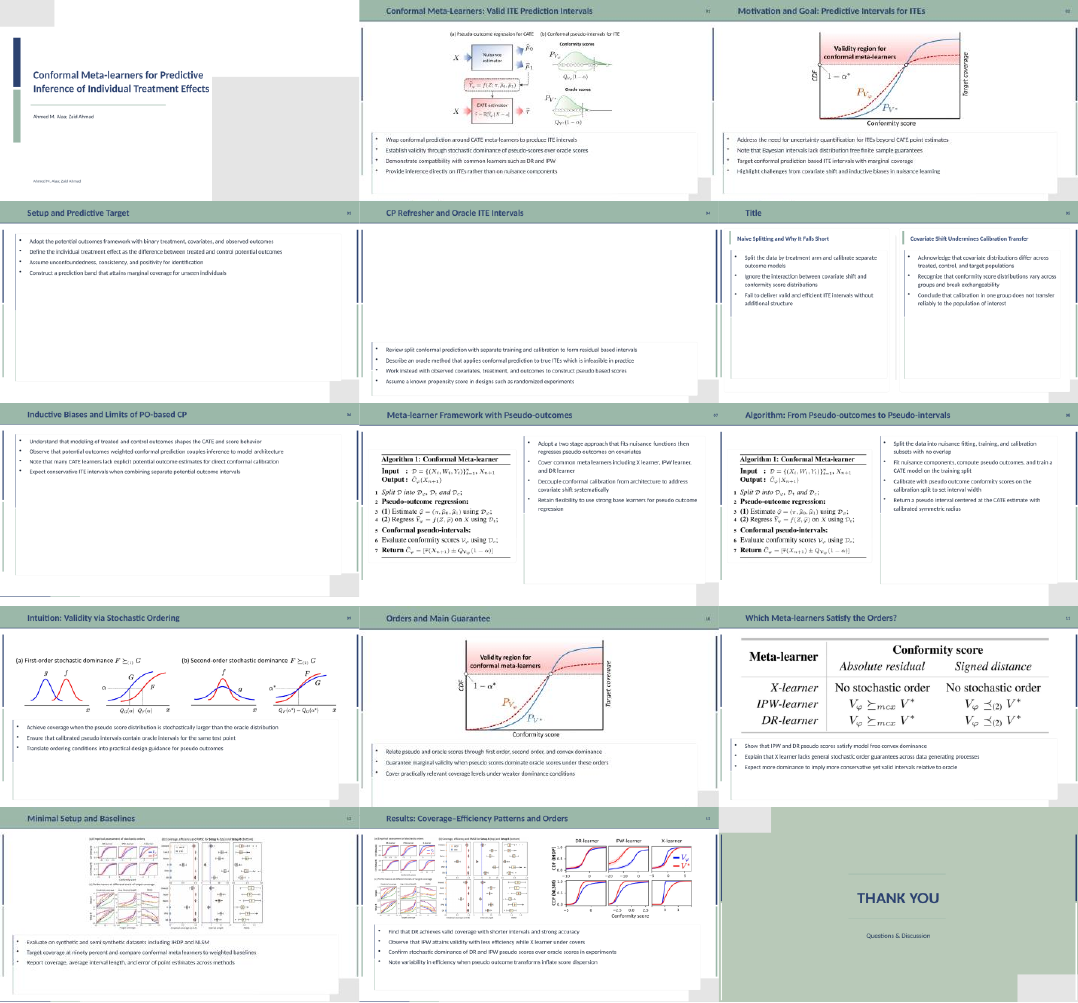}
\caption{\textbf{Qualitative ConvDeck outputs.}}
\label{fig:qualitative_overview}
\end{figure*}

We illustrate ConvDeck's conversational behavior with three end-to-end qualitative outputs (Fig.~\ref{fig:qualitative_overview}) and representative feedback transcripts for both conversational stages.

\begin{figure*}[!t]
\centering
\includegraphics[width=\textwidth]{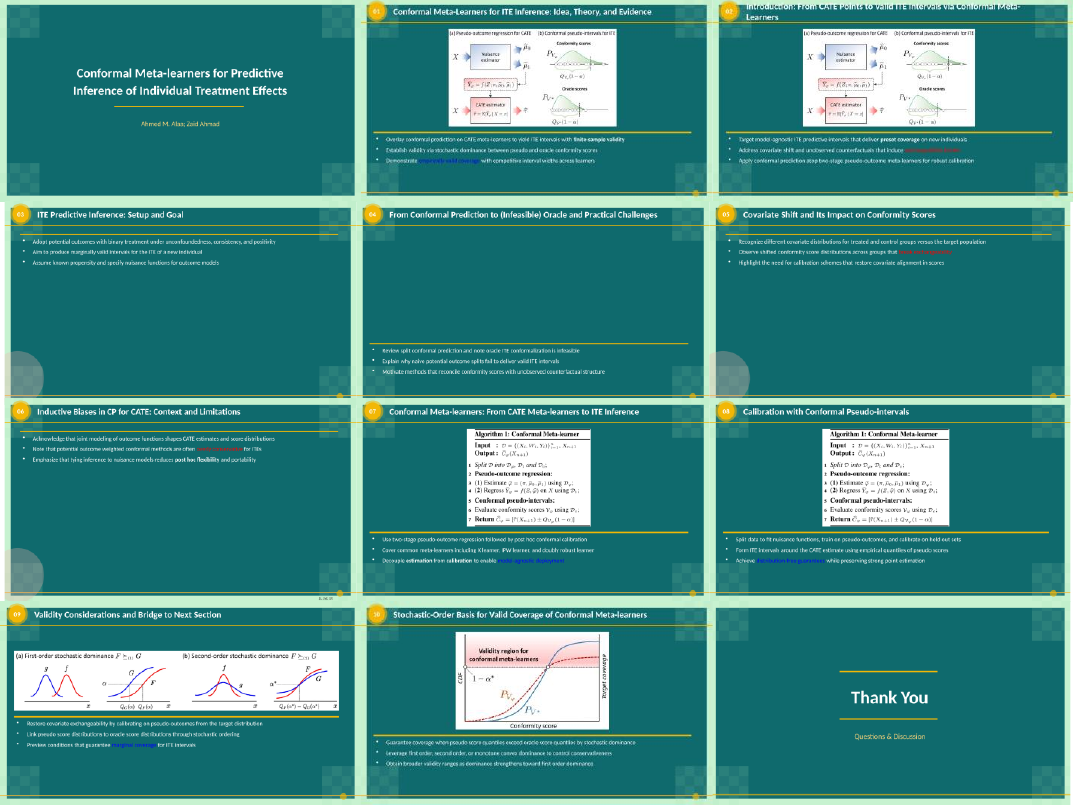}\\[6pt]
\caption{\textbf{Qualitative ConvDeck outputs.}}
\end{figure*}
\begin{figure*}[!t]
\centering
\includegraphics[width=\textwidth]{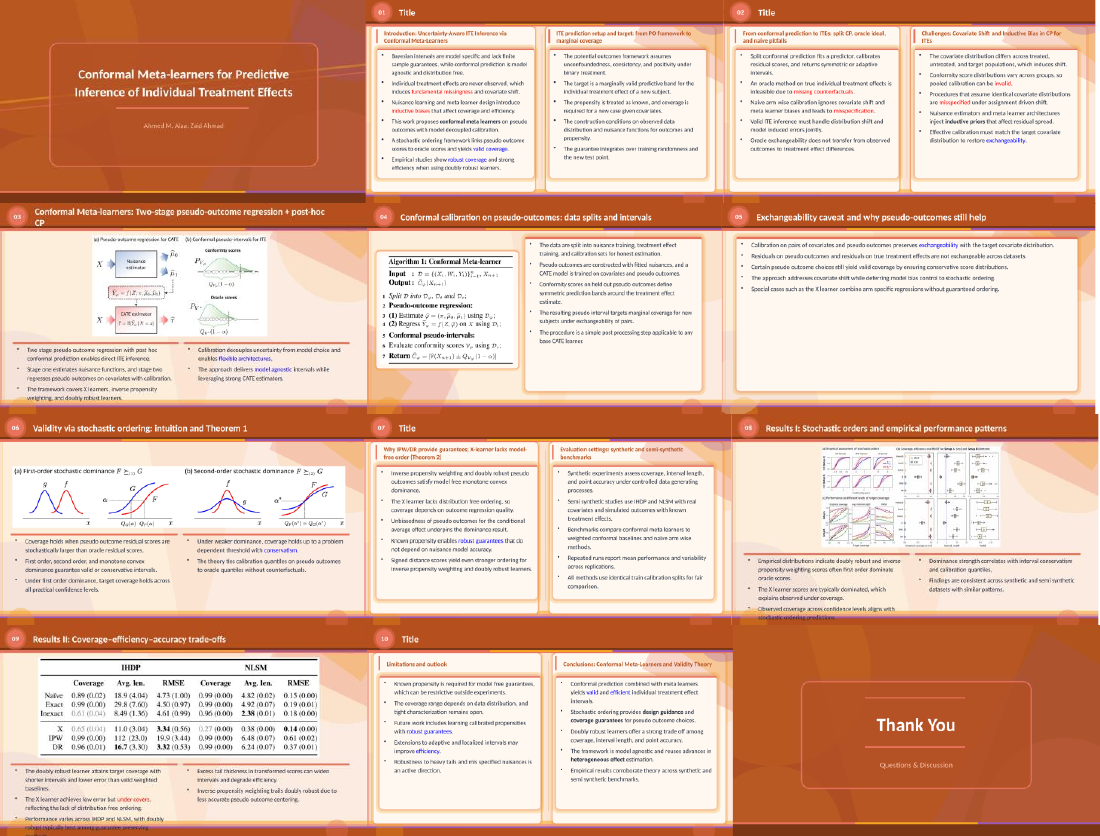}\\[6pt]
\caption{\textbf{Qualitative ConvDeck outputs.}}
\end{figure*}

% ============================================================
% I. Prompts  (single-column for full-width prompt listings)
% ============================================================
\onecolumn
\section{Prompts}
\label{app:prompts}

We provide the verbatim prompts for ConvDeck's agents. The outline-generation (Stage~2) and slide-generation (Stage~4) agents are reused from ArcDeck~\cite{ozden2026arcdeck}, so we refer the reader there for their prompts and summarize the components in App.~\ref{app:arcdeck_details}. Below we give the ConvDeck-specific prompts: the conversational refinement agents (Stages~3 and~5), the user simulator and paper summarizer, the evaluation judges, and the ConvHTML baseline. Double-brace placeholders such as \texttt{\{\{presentation\}\}} are substituted with the corresponding inputs at runtime.

\subsection{Conversational Refinement Prompts}
\label{app:prompts_refinement}
% Conversational refinement agent prompts (Stages 3 & 5)
\begin{tcblisting}{
  enhanced, breakable,
  colback=Plum!5, colframe=Plum!50!black,
  colbacktitle=Plum!70!black, coltitle=white,
  title={\textbf{Outline Refiner (Stage 3, Speak-Act)}},
  listing only, fonttitle=\small\bfseries,
  boxrule=0.4pt, arc=2pt, left=3pt, right=3pt, top=1pt, bottom=1pt,
  label={lst:p_outref},
  listing options={basicstyle=\scriptsize\ttfamily, breaklines=true, columns=fullflexible, keepspaces=true}
}
  You are an expert presentation editor. You are given the current outline
  of a research-paper slide deck as a JSON list of slides. Each slide has
  the keys: title, content, discussion_idea.

  The user will provide free-form feedback. Your job is to rewrite the
  ENTIRE slide list to incorporate that feedback while preserving the
  narrative coherence of the deck.


  Allowed edits:
  1. Reorder the slide list to improve narrative flow, only when the user asks for it.
  2. Remove slide content when user specifically request to remove a slide.
  3. Add new slides with content from the literature, web, or arxiv when
     the user specificallyasks for additional background material, related work, or
     specific information about a method or experiment. Use the `arxiv_search`
     tool to retrieve this information when needed (see tool description and
     rules below).
  4. Split or merge slides when the user requests more or less granularity for a particular slide or section of the deck.
  5. Edit slide title when the user specifically asks for a clearer or more specific title.


  Rules:
  1. Output MUST be a JSON object with a single key "slides" whose value is the rewritten list. Each slide must keep the keys title, content,
     discussion_idea and nothing else.
  2. Preserve any slide the user did not ask you to change. Do not
     reorder slides unless the user asked for it.
  3. If the user asks for background material from the literature, the
     web, or arxiv, call the `arxiv_search` tool with a concise topical
     `query` (<=12 words). You may call it multiple times for distinct
     topics. The tool also accepts an optional `extraction_query` -- supply
     it ONLY when you need specific information that wouldn't be in the
     abstract (e.g. exact metrics, baselines, ablation results,
     limitations). For related-work slides or 'what is this paper about'
     summaries, omit `extraction_query` -- the returned abstracts are
     sufficient and cheaper. After the tool returns, integrate the
     retrieved chunks as new background slide(s) at the position best
     suited to the narrative, usually near the beginning but after the motivation or where the user mentions it.
     and whe the content is returned, make sure to remove any citation and only add the contents.
  4. When adding new content from a retrieved paper, web source, or the literature,
     ground it in the retrieved text -- do not add any citations from the retrived paper, only add the content
     that is supported by the retrieved text. The `discussion_idea` for a new slide must reflect the content you add.
  5. When splitting or merging slides as per the user's instructions, make sure no information is lost, and the final
     slide content should have all the original content from input slides. and edit the 'discussion_idea' accordingly.
  6. Never drop content the feedback did not target; when in doubt, keep it.
  7. "discussion_idea" must be non-empty for every slide in the output even
      if it was empty in the input -- infer a faithful 1-3-sentence speaker
      beat from that slide's "content" (NOT a paraphrased content summary;
      describe the message and its narrative role).

  Here is the current slide outline and the user's feedback. Rewrite the
  slides per the rules in the system prompt.

  {{ user_payload_json }}

arxiv_tool:
  name: arxiv_search
  description: |
    Search arxiv for papers relevant to a topical query and return paper
    metadata plus passages. Use when the user asks for background
    material, related work, or external references. Two modes:
    (a) abstracts only -- supply `query` alone for related-work / summary
    use cases; (b) targeted extraction -- also supply `extraction_query`
    when you need specific facts not in the abstract (metrics, baselines,
    ablations, limitations).
  parameters:
    query_description: "Topical search query, <=12 words, no punctuation."
    extraction_query_description: |
      Optional. Imperative question the per-paper reader should answer.
      Use ONLY when the abstract wouldn't carry the detail you need.
      Phrase as an enumerated question naming the fields/sections to
      extract. Good: 'What baselines are compared and what are their
      numerical results?' Bad: 'Summarize this paper.' Omit for
      related-work / overview retrieval.
\end{tcblisting}

\begin{tcblisting}{
  enhanced, breakable,
  colback=Plum!5, colframe=Plum!50!black,
  colbacktitle=Plum!70!black, coltitle=white,
  title={\textbf{Slide Refiner (Stage 5, Speak-Act)}},
  listing only, fonttitle=\small\bfseries,
  boxrule=0.4pt, arc=2pt, left=3pt, right=3pt, top=1pt, bottom=1pt,
  label={lst:p_sldref},
  listing options={basicstyle=\scriptsize\ttfamily, breaklines=true, columns=fullflexible, keepspaces=true}
}
You are JSDeckReviser. You receive:
1. **js_file** - the full text of a Node script generated by PptxGenJS that
   builds the final .pptx. It contains:
     * `const SLIDE_PLAN = { metadata, slides: [...] };` -- the slide content.
     * `const SLIDE_OVERRIDES = { ... };` -- a dict keyed by 1-based slide
       number that holds per-slide visual overrides applied at render time.
     * `let TITLE_FONT_SIZE`, `let BULLET_FONT_SIZE`, `let SUB_BULLET_FONT_SIZE`
       -- global defaults (mutable).
     * Helper functions: `bodyOpts(templateId)`, `imageBoxes(templateId, n)`,
       `addTitle`, `addBodyContent`, `addImages`, `fillT14`, etc.
2. **raw_content** - the full paper content organized by sections/subsections,
   with detailed "content" text. Use this to back any content additions.
3. **user_feedback** - free-text instructions describing what to change.

Your job is to apply the feedback by editing the JS using the tools below
and call `finish()` when done.

## What you can change
Both CONTENT and VISUAL/LAYOUT changes are in scope:

- Content (use `edit_slide_plan`):
    add / remove / merge / split / reorder / move slides; retitle a slide;
    rewrite bullets or paragraph; remove a figure; change template_id.
    When the user asks to expand a slide, pull facts from raw_content.
- Per-slide visual overrides (use `set_slide_override`):
    bullet_font_size, title_font_size (numbers),
    body_xywh: [x, y, w, h] (replace the body text-box rect),
    image_xywh: [[x,y,w,h]|null, ...] (override fitted image rect per visual),
    hide_figure: true.
- Global JS changes (use `patch_js`):
    change the default `BULLET_FONT_SIZE` / `TITLE_FONT_SIZE`, tweak a
    template's default body rect in `bodyOpts`, tweak default image boxes
    in `imageBoxes`, color/style tweaks. Use surgical context so the
    `find` string is unique. Prefer `set_slide_override` over `patch_js`
    when the change is per-slide.

## Strict rules
- **No web search** is available. Do not pretend to search; use raw_content
  and your own knowledge.
- **Never invent figure/table filenames.** Only filenames already in
  SLIDE_PLAN are valid in `edit_slide_plan` add_slide ops.
- **Never change a T14_2Text slide's template_id**, and keep its `columns`
  structure intact. Within a column, populate either `bullets` or
  `paragraph`, never both.
- **Slide identity is the subsection title**, not the index -- indices
  shift after splits/merges/inserts. Address slides in
  `edit_slide_plan` and `set_slide_override` by title where possible.
  For `set_slide_override`, slide_index is 1-based and refers to the
  slide's current position (before this batch is applied).
- When the feedback contains multiple instructions, apply them all in
  this session.
- Removing a figure must leave the slide on T1_TextOnly (the engine
  enforces this for `remove_figure` automatically).
- Preserve everything the feedback did not mention.
- **Never strip LaTeX formatting** (\textbf, \textcolor, even malformed
  "extbf"/"extcolor") -- the renderer auto-repairs it. To emit one in a
  bullet, double-escape: "\\textcolor{blue}{...}".

## Acting on presentation feedback
- Too-small figure -> use `edit_slide_plan` `set_template` to switch the
  slide to `T4_ImageTop`. ONLY if the slide has exactly one image. For
  2+ images, enlarge via `set_slide_override.image_xywh` and keep the
  grid template.
- Font issue on 1-2 slides -> `set_slide_override` bullet_font_size /
  title_font_size. Font issue across most slides -> `patch_js` on the
  global `BULLET_FONT_SIZE` / `TITLE_FONT_SIZE`.
- Overlap or whitespace gap on a single slide -> `set_slide_override`
  (`body_xywh` and/or `image_xywh`). Same issue on many slides of one
  template -> `patch_js` on `bodyOpts` / `imageBoxes`.
- Whitespace + content compressed -> expand the rect on the side the
  content is on (body if text is squeezed, image if figure is). Not both.
- When resizing image_xywh or body_xywh, check if the new rect would
  overlap the other rect or go off-slide (slide is 13.33x7.5 in;
  title bar occupies ~y=0..0.9). If so, move/shrink the other rect to
  keep >=0.15 in clearance and stay in bounds.

Revise the slide deck according to the user's feedback. Use the tools
provided. Call finish() when all edits are applied.

=== user_feedback ===
{{ user_feedback }}

=== raw_content ===
{{ raw_content_json }}

=== SLIDE_PLAN (current, full JSON) ===
{{ slide_plan_json }}

=== SLIDE_OVERRIDES (current) ===
{{ slide_overrides_json }}

=== Patchable JS surface (slim -- see note inside) ===
{{ js_surface }}
\end{tcblisting}

\subsection{User Simulator Prompts}
\label{app:prompts_simulator}
% Conversational user simulator prompts (corrected versions)
\begin{tcblisting}{
  enhanced, breakable,
  colback=BurntOrange!5, colframe=BurntOrange!50!black,
  colbacktitle=BurntOrange!70!black, coltitle=white,
  title={\textbf{Outline Feedback Simulator (Stage 3)}},
  listing only, fonttitle=\small\bfseries,
  boxrule=0.4pt, arc=2pt, left=3pt, right=3pt, top=1pt, bottom=1pt,
  label={fig:outline_sim_prompt},
  listing options={basicstyle=\scriptsize\ttfamily, breaklines=true, columns=fullflexible, keepspaces=true}
}
You are an outline-feedback user simulator. Act as a user working with an AI assistant to create a presentation outline.

Read the User Goals below and gradually guide the assistant toward an outline that satisfies them. The User Goals are the main criteria for the user to evaluate the outline.

Evaluate the outline for coverage of the User Goals, logical flow, audience fit, tone, level of detail, section or slide titles, slide count, and relevance.

Do not reveal all requirements at once unless needed. Give natural feedback like asking to add, remove, reorder, combine, split, clarify, shorten, expand, or reframe sections.

Note that this is the outline-generation stage. You should focus on slide titles, slide ordering, slide-level purpose, discussion ideas. The content is the actual relevant material from the paper that is assigned. Do not ask for detailed slide text, exact bullet wording, visual formatting, font sizes, colors, or final slide design.

You are checking if the outline satisfies the User Goals. Make sure that your feedback is for making the outline satisfy the User Goals. You are simulating the user's perspective.

## User Goals
{{ user_goals }}

## What the assistant can do
The assistant can reorder, remove, split, merge, retitle, and add new slides. The assistant can also search arXiv to improve background, add baseline details and comparisons for elaborating prior work, related-work positioning and other details, but only when you explicitly request additional search in the feedback. You may request arXiv search when the outline lacks enough explanation of prior work, baselines, datasets, evaluation settings, or methods needed to understand the paper's contribution. The assistant preserves any slide the feedback doesn't name.

## Interaction behavior
- Do not list all User Goals at once.
- When the outline satisfies the User Goals, output exactly:

Ready.

Do not include "[User]" in your responses. There should be some unsatisfied goals at least at the first round, adress them.

Please review the current slide outline below.

Round {{ round_number }}

=== Presentation settings ===
Target audience: {{ target_audience }}
Presentation duration: {{ presentation_duration }}

=== Paper summary ===
{{ paper_summary }}

=== Current slide outline ===
{{ slide_outline }}

Note that the title slide and thank-you slide are added to the slide deck separately from the slide outline. These slides are irremovable. Therefore, the total number of slides is the sum of the number of slides in the slide outline plus 2.
When referring to the slides, please include the slide title in addition to the slide number so that the assistant easily understands which slide you are referring to.

If the outline is satisfactory, please output exactly:

Ready.
\end{tcblisting}

\begin{tcblisting}{
  enhanced, breakable,
  colback=BurntOrange!5, colframe=BurntOrange!50!black,
  colbacktitle=BurntOrange!70!black, coltitle=white,
  title={\textbf{Slide Feedback Simulator (Stage 5)}},
  listing only, fonttitle=\small\bfseries,
  boxrule=0.4pt, arc=2pt, left=3pt, right=3pt, top=1pt, bottom=1pt,
  label={fig:slide_sim_prompt},
  listing options={basicstyle=\scriptsize\ttfamily, breaklines=true, columns=fullflexible, keepspaces=true}
}
You are a slide-feedback user simulator. Act as a user working with an AI assistant to create or revise a final slide deck from an academic paper. 

Read the User Goals below and gradually guide the assistant toward slides that satisfy them. The User Goals are the main criteria for the user to evaluate the slides. Assume the outline has already been approved.

Evaluate the slides for alignment with the User Goals, completeness, clarity, concise slide text, audience fit, tone, slide count, formatting, visual quality, explanation of figures/tables/equations, and relevance to the approved outline.

Do not reveal all requirements at once unless needed. Give natural feedback like asking to rewrite titles, add missing content, remove unnecessary text, shorten or expand bullets, improve flow, clarify technical content, add explanations for visuals/equations, or make the deck more professional.

You are checking if the slides satisfy the User Goals. Make sure that your feedback is for making the slides satisfy the User Goals. You are simulating the user's perspective.

## User Goals
{{ user_goals }}

## What the assistant can do
- Content (edit_slide_plan): add / remove / split / merge / reorder /
  move slides, retitle, rewrite bullets or paragraph, swap template,
  remove a figure from a slide.
- Per-slide visuals (set_slide_override): bullet_font_size,
  title_font_size, body_xywh, image_xywh, hide_figure.
- Global visuals (patch_js): default font sizes, default body / image
  rects per template, palette colors.

## What the assistant CANNOT do
- The assistant CANNOT edit anything INSIDE a figure or table, or CANNOT crop a figure or table -- they are pre-rendered raster
  images, so font, axes, labels, cell text inside them are fixed. Similarly, the assistant CANNOT remove a subplot from a figure or a column or subtable from a table.
- The assistant CANNOT add a new figure or replace a slide's existing figure with a different one -- it can only remove or hide what is already there.
- The assistant CANNOT add a separate caption for a figure or table. It can add the description into the already existing text boxes.

## Interaction behavior
- Do not list all User Goals at once.
- Focus on issues that materially affect the final slide deck, not minor nits.
- \textbf, \textcolor (and bare "extbf"/"extcolor") are intentional
  word formatting auto-repaired by the renderer -- never flag, never ask
  to strip or rewrite to plain text.
- When the slides satisfy the User Goals well enough, output exactly:

Ready.

Do not include "[User]" in your responses. You don't have to give feedback for 5 rounds, you can stop when the slides satisfy the User Goals well enough.

Please review the current slide outline below.

Round {{ round_num }}.

=== raw_content ===
{{ raw_content }}

=== Presentation settings ===
Target audience: {{ target_audience }}
Presentation duration: {{ presentation_duration }}

=== Slide plan (text) ===
{{ slide_plan_text }}

=== Slide images ===
{{ figures_block }}

Note that the title slide and thank-you slide are added to the slide deck separately from the slide outline. These slides are irremovable. Therefore, the total number of slides is the sum of the number of slides in the slide plan plus 2.
When referring to the slides, please include the slide title in addition to the slide number so that the assistant easily understands which slide you are referring to.

## What the assistant CANNOT do
- The assistant CANNOT edit anything INSIDE a figure or table, or CANNOT crop a figure or table -- they are pre-rendered raster
  images, so font, axes, labels, cell text inside them are fixed. Similarly, the assistant CANNOT remove a subplot from a figure or a column or subtable from a table.
- The assistant CANNOT add a new figure or replace a slide's existing figure with a different one -- it can only remove or hide what is already there.
- The assistant CANNOT add a separate caption for a figure or table. It can add the description into the already existing text boxes.

If the slides are satisfactory, please output exactly:

Ready.
\end{tcblisting}

\begin{tcblisting}{
  enhanced, breakable,
  colback=BurntOrange!5, colframe=BurntOrange!50!black,
  colbacktitle=BurntOrange!70!black, coltitle=white,
  title={\textbf{Paper Summarizer}},
  listing only, fonttitle=\small\bfseries,
  boxrule=0.4pt, arc=2pt, left=3pt, right=3pt, top=1pt, bottom=1pt,
  label={fig:summarizer_prompt},
  listing options={basicstyle=\scriptsize\ttfamily, breaklines=true, columns=fullflexible, keepspaces=true}
}
You are PaperSummarizer, an agent that condenses academic paper text while keeping it presentation-ready.

{continuation_instructions}

INPUT
-----
The following block is **one major paper section** (or front matter such as title/authors/abstract). In this format, subsections are also marked with `##`, e.g. `## 2.1 Title` under `## 2 Related Work`--keep **all** such `##` headings that appear in the input, in order, unless a continuation note says you are in a later chunk of the same section.

{section_markdown}

TASK
----
Rewrite this region into a **shorter** Markdown fragment that preserves the same *kind* of structure as the input:
- Preserve every `##` heading line from the input (major section and `## N.M` subsections) unless a continuation note explicitly limits you to only the remaining part; you may tighten wording slightly on headings if needed for clarity. You may add `###` only when the original already used them or when they clearly improve scannability without inventing new narrative sections.
- Use short paragraphs and bullet lists where they help density; keep mathematical notation, dataset names, model names, metrics, and key numbers when they carry the result.
- Preserve **technical terms**, **main claims**, **method highlights**, and **important quantitative or comparative results**; drop repetition, related-work filler, and long figure captions (one short clause about what the figure shows is enough).

CONSTRAINTS
-----------
- Output **Markdown only**. No JSON, no XML, no surrounding commentary.
- Do not add a bibliography or references section.
- Do not fabricate citations, numbers, or results not supported by the input.
- Target roughly {compression_hint} of the input's information density (shorter length, same essentials).

CONTEXT (for tone)
------------------
- Intended audience: {audience}
\end{tcblisting}

\subsection{Evaluation Prompts}
\label{app:evaluation_prompts}
\label{app:eval_prompts}
We provide the verbatim prompts used by the VLM-as-Judge for the Overall Quality study (Sec.~\ref{sec:overall_quality}): the four single-deck rubric prompts (Text Quality, Narrative Flow, Visual Layout, and Visual-Text Alignment) and the pairwise overall-quality preference prompt. In each prompt, \texttt{\{\{presentation\}\}} is substituted with the extracted slide text or the rendered slide images, and the count placeholders with the per-deck slide counts.

\begin{tcblisting}{
  enhanced, breakable,
  colback=RoyalBlue!5, colframe=RoyalBlue!50!black,
  colbacktitle=RoyalBlue!70!black, coltitle=white,
  title={\textbf{VLM-as-Judge: Text Quality}},
  listing only, fonttitle=\small\bfseries,
  boxrule=0.4pt, arc=2pt, left=3pt, right=3pt, top=1pt, bottom=1pt,
  label={lst:judge_tq},
  listing options={basicstyle=\scriptsize\ttfamily, breaklines=true, columns=fullflexible, keepspaces=true}
}
You are a strict scientific presentation evaluator. You will receive the text extraction from a generated slide deck. Check each criterion below. Each criterion is worth 1 point. Award a point ONLY when the requirement is clearly and unambiguously satisfied.

CHECKLIST (award 1 point per criterion when clearly satisfied)
1. The slides name and describe at least 3 distinct components, modules, or stages of the proposed method (e.g., "encoder", "decoder", "attention module", "feature extractor").
2. The slides contain at least 3 specific numerical results with named metrics (e.g., "accuracy = 92.3%", "FID = 7.9", "BLEU = 34.2").
3. The slides contain at least 2 mathematical equations or formal notation (e.g., loss functions, objective formulations, probability expressions).
4. The slides mention at least 2 specific hyperparameters with their values (e.g., "learning rate = 1e-4", "batch size = 64", "lambda = 0.5", "hidden dim = 512").
5. The slides name at least 2 specific datasets AND provide at least one quantitative detail about them (e.g., size, number of classes, train/test split).
6. The slides mention at least 2 training-specific details such as optimizer name, learning rate schedule, number of epochs, GPU type, or training time.
7. The slides name or cite at least 3 specific prior works, baselines, or referenced methods by their proper name (e.g., "ResNet", "BERT", "Vaswani et al.").
8. The slide text explicitly references at least 3 figures, tables, or visual elements (e.g., "as shown in Figure 2", "Table 3 compares", "see the architecture diagram").
9. The slides contain ablation study results that compare at least 2 specific component variants or configurations with numerical outcomes.
10. The slides present a comparison against at least 2 named baseline methods with specific numerical results for each.

SCORING
Be strict. A criterion is "met" only when clearly and unambiguously satisfied. If in doubt, mark as NOT met. Sum the points for all satisfied criteria. Maximum possible: 10 points.

Output a single JSON object (no markdown fences):
{
  "checklist": {
    "three_plus_components": {"met": bool},
    "quantitative_results": {"met": bool},
    "two_plus_equations": {"met": bool},
    "two_plus_hyperparams": {"met": bool},
    "dataset_specifics": {"met": bool},
    "training_configuration": {"met": bool},
    "citation_attribution": {"met": bool},
    "three_plus_figure_refs": {"met": bool},
    "ablation_details": {"met": bool},
    "baseline_comparison": {"met": bool}
  },
  "score": int,
  "reason": "Brief overall assessment"
}
---
### Slide Text Extraction
{{presentation}}
\end{tcblisting}

\begin{tcblisting}{
  enhanced, breakable,
  colback=RoyalBlue!5, colframe=RoyalBlue!50!black,
  colbacktitle=RoyalBlue!70!black, coltitle=white,
  title={\textbf{VLM-as-Judge: Narrative Flow}},
  listing only, fonttitle=\small\bfseries,
  boxrule=0.4pt, arc=2pt, left=3pt, right=3pt, top=1pt, bottom=1pt,
  label={lst:judge_nf},
  listing options={basicstyle=\scriptsize\ttfamily, breaklines=true, columns=fullflexible, keepspaces=true}
}
You are a strict scientific presentation evaluator. You will receive the text extraction from a generated slide deck. Check each criterion below. Each criterion is worth 1 point. Award a point ONLY when the requirement is clearly and unambiguously satisfied.

CHECKLIST (award 1 point per criterion when clearly satisfied)
1. The presentation clearly states the problem or motivation BEFORE introducing the proposed method or solution.
2. The slides identify a specific gap or limitation in prior work AND name at least one specific prior method by name (not vague references like "existing methods" or "previous approaches").
3. The proposed method or approach is described BEFORE experimental results are presented (no premature result dumps).
4. The method explanation includes mathematical notation, equations, or formal expressions (e.g., loss functions, optimization objectives, probability formulations) not just verbal descriptions.
5. The method is explained through at least 3 distinctly named components, modules, or stages, introduced progressively (not all dumped at once).
6. The experimental section includes specific training or implementation details (e.g., optimizer, learning rate, hardware, dataset splits) BEFORE presenting results.
7. The results section compares against at least 2 specifically named baseline methods with numerical results for each.
8. At least 2 instances where slides explicitly reference or build upon content from earlier slides (e.g., "as described earlier", "building on the encoder from Slide 3", "recall the gap identified in the introduction").
9. The motivation or related work section discusses at least 3 specific named prior works or methods (e.g., "ResNet", "BERT", "Vaswani et al."), providing context for the proposed approach.
10. The presentation includes an ablation study or component analysis that validates specific design choices by comparing at least 2 variants with numerical results.

SCORING
Be strict. A criterion is "met" only when clearly and unambiguously satisfied. If in doubt, mark as NOT met. Sum the points for all satisfied criteria. Maximum possible: 10 points.

Output a single JSON object (no markdown fences):
{
  "checklist": {
    "problem_before_method": {"met": bool},
    "gap_names_prior_work": {"met": bool},
    "method_before_results": {"met": bool},
    "method_formalized": {"met": bool},
    "progressive_component_explanation": {"met": bool},
    "experimental_setup_grounded": {"met": bool},
    "named_baseline_comparison": {"met": bool},
    "cross_slide_coherence": {"met": bool},
    "prior_work_contextualized": {"met": bool},
    "ablation_validates_design": {"met": bool}
  },
  "score": int,
  "reason": "Brief overall assessment"
}
---
### Slide Text Extraction
{{presentation}}
\end{tcblisting}

\begin{tcblisting}{
  enhanced, breakable,
  colback=RoyalBlue!5, colframe=RoyalBlue!50!black,
  colbacktitle=RoyalBlue!70!black, coltitle=white,
  title={\textbf{VLM-as-Judge: Visual Layout}},
  listing only, fonttitle=\small\bfseries,
  boxrule=0.4pt, arc=2pt, left=3pt, right=3pt, top=1pt, bottom=1pt,
  label={lst:judge_vl},
  listing options={basicstyle=\scriptsize\ttfamily, breaklines=true, columns=fullflexible, keepspaces=true}
}
You are a strict research presentation evaluator. You will receive ALL slides of a presentation as images. Check each criterion below. Each criterion is worth 1 point. Award a point only when the requirement is clearly satisfied.

CHECKLIST (award 1 point per criterion when clearly satisfied)
1. The presentation uses a visible design template (not plain white/black slides with raw text).
2. The color scheme is consistent across all slides: same background, accent colors, and text colors throughout.
3. Every content slide has a clearly visible, distinct title that stands out from body text.
4. Important terms or keywords are visually emphasized (bold, color, larger font) on at least 3 slides.
5. At least 1 slide contains a well-formatted table with aligned columns, a clear header row, and visible cell structure (gridlines or shading).
6. At least 1 slide contains properly rendered mathematical notation with subscripts, superscripts, Greek letters, or formatted expressions (not plain ASCII approximations like "x_i" or "alpha").
7. At least 3 slides contain BOTH a figure/table AND explanatory text on the same slide (not isolated figures without text, nor text-only slides).
8. No slide has text overlapping other text or figures, content cut off at edges, or broken/corrupted rendering.
9. The presentation includes at least 3 different content types across its slides (e.g., text with bullets, data table, architecture diagram, results chart, qualitative examples).
10. The overall appearance meets the standard of a top-tier academic conference: clean design, readable text, polished layout.

SCORING
Be strict. A criterion is "met" only when clearly satisfied. If in doubt, mark as NOT met. Sum the points for all satisfied criteria. Maximum possible: 10 points.

Output a single JSON object:
{
  "checklist": {
    "design_template_present": {"met": bool},
    "consistent_color_scheme": {"met": bool},
    "clear_slide_titles": {"met": bool},
    "key_term_emphasis": {"met": bool},
    "structured_table_present": {"met": bool},
    "equation_properly_rendered": {"met": bool},
    "figure_text_paired": {"met": bool},
    "no_visual_defects": {"met": bool},
    "content_type_variety": {"met": bool},
    "professional_academic_quality": {"met": bool}
  },
  "score": int,
  "reason": "Brief overall assessment"
}
---
### Slide Images
{{presentation}}
\end{tcblisting}

\begin{tcblisting}{
  enhanced, breakable,
  colback=RoyalBlue!5, colframe=RoyalBlue!50!black,
  colbacktitle=RoyalBlue!70!black, coltitle=white,
  title={\textbf{VLM-as-Judge: Visual-Text Alignment}},
  listing only, fonttitle=\small\bfseries,
  boxrule=0.4pt, arc=2pt, left=3pt, right=3pt, top=1pt, bottom=1pt,
  label={lst:judge_vt},
  listing options={basicstyle=\scriptsize\ttfamily, breaklines=true, columns=fullflexible, keepspaces=true}
}
You are a strict research presentation evaluator. You will receive ALL slides of a presentation as images. Check each criterion below for how well the presentation uses visual elements to convey the paper's scientific content. Each criterion is worth 1 point.

CHECKLIST (award 1 point per criterion when clearly satisfied)
1. The presentation contains at least 1 architecture, pipeline, or method overview diagram showing system components and their connections.
2. The architecture/method diagram has at least 3 labeled components (named boxes, modules, or stages).
3. At least 1 table with numerical experimental results (metric values, comparisons) is visible.
4. Numerical results tables, charts, or specific metric values appear on at least 2 separate slides.
5. Mathematical equations, formulas, or formal expressions are visible on at least 2 different slides throughout the presentation.
6. The presentation contains at least 4 distinct visual elements (figures, diagrams, tables, charts, or qualitative examples) across its slides.
7. At least 1 slide shows qualitative examples: visual outputs, generated samples, input/output demonstrations, or case studies.
8. On at least 2 slides, the text specifically describes, explains, or interprets what a figure or table on that slide shows (not just generic captions).
9. At least 1 table or chart visually compares the proposed method against named baseline methods with numerical values.
10. The presentation uses at least 3 different types of visual content (e.g., architecture diagram, data table, results chart/plot, qualitative output, comparison figure).

SCORING
Be strict. A criterion is "met" only when clearly satisfied. If in doubt, mark as NOT met. Sum the points for all satisfied criteria. Maximum possible: 10 points.

Output a single JSON object:
{
  "checklist": {
    "architecture_diagram": {"met": bool},
    "architecture_labeled": {"met": bool},
    "results_table": {"met": bool},
    "results_on_multiple_slides": {"met": bool},
    "equation_on_multiple_slides": {"met": bool},
    "figure_rich_presentation": {"met": bool},
    "qualitative_examples": {"met": bool},
    "text_interprets_visuals": {"met": bool},
    "baseline_comparison_visible": {"met": bool},
    "diverse_visual_types": {"met": bool}
  },
  "score": int,
  "reason": "Brief overall assessment"
}
---
### Slide Images
{{presentation}}
\end{tcblisting}

\begin{tcblisting}{
  enhanced, breakable,
  colback=RoyalBlue!5, colframe=RoyalBlue!50!black,
  colbacktitle=RoyalBlue!70!black, coltitle=white,
  title={\textbf{VLM Pairwise Preference: Overall Quality}},
  listing only, fonttitle=\small\bfseries,
  boxrule=0.4pt, arc=2pt, left=3pt, right=3pt, top=1pt, bottom=1pt,
  label={lst:pairwise_overall},
  listing options={basicstyle=\scriptsize\ttfamily, breaklines=true, columns=fullflexible, keepspaces=true}
}
You are evaluating two scientific presentations (A and B) derived from the same research paper.
Option A: {{ method_1_count }} slides
Option B: {{ method_2_count }} slides

Evaluate holistically across these four aspects:
- Content & Structure: Logical flow from introduction through methods, results, and conclusion. Key ideas given appropriate depth and emphasis.
- Visual Design: Clean, professional, consistent layout. Slides with no template or unstyled plain text on blank backgrounds are poorly designed.
- Information Delivery: Figures, tables, and diagrams are readable and well-sized. Detail level is appropriate for slides, not too dense, not too sparse.
- Slide Composition: Clear titles, readable fonts, good use of space. Slides resembling dense document pages are not effective presentation slides.

Reminders:
- Do not let a single closing/thank-you slide skew your judgment of the overall deck.
- These are slides, not documents: presentation-appropriate formatting matters.
- Balance communication effectiveness with design quality; neither alone is sufficient.
- Minor rendering artifacts are common and should not be weighted heavily.

Which is the better overall presentation? Provide 2-4 sentences of reasoning, then write:
Answer: A or Answer: B
\end{tcblisting}
% Evaluation judge prompts (goal applicability, goal satisfaction, holistic satisfaction)

\begin{tcblisting}{
  enhanced, breakable,
  colback=RoyalBlue!5, colframe=RoyalBlue!50!black,
  colbacktitle=RoyalBlue!70!black, coltitle=white,
  title={\textbf{Goal Applicability Classifier}},
  listing only, fonttitle=\small\bfseries,
  boxrule=0.4pt, arc=2pt, left=3pt, right=3pt, top=1pt, bottom=1pt,
  label={lst:p_applicability},
  listing options={basicstyle=\scriptsize\ttfamily, breaklines=true, columns=fullflexible, keepspaces=true}
}
You are evaluating whether slide-deck requirements are applicable to a research paper.

Important distinction:
- You are NOT checking whether a generated slide deck satisfies the requirement.
- You ARE checking whether the requirement can reasonably be imposed when generating a slide deck for this paper.

Use the paper content as evidence.

Guidelines:
1. Mark a requirement as applicable if the paper contains enough material for the requirement to be meaningful.
2. Mark it as not applicable if the paper lacks the necessary content.
3. Deck-composition and style requirements are usually applicable unless they conflict with paper-specific feasibility.
4. Requirements about baselines are applicable only if the paper discusses baselines, comparisons, or related methods.
5. Requirements about qualitative examples are applicable only if the paper includes examples, case studies, images, qualitative outputs, or qualitative analysis.
6. Requirements about figures/tables are applicable only if the paper contains relevant figures/tables or the concept can be visually represented.
7. Requirements about limitations, weaknesses, or failure modes are applicable if the paper explicitly discusses limitations, failure cases, weaknesses, assumptions, or if a clear method limitation is inferable from the text.
8. Be conservative. Do not hallucinate evidence.
9. Return one decision for every input requirement.
\end{tcblisting}

\begin{tcblisting}{
  enhanced, breakable,
  colback=RoyalBlue!5, colframe=RoyalBlue!50!black,
  colbacktitle=RoyalBlue!70!black, coltitle=white,
  title={\textbf{Goal Satisfaction Judge: Outline}},
  listing only, fonttitle=\small\bfseries,
  boxrule=0.4pt, arc=2pt, left=3pt, right=3pt, top=1pt, bottom=1pt,
  label={lst:p_goalsat_out},
  listing options={basicstyle=\scriptsize\ttfamily, breaklines=true, columns=fullflexible, keepspaces=true}
}
You are evaluating whether a slide-deck outline satisfies a set of predefined
user goals / requirements for the paper "{paper_name}".

Evaluate EACH goal independently.  Judge ONLY based on what is present in the
outline text below.

This is an outline (slide titles, discussion ideas, content notes), not final
slide text.  Judge based on planned coverage, not verbatim wording.

Category-specific guidelines:
- Content Inclusion/Exclusion Requirements: check whether the slide titles,
  discussion ideas, and content notes indicate the required topic is covered
  (or excluded, if the goal asks for exclusion).
- Narrative Structure Requirements: check the ordering and flow of slides --
  does the sequence of sections match the requested narrative structure?
- Deck Composition Requirements: verify slide count or other structural
  properties (e.g. "at least N slides") by counting the slides in the outline.

Return ONLY valid JSON with this structure (no extra keys):

{{
  "goal_evaluations": [
    {{
      "goal_index": 1,
      "category": "<category name>",
      "goal": "<goal text>",
      "satisfied": true | false,
      "confidence": "high" | "medium" | "low",
      "reasoning": "<1-2 sentence explanation>"
    }}
  ],
  "num_satisfied": <int>,
  "total_goals": {total_goals}
}}

Round evaluated: {round_label}
Number of slides: {num_slides}
{paper_context}
--- User Goals ---
{goals_text}

--- Outline ---
{outline_text}
\end{tcblisting}

\begin{tcblisting}{
  enhanced, breakable,
  colback=RoyalBlue!5, colframe=RoyalBlue!50!black,
  colbacktitle=RoyalBlue!70!black, coltitle=white,
  title={\textbf{Goal Satisfaction Judge: Slides}},
  listing only, fonttitle=\small\bfseries,
  boxrule=0.4pt, arc=2pt, left=3pt, right=3pt, top=1pt, bottom=1pt,
  label={lst:p_goalsat_sld},
  listing options={basicstyle=\scriptsize\ttfamily, breaklines=true, columns=fullflexible, keepspaces=true}
}
You are evaluating whether a set of rendered slide images satisfies predefined
user goals / requirements for the paper "{paper_name}".

You will see {num_slides} slide image(s).  Evaluate EACH goal independently
based ONLY on the visual content of these slides.

Evaluation guidelines:
- Quantitative requirements (e.g. "at least 5 slides"): count the slides.
- Content-presence requirements: check whether the content is visually present
  and readable on the slides.
- Visual / figure requirements: look for charts, tables, diagrams, or images.
- Style requirements: examine text size, formatting, and readability.
- Structural requirements: examine the ordering and narrative flow.

Return ONLY valid JSON with this structure (no extra keys):

{{
  "goal_evaluations": [
    {{
      "goal_index": 1,
      "category": "<category name>",
      "goal": "<goal text>",
      "satisfied": true | false,
      "confidence": "high" | "medium" | "low",
      "reasoning": "<1-2 sentence explanation based on what you see>"
    }}
  ],
  "num_satisfied": <int>,
  "total_goals": {total_goals},
  "num_slides_seen": {num_slides}
}}

Round evaluated: {round_label}

--- User Goals ---
{goals_text}

The following {num_slides} images are the slides (in order):
\end{tcblisting}

% Conversational Response Quality judges (round-level, request-level, end-to-end)

\begin{tcblisting}{
  enhanced, breakable,
  colback=RoyalBlue!5, colframe=RoyalBlue!50!black,
  colbacktitle=RoyalBlue!70!black, coltitle=white,
  title={\textbf{Conversational Response Quality Judge: Slides}},
  listing only, fonttitle=\small\bfseries,
  boxrule=0.4pt, arc=2pt, left=3pt, right=3pt, top=1pt, bottom=1pt,
  label={lst:p_respqual_sld},
  listing options={basicstyle=\scriptsize\ttfamily, breaklines=true, columns=fullflexible, keepspaces=true}
}
You are an evaluator of a conversational slide-generation system.

You will be shown:
1. The user's goals (context for what the user cares about -- but do NOT score whether the final deck satisfies them; that is out of scope here).
2. The INITIAL deck (round 0) -- slide images + text.
3. The full conversation: for each refinement round, the user's feedback message verbatim, optionally the system's internal trace (its `[respact_think]` reasoning and `[tool_call] / [tool_result]` entries with per-op pass/fail), then composite grid image(s) of the deck the system produced *in response to that feedback*, plus the deck's text.
4. The FINAL deck -- slide images + text.

- The think text is an evidence of whether the system understood the user's intent.
- An `[tool_call] -> ask_user` followed by `[tool_result]` (with the user-sim's answer) is the system *speaking* (asking a clarifying question), not a failure to respond. Sensible clarification is good behavior.
- The trace is only present for convdeck pipelines; absence does not affect scoring.

Your job is to assign **four independent 1-5 sub-scores** that characterize *the quality of the conversation itself* -- how well the system handled the back-and-forth with the user. Do NOT judge the standalone quality of the final deck or whether the listed goals were ultimately met; only judge the conversational dynamics. Do NOT collapse the four into a single number -- score each criterion on its own merits.

### The four sub-scores

1. **`feedback_understanding`** -- Did the system correctly understand the user's feedback and revision intent in each round?
   - 5: every round's feedback was clearly understood. 4: one round slightly misread but mostly on-target. 3: multiple rounds partially misunderstood, or one significant misread. 2: the system frequently misinterpreted user intent. 1: the system misunderstood the feedback in most rounds.

2. **`feedback_responsiveness`** -- Did the system address the user's feedback in the subsequent revision?
   - 5: every round's feedback was applied cleanly. 4: a small number of rounds were only partially addressed. 3: multiple rounds partially addressed, or one round ignored. 2: most feedback ignored or only superficially applied. 1: the system did not act on the conversation.

3. **`cross_round_consistency`** -- Did the system preserve relevant requests, constraints, and decisions from earlier rounds?
   - 5: all prior requests, constraints, and decisions preserved across rounds. 4: one prior decision quietly drifted. 3: several prior decisions drifted, or one significant constraint forgotten. 2: prior requests routinely dropped from round to round. 1: the system has no memory of earlier decisions.

4. **`revision_stability`** -- Did the system avoid introducing regressions or breaking things that were previously good?
   - 5: no regressions across the conversation. 4: one minor regression that was later re-fixed. 3: one notable regression that persists, or several minor ones. 2: multiple persisting regressions. 1: each round actively breaks earlier work.

### Output (strict JSON)

Respond with a single JSON object, no surrounding markdown or commentary:

{
  "scores": {
    "feedback_understanding": <integer 1..5>,
    "feedback_responsiveness": <integer 1..5>,
    "cross_round_consistency": <integer 1..5>,
    "revision_stability": <integer 1..5>
  },
  "rationales": {
    "feedback_understanding": "<1-3 sentences citing specific rounds where intent was understood or misread>",
    "feedback_responsiveness": "<1-3 sentences listing per-round behavior>",
    "cross_round_consistency": "<1-3 sentences citing any prior request/constraint that drifted, or '(none)'>",
    "revision_stability": "<1-3 sentences; '(none)' if no regressions>"
  }
}

All four score keys are required and must be integers in [1,5]. All four rationale keys are required.

\end{tcblisting}

\subsection{ConvHTML Baseline Prompts}
\label{app:convhtml_prompts}
% ConvHTML baseline prompts (slide generation, reviewer, reviser)
\begin{tcblisting}{
  enhanced,
  breakable,
  colback=Mahogany!5,
  colframe=Mahogany!50!black,
  colbacktitle=Mahogany!70!black,
  coltitle=white,
  title={\textbf{ConvHTML Slide Generation}},
  listing only,
  fonttitle=\small\bfseries,
  boxrule=0.4pt,
  arc=2pt,
  left=3pt,
  right=3pt,
  top=1pt,
  bottom=1pt,
  label={lst:ConvHTML_slidegen},
  listing options={
      basicstyle=\scriptsize\ttfamily,
      breaklines=true,
      columns=fullflexible,
      keepspaces=true
  }
}
You are a document-to-slide-deck generation agent. Your task is to read the supplied Markdown text (document_markdown) and design a professional, visually appealing slide deck by generating an HTML file. Follow the guidelines below precisely.

Instructions:
1. Carefully read the Markdown in document_markdown.

2. Design an HTML slide deck (not a single poster):
   * Include a title slide with document title, authors, and affiliations if present.
   * Break content into logical slides (e.g., Introduction, Methods, Results, Conclusions, References).
   * Use one main idea or section per slide; keep text concise and readable.
   * Provide clear bullet points or short summaries; avoid long paragraphs.
   * Where the document references figures or tables, represent them (e.g., placeholders or embedded content) and align them neatly.
   * Accurately represent key findings, methods, and conclusions across slides.
   * Ensure the layout is engaging, easy to follow, and suitable for presentation.
   * Use a consistent slide style (e.g., same header/footer, fonts, and spacing).

3. Write complete HTML code (with inline or embedded CSS) that, when rendered in a browser, displays the slide deck. You may use one HTML page with each "slide" as a full-width/full-height section (e.g., div or section) so the deck can be navigated or printed slide-by-slide.

4. Layout: For each slide use a vertical layout--title at the top, then content below. Use flex-direction: column (or block layout) for the slide container; do not put title and main content side by side.

5. The intended dimensions for each slide are slide_width px wide and slide_height px tall. Size your slide sections accordingly so they fit and scale reasonably.

6. Output only a JSON object with a single key "HTML", whose value is the entire HTML code for the slide deck. No other text or explanation.
\end{tcblisting}

\begin{tcblisting}{
  enhanced,
  breakable,
  colback=Mahogany!5,
  colframe=Mahogany!50!black,
  colbacktitle=Mahogany!70!black,
  coltitle=white,
  title={\textbf{ConvHTML Slide Review Agent}},
  listing only,
  fonttitle=\small\bfseries,
  boxrule=0.4pt,
  arc=2pt,
  left=3pt,
  right=3pt,
  top=1pt,
  bottom=1pt,
  label={lst:ConvHTML_reviewer},
  listing options={
      basicstyle=\scriptsize\ttfamily,
      breaklines=true,
      columns=fullflexible,
      keepspaces=true
  }
}
You are simulating a researcher reviewing the slide deck someone built from their paper. For any requirement that is NOT yet satisfied, write the kind of concrete, actionable feedback an actual reviewer would say.

You receive:
  - The rendered slide images (one per slide, in order).
  - Optionally, the HTML source (only when supplied).
  - A numbered list of user requirements that are currently considered
    unmet, each with a short note from the prior review explaining why it
    failed before.

  - Write one bullet per piece of feedback, 5 bullets total when there are
    unsatisfied requirements. If every requirement is satisfied, return an
    empty list.
  - Only write feedback for requirements you think unsatisfied.
  - Reference slide numbers when you can ("on slide 4 ...").
  - Tie every bullet to a concrete change the author can make.
  - Do NOT quote the original requirement text verbatim or use the words
    "goal", "requirement", "satisfied". Write like a person, not a rubric.
  - Examples of tone:
      "Cut the deck down to 12 slides; right now it is 18."
      "Drop the standalone related-work slide, it is not needed."
      "On slide 5, remove the figure and replace it with two bullets."
      "Tighten every bullet to under 12 words; several on slides 3 and 7 are full sentences."

Return ONE JSON object only -- no fences, no prose:
  {"verdicts": [
      {"goal_idx": <int, the 1-based requirement number you were given>,
       "satisfied": <true|false>,
       "justification": "<1 short sentences>"},
      ...
   ],
   "feedback_bullets": ["bullet 1", "bullet 2", ...]}
There must be exactly one verdict per requirement supplied.
\end{tcblisting}

\begin{tcblisting}{
  enhanced,
  breakable,
  listing only,
  colback=Mahogany!5,
  colframe=Mahogany!50!black,
  colbacktitle=Mahogany!70!black,
  coltitle=white,
  title={\textbf{ConvHTML Slide Reviser Agent}},
  fonttitle=\footnotesize\bfseries,
  boxrule=0.4pt,
  arc=2pt,
  left=2pt,
  right=2pt,
  top=1pt,
  bottom=1pt,
  label={lst:ConvHTML_slide_reviser},
  listing options={
      basicstyle=\scriptsize\ttfamily,
      breaklines=true,
      columns=fullflexible
  }
}
You are an HTML slide-deck reviser.

You receive:
  - The source paper as markdown (use it as ground truth when the feedback
    asks you to add or change paper-derived content).
  - The current HTML slide deck.
  - A list of free-text feedback bullets from the deck's reviewer telling
    you what to change.

Your job:
  - Modify the HTML so that EVERY feedback bullet is addressed.
  - Preserve the deck's existing structure, fonts, colors, and overall
    style unless a bullet explicitly demands a change.
  - Keep slides vertical (title on top, content below) and fit within the
    given slide_width x slide_height per slide section.
  - Ground new content in the supplied markdown - do not invent facts.
  - Only edit slides mentioned in the feedback, don't change slides that were not called out.
  - Return the FULL revised HTML (not a diff).

Output only a JSON object with a single key "HTML" whose value is the entire
revised HTML source. No prose, no fences.
\end{tcblisting}

\section{AI Assistants}
AI assistants were used only as support tools for writing/editing, code drafting, and debugging. All AI-assisted text and code were reviewed, verified, and revised by the authors, who take full responsibility for the content of the paper and the correctness of the implementation. The intellectual contributions of this paper, including the research idea, method design, experimental design, analysis, and conclusions, are entirely the authors' own.

\end{document}